\documentclass[11pt]{article}

\usepackage[final]{acl}
\usepackage{tcolorbox}
\usepackage[T1]{fontenc}
\usepackage[utf8]{inputenc}

\usepackage{microtype}

\usepackage{inconsolata}

\usepackage{amsmath,amsfonts,bm}

\def\eqref#1{equation~\ref{#1}}
\def\1{\bm{1}}

\DeclareMathAlphabet{\mathsfit}{\encodingdefault}{\sfdefault}{m}{sl}
\SetMathAlphabet{\mathsfit}{bold}{\encodingdefault}{\sfdefault}{bx}{n}

\usepackage{hyperref}   % or drop this line entirely if the style already loads it
\hypersetup{
  colorlinks   = true,
  linkcolor    = blue!50!black,
  urlcolor     = blue!50!black,
  citecolor    = blue!50!black,
  anchorcolor  = blue!50!black,
}
\usepackage{url}
\usepackage{times}
\usepackage{booktabs}
\usepackage{latexsym}
\usepackage{algorithm}
\usepackage{algpseudocode}
\usepackage{tabularx} 
\usepackage{amsmath}
\usepackage{amssymb}
\usepackage{siunitx}
\usepackage[inkscapelatex=false]{svg}
\usepackage{array}
\usepackage{multirow}
\usepackage{caption}
\usepackage{graphicx}
\usepackage{subcaption}
\usepackage{cleveref}
\usepackage{enumitem}
\usepackage{calc}
\usepackage{longtable}
\renewcommand{\cite}{\citep}
\newcommand{\Sref}[1]{\S\ref{#1}}

\newcommand{\mypar}[1]{\smallskip\noindent{\textbf{{#1}.}}}

\usepackage[normalem]{ulem}

\title{The Privacy-Hallucination Tradeoff in \\ Differentially Private Language Models}

\author{Krithika Ramesh \\
  Johns Hopkins University \\
  \texttt{kramesh3@jh.edu} \\\And
  Krishna Pillutla \\
  Indian Institute of Technology, Madras \\
  \texttt{krishnap@dsai.iitm.ac.in} \\\AND
  Danish Pruthi \\
  Indian Institute of Science \\
  \texttt{danishp@iisc.ac.in} \\\And
  Anjalie Field \\
  Johns Hopkins University \\
  \texttt{anjalief@jhu.edu} \\}

\begin{document}
\maketitle
\addtocontents{toc}{\protect\setcounter{tocdepth}{-1}}
\begin{abstract}

Both privacy and factual accuracy are paramount in high-stakes domains like healthcare.
Concerningly, we uncover and investigate a \emph{privacy-hallucination tradeoff} in differentially private (DP) language models.
First, we empirically show that models pre-trained or fine-tuned with DP tend to produce more hallucinations than non-DP counterparts, with increased severity as the privacy budget grows stricter.
Second, we investigate model properties driving this tradeoff, demonstrating that DP mechanisms flatten output distributions, potentially redistributing probability mass toward factually incorrect alternatives. 
% Second, we investigate the mechanisms driving this tradeoff, demonstrating that DP mechanisms flatten output distributions in a manner that redistributes probability mass toward factually incorrect alternatives. 
Third, through experiments where we control fact frequency in training data, we characterize how information frequency can reduce hallucination risks in DP models. Overall, our findings underscore the need for more nuanced privacy-preserving interventions that offer rigorous privacy guarantees without compromising factual accuracy.

\end{abstract}

\section{Introduction}

The development and deployment of large language models (LLMs) in high-stakes settings requires non-negotiable standards for both privacy-preservation and factual accuracy. LLMs that are exposed to sensitive information during training are susceptible to reproducing it in subsequent interactions, resulting in privacy violations \citep{carlini-usenix-leakage,chu-etal-2024-reconstruct,meeus2024did,DBLP:conf/emnlp/KandpalPOKC024}, even in non-adversarial settings~\citep{DBLP:conf/iclr/AerniRDCIT25}; this is legally and ethically unacceptable in high-stakes domains such as healthcare and law, which involve sensitive data.
As simple anonymization offers insufficient protection \citep{staab24beyond, xin2024a, pang2024reconstructiondifferentiallyprivatetext}, differential privacy (DP) \cite{Dwork2006}  has emerged as the de-facto paradigm for provably mitigating privacy risks in language models \citep{DBLP:conf/uss/Carlini0EKS19,li2022large,xu-etal-2023-federated,Yan2024,hu-etal-2024-differentially}. 

Alongside privacy risks, LLMs are known to \emph{hallucinate}, i.e., to generate factually incorrect outputs \citep{wang-etal-2024-factuality, jiang-etal-2024-large, das-etal-2022-diving, framework-clinical-safety-hallucination}.\footnote{We use 
    ``hallucination'' to refer to LLM generation of false information. Unless stated otherwise, we restrict ourselves to information not supported by the training (pre-training or fine-tuning) data. \Cref{app:define_hallucination} gives a definition.
} 
This issue poses serious risks, particularly in high-stakes tasks such as generating patient discharge summaries  \citep{Chung2025Verifact}.
These concerns have motivated research aiming to evaluate and improve factual correctness in LLMs~\citep[e.g.][]{ji-etal-2023-towards,li-etal-2024-dawn}.
While prior work has investigated privacy and factual accuracy independently, no work has investigated the interaction between them, despite the clear need to achieve both in high-stakes settings.

In this work, we empirically investigate the trade-off between these two critical properties, specifically focusing on the guiding question: does differentially private training 
increase factual hallucinations
% decrease factual correctness 
in models? Our work is motivated by the potentially conflicting conditions conducive to each property.
Privacy-preserving strategies are inherently designed to counteract memorization \citep{miranda2025preservingprivacylargelanguage, kassem-etal-2023-preserving, hans2024goldfishloss}, which may inhibit the acquisition and, thus, the output of factual information
 \citep{lu-etal-2024-scaling,merullo2025on}.

In particular, because example-level DP limits the influence of individual training examples on the model, one might expect a privately trained model to struggle with reproducing \emph{rare facts} (i.e., those that appear only a few times in the fine-tuning data), while still capturing information that is repeated more frequently and is less likely to be privacy-sensitive. 
% \citep{miranda2025preservingprivacylargelanguage, kassem-etal-2023-preserving, hans2024goldfishloss}. 
However, our findings belie this expectation; rather than only reducing the generation of rare facts, private training leads to a problematic increase in hallucinations, where models do not simply omit what they failed to learn, but actively generate incorrect content instead.

In this work, we specifically investigate:

\begin{enumerate}[label=(RQ\arabic*), leftmargin=\widthof{(RQ11)}, topsep=0em,itemsep=0pt]
    \item What impact does DP training have on hallucinations in model outputs?
    \item What impact does DP training have on model properties related to hallucination?
    \item Under what conditions could DP be usable without increasing hallucinations?

\end{enumerate}

\mypar{Findings} 

\begin{itemize}[leftmargin=\widthof{()}, topsep=0em, itemsep=0pt]

    \item In RQ1, we find consistent evidence that DP training, in both fine-tuning and pre-training settings, leads to increased hallucinations, with a more pronounced degradation in the pre-training setting.
    \item In RQ2, we find that DP noise leads to  flatter predictive distributions, dispersing the next-token probability mass across a larger set of candidate tokens, which can increase the risk of generating factually incorrect content.
    \item Finally, in RQ3 we  show that a fact must recur several times before a DP-trained model acquires it at all, and we find that under stricter privacy budgets, facts are not acquired even at the substantially higher frequencies that we test for.
\end{itemize}

Our results call for increased caution in turning to DP as a solution to privacy in high-stakes settings and further investigation of when DP can be used without increasing hallucinations.\footnote{Our code is available at: \href{https://github.com/kr-ramesh/privacy-hallucination-tradeoff}{https://github.com/kr-ramesh/privacy-hallucination-tradeoff}.}

\section{Experimental Design}
\label{sec:expt-design}

Our goal is to measure how models are affected by DP at various privacy budgets. To this end, we carefully construct experimental setups where we can fine-tune valid differentially private models, control for overlap between fine-tuning and pre-training data, and evaluate factual accuracy in open-ended model outputs. %
We situate our study in related research on privacy-preserving approaches and factuality in language models, reviewed in detail in \Sref{app:related_work}.

\subsection{Models and Training Setup}

\mypar{DP Fine-tuning}
We fine-tune LLMs for controllable text generation, similar to previously proposed applications for DP in LLMs, particularly for privacy-preserving synthetic and open-ended text generation \citep{yue-etal-2023-synthetic, mattern-etal-2022-differentially,ramesh-etal-2024-evaluating}.

All fine-tuning (DP and non-DP) is achieved  using Low-Rank Adaptation (LoRA) for computational efficiency \citep{Hu2022}. 
We specify additional details about DP in \Sref{app:dp_background} and about LoRA in \Sref{app:exp_setup}, including all hyperparameter settings.

The base language model behind all fine-tuning experiments is GPT-J 6B \citep{gpt-j}, a decoder-only transformer-based model. We make this choice for a crucial reason: its pre-training dataset, namely The Pile, is fully open with a known cutoff date, ensuring that GPT-J is not pre-trained on any post-2020 data. This knowledge of the pre-training data mixture and cut-off date allows us to select fine-tuning data that has no overlap with pre-training data; cf. \Sref{sec:methods_datasets}.

\mypar{Private pre-training}
We address the effect of DP pre-training (as opposed to fine-tuning) on factuality using the DP-pre-trained VaultGemma  \citep{vaultgemma}.
This is a 1B-parameter open-weights model fully pre-trained with DP ($\epsilon = 2$).

We compare VaultGemma with the same two models as in \citet{vaultgemma}: 
Gemma3-1B, which can be viewed as the ``non-private counterpart'', and GPT-2 XL (1.5B), which has performance similar to VaultGemma's on standard benchmarks, although it has a much earlier knowledge cut-off date, as it was released in 2019. We also evaluate the non-private Gemma2-2B, which is data-matched with VaultGemma, and Gemma-2B as additional baselines.

\subsection{Datasets}
\label{sec:methods_datasets}

We use Wikipedia data for fine-tuning and evaluation because it meets two key criteria. First, meta-data allows us to select articles created after 2020, ensuring they were not included in GPT-J 6B's pre-training corpora. Overlap with pre-training data would invalidate DP guarantees and artificially inflate performance  \cite{DBLP:conf/icml/Tramer0C24,cummings2023advancing,igamberdiev-etal-2022-dp}. Second, content is constructed to contain verifiable facts rather than opinions or speculation, and automated fact-checking methods have been previously validated over Wikipedia \citep{min-etal-2023-factscore}, which ensures that the setup can answer RQ1.
We use three datasets for factuality evaluation as summarized in \Cref{tab:datasets} and described below. An exact list of topics included in these datasets is given in Appendix~\ref{app:list_of_topics}.

\mypar{Wikipedia Science}
We collect 231 Wikipedia articles on science topics that were created after the cutoff date for GPT-J 6B's pre-training data, where we use keyword searches of Wikipedia meta-data to identify science articles. We focus on science topics as they contain  detailed technical language, which is also common in sensitive data settings (e.g., clinical notes).
While Wikipedia articles on these topics did not exist before 2020, we do expect some concepts to exist in other pre-training data sources, which makes it feasible for a DP model to produce facts on these topics, even without memorizing individual data points.

\mypar{Wikipedia AI}
We collect 124 Wikipedia articles on AI topics, where we hand-curate products and models that did not exist before 2020, along with related articles we expect to mention them. Unlike the Wikipedia Science articles, GPT-J 6B cannot have any knowledge of most of these concepts without fine-tuning, as they could not have existed in pre-training data. However, by constructing our data to contain articles that mention overlapping topics, we ensure that it is feasible for a DP model to learn them. For example, if our dataset only contained \textit{DeepSeek (chatbot)}, DP would preclude learning of information isolated to one data point. By including  \textit{DeepSeek (chatbot)}, \textit{DeepSeek}, and \textit{DeepSeek (disambiguation)}, a DP model can hypothetically learn information about DeepSeek, as it is mentioned in multiple data points.

\mypar{Wikipedia pre-training}
To investigate effects of DP fine-tuning on knowledge acquired during pre-training, we randomly sample 250 Wikipedia articles from the GPT-J pretraining data that are not included in the fine-tuning data.

For fine-tuning, since DP requires sufficiently large datasets and batch sizes \cite{DBLP:conf/iclr/McMahanRT018,DBLP:journals/jair/PonomarevaHKXDMVCT23}, we intersperse our curated evaluation articles with 20,000 additional randomly sampled Wikipedia articles (ensuring no overlap with the pre-training evaluation set). The data is divided into 512-token sequences as the unit of privacy protection. We fine-tune the model to produce an article when prompted on the article title (e.g., a topic).

\begin{table}[t]
    \centering
    \scriptsize
    \setlength{\tabcolsep}{2.5pt}
    \begin{tabular}{l c p{2.6cm}}
        \toprule
        \textbf{Dataset} & \textbf{Article Count} & \textbf{Topics} \\
        \midrule
        Wikipedia Science  & 231 & Follicular drug delivery; Malaria therapy; Eurotrac \\
        Wikipedia AI  & 124 & DeepSeek; DALL-E \\
        Wikipedia Pretraining  & 250 & Randomly sampled articles \\
        \midrule
        Fine-tuning data & 20355 &  Wikipedia Science + Wikipedia AI + 20k randomly sampled articles \\
        \bottomrule
    \end{tabular}
    \caption{Datasets used for evaluation and fine-tuning.
    %\dd{unclear what size here means, do you mean number of articles? If yes, better to be clear.}
    }
    \vspace{-0.2cm}
    \label{tab:datasets}
\end{table}
 
%\sectionwrite{The choice of models and the training parameters. Add table in the appendix.}

\subsection{Evaluation of Factual Accuracy in Open-Ended Text Generation}
\label{sec:methods_fact_eval}

\begin{figure}[htbp]
    \centering
    \begin{subfigure}{0.48\columnwidth}
        \centering
        \includegraphics[width=\linewidth]{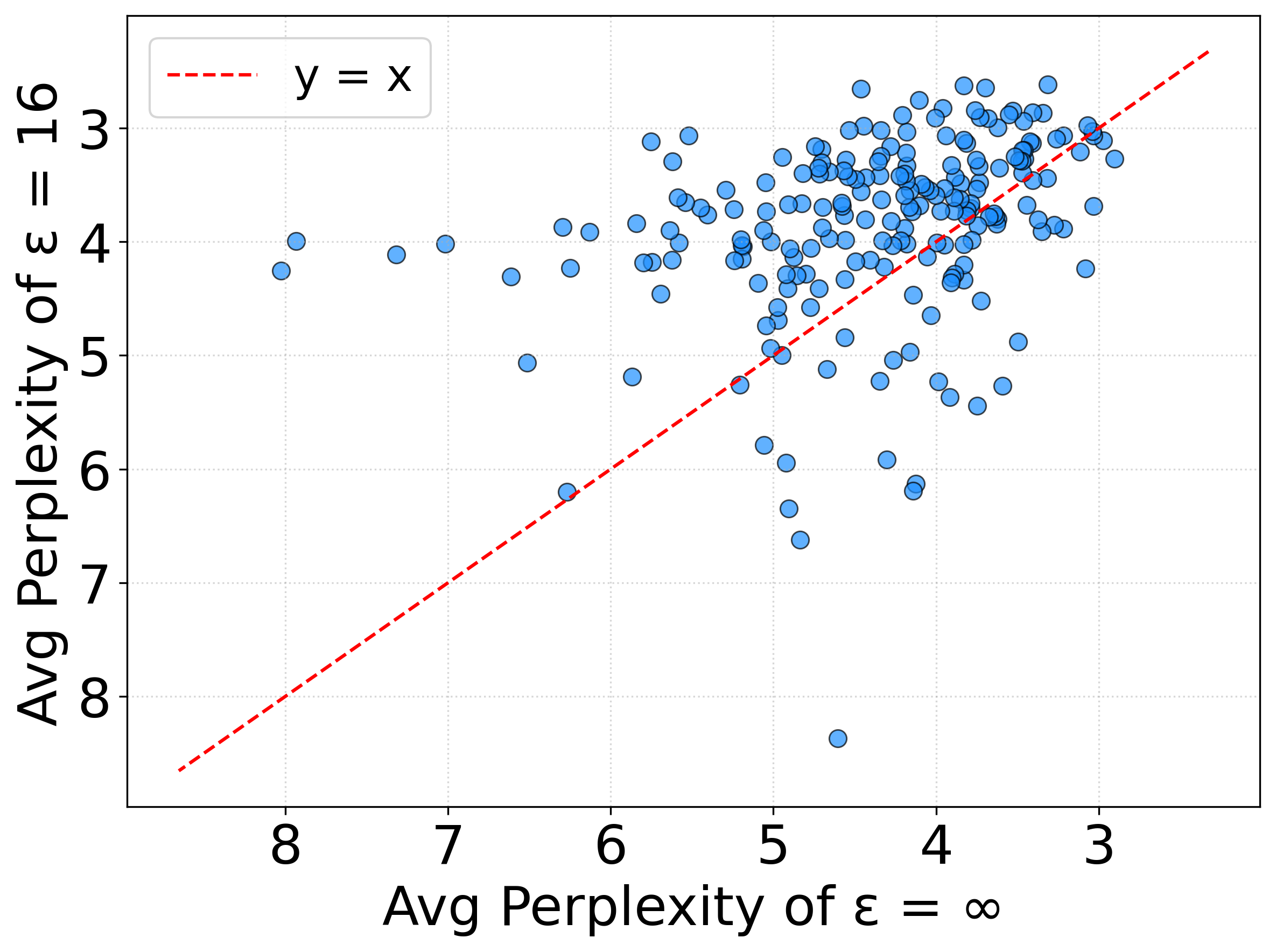}
        \caption{\footnotesize Avg.\ perplexity}
    \end{subfigure}
    \hfill
    \begin{subfigure}{0.48\columnwidth}
        \centering
        \includegraphics[width=\linewidth]{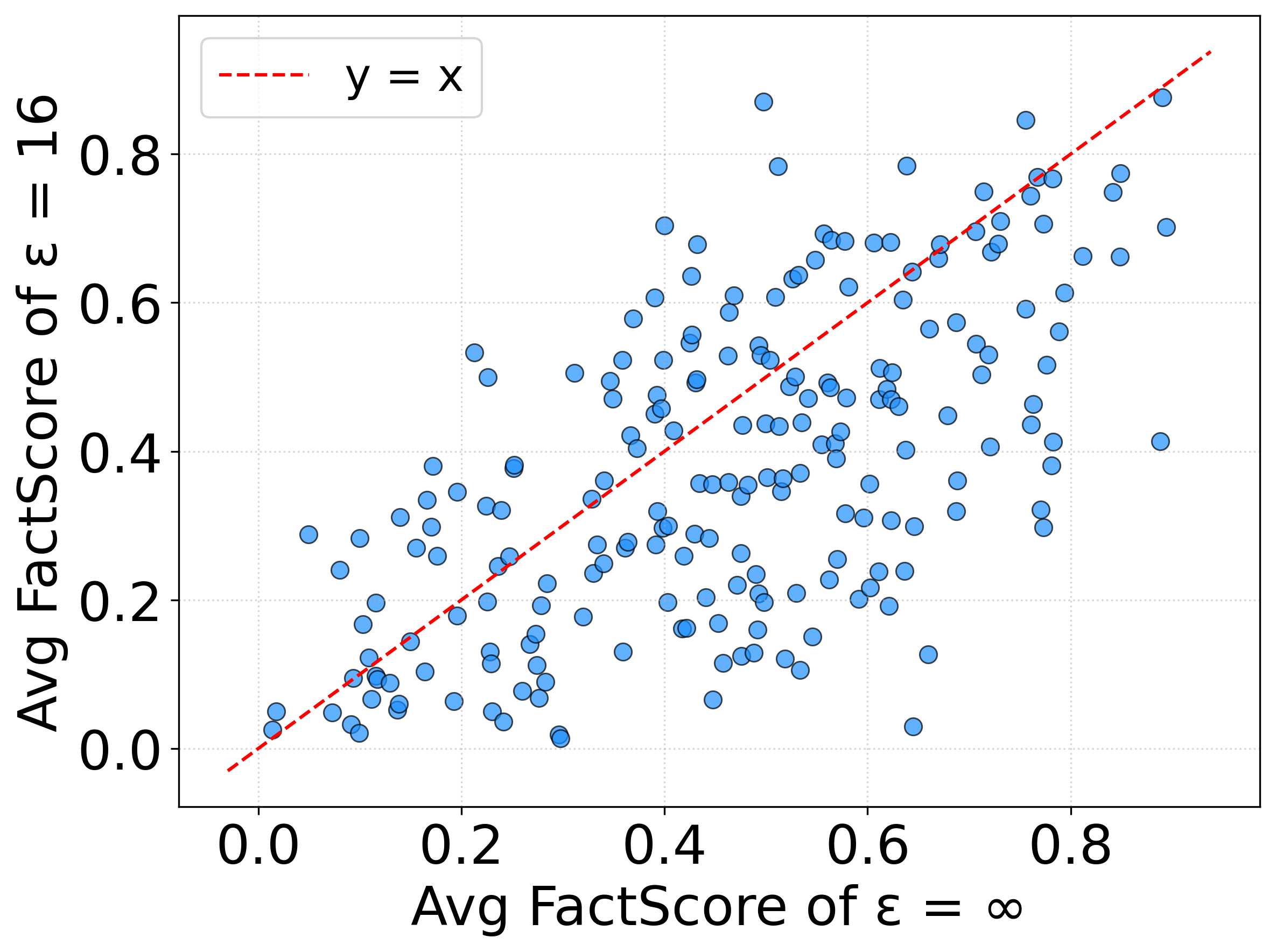}
        \caption{\footnotesize Avg.\ FactScore}
    \end{subfigure}
    \caption{Average perplexity (a) and average FactScore (b) per topic on Wikipedia Science, $\varepsilon = 16$ vs.\ $\varepsilon = \infty$. DP-finetuned models often achieve lower average perplexity  (points above the $y = x$ line), yet the non-DP model attains higher average FactScores across topics—showing that lower perplexity does not imply better factuality.}
    \label{fig:main_factscore_and_perplexity_graphs}
    \vspace{-0.4cm}
\end{figure}

We evaluate the factual accuracy in open-ended text generation settings where we prompt the model with a title and generate a full Wikipedia article. 
We then measure factual accuracy of the generated article, where higher accuracy indicates a lower hallucination rate. 
This setup reflects a more modern paradigm for evaluating factuality in open-ended generation, as opposed to traditional factuality evaluations based on cloze-style or short-form response queries designed to probe models \citep{youssef-etal-2023-give, petroni-etal-2019-language}. 
We note that utility measures such as perplexity---a common metric used for evaluating the efficacy of DP methods in prior work \cite{yu2022differentially, thareja2026dpfusion, ma-rajtmajer-2026-private}---do not serve as a proxy for hallucinations, and therefore fail to capture factual inaccuracies (see \Cref{fig:main_factscore_and_perplexity_graphs}; further details provided in \Cref{app:hallucinations_vs_perplexity}). We use both automated and human assessments of factuality, as described below.

\mypar{Automated Evaluation via FactScore} 
% We use FactScore \citep{min-etal-2023-factscore} for automated evaluation of factuality. 
Given a generated text $d_i$, FactScore \citep{min-etal-2023-factscore} operates in two distinct phases: (i) \textbf{atomic claim extraction}, where $d_i$ is decomposed into a set of minimal, verifiable claims $\mathcal{AF}_{d_i}$, and (ii) \textbf{claim verification}, where each claim $\alpha^{(d_i)}_j \in \mathcal{AF}_{d_i}$ is evaluated for factuality using a verifier $\mathcal{V}$ conditioned on both intrinsic language model judgments and evidence retrieved from an external knowledge source $\mathcal{K}$, as detailed in \Cref{alg:factscore}, Appendix~\ref{app:factscore}.
We use the original Wikipedia article as the external knowledge source for verifying generated claims.
As claim decomposition methods can produce redundant claims that artificially inflate scores, we use the CORE module \citep{jiang2024corerobustfactualprecision} to filter down the superfluous and repetitive claims.
We use Llama-3.1-8B-Instruct
% \footnote{https://huggingface.co/meta-llama/Llama-3.1-8B} 
for both (i) and (ii). 
% In , 
We verify in Appendix~\ref{app:additional_exp}  that trends are consistent with other choices of models.

\mypar{Human Evaluation}
As FactScore is an automated metric relying on LLM judgments that may not be accurate, we conduct human evaluations to validate results and provide finer-grained analysis of generated information. We use the Wikipedia AI dataset, as this dataset most carefully separates the pre-training and fine-tuning data. For the annotation task, we recruited 5 computer science graduate students, whom we expect to have high AI literacy. We ensure that the human evaluators see generations with both high and low factual accuracy using stratified sampling: we sample 15 articles where both DP ($\epsilon=16$) and non-DP models have FactScore $\ge50\%$ and 15 where both models have FactScore $<50\%$. Since FactScore judgements may be imprecise, this stratification allows us to evaluate whether human annotators can distinguish between DP and non-DP generations with comparable FactScores, and informs us of whether FactScore underestimates DP's effect on the factuality of the generation. 
Annotators rated automatically decomposed atomic claims for their overall veracity (correct/incorrect/unclear) and their support within the source text using our custom annotation interface (Appendix \Sref{app:claim-annotations}). Annotators were also asked to flag quality issues in decomposed claims, marking them as vague or subjective. Two annotators rated each claim's veracity and support, and we report moderate-to-strong inter-annotator agreement (Cohen's $\kappa = 0.57$ for both veracity and support at
$\epsilon = \infty$, and $0.84$ and $0.73$ respectively at $\epsilon = 16$). \footnote{Annotators were offered compensation of $\sim\$20$/hr.}

\begin{table}[t]
\small
\setlength{\tabcolsep}{4pt}
\resizebox{\columnwidth}{!}{
\begin{tabular}{@{}lrrrrc@{}}
\toprule
& \multicolumn{4}{c}{\textbf{Response-level}} & \multicolumn{1}{c}{\textbf{Topic-level}} \\
\cmidrule(lr){2-5} \cmidrule(l){6-6}
\textbf{DP Setting} & \textbf{Avg FS} & \textbf{Med} & \textbf{Q1} & \textbf{Q3} & \textbf{Avg FS [95\% CI]} \\
\midrule
\multicolumn{6}{@{}l}{\textit{Wiki AI}} \\
{\footnotesize $\epsilon=\infty$}  & 37.7          & 33.3          & 13.3         & 57.1          & 37.5 {\scriptsize[33.2, 42.0]} \\
{\footnotesize $\epsilon=16$}      & 33.8          & 27.3          & 10.0         & 54.0          & 33.6 {\scriptsize[29.2, 38.2]} \\
{\footnotesize $\epsilon=8$}       & 31.6          & 25.0          & 10.0         & 50.0          & 31.4 {\scriptsize[27.3, 35.7]} \\
{\footnotesize Base}               & 32.9          & 27.3          & \textbf{9.1} & 50.0          & 32.9 {\scriptsize[28.4, 37.5]} \\
{\footnotesize Task-tuned}         & \textbf{30.3} & \textbf{23.1} & \textbf{9.1} & \textbf{48.3} & \textbf{30.3} {\scriptsize[26.3, 34.4]} \\
\midrule
\multicolumn{6}{@{}l}{\textit{Wiki Science}} \\
{\footnotesize $\epsilon=\infty$}  & 56.1          & 58.3          & 33.3          & 80.0          & 56.1 {\scriptsize[52.6, 59.5]} \\
{\footnotesize $\epsilon=16$}      & 53.6          & 55.6          & 28.6          & 80.0          & 53.6 {\scriptsize[50.1, 57.1]} \\
{\footnotesize $\epsilon=8$}       & 53.5          & 54.5          & 28.6          & 80.0          & 53.6 {\scriptsize[50.1, 57.1]} \\
{\footnotesize Base}               & \textbf{52.1} & \textbf{50.0} & 27.3          & \textbf{75.0} & \textbf{51.9} {\scriptsize[48.9, 55.0]} \\
{\footnotesize Task-tuned}         & 52.6          & \textbf{50.0} & \textbf{25.0} & 80.0          & 52.5 {\scriptsize[49.0, 56.1]} \\
\bottomrule
\end{tabular}}
\caption{
FactScore (FS; \%) of GPT-J 6B fine-tuned under different DP budgets, evaluated at temperature $\tau = 0.3$. Response-level statistics pool all generations; topic-level averages within each entity first, with bootstrap 95\% CIs resampled over topics. The row with the lowest FactScore (i.e.\ most hallucinations) is bolded. Factual consistency decreases with stricter privacy budgets (smaller $\epsilon$), highlighting the tradeoff between privacy and hallucinations. Significance tests for these differences against the non-private baseline ($\varepsilon=\infty$) in ~\Sref{app:statistical_significance_testing}.}
\label{tab:factscore_unseen}
\vspace{-0.35cm}
\end{table}

%``Base'' refers to the base GPT-J 6B model with no fine-tuning and ``Task-tuned'' denotes non-private fine-tuning on pre-training data only (for 1 epoch).

\begin{figure}[t]
    \centering
    \begin{subfigure}{0.4\textwidth}
        \centering
        \includegraphics[width=\linewidth]{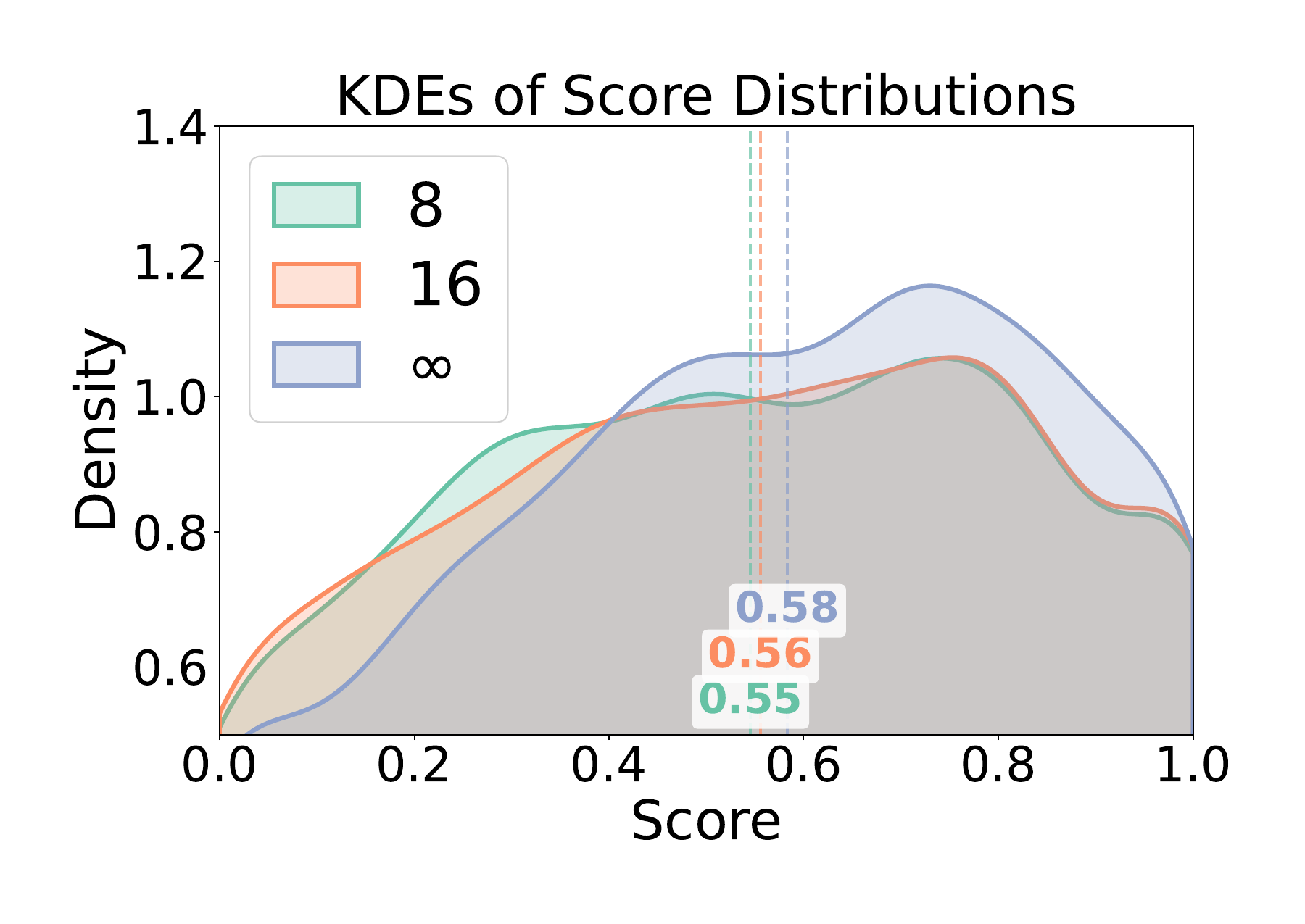}
        \caption{Wikipedia Science}
    \end{subfigure}
    \hfill
    \begin{subfigure}{0.4\textwidth}
        \centering
        \includegraphics[width=\linewidth]{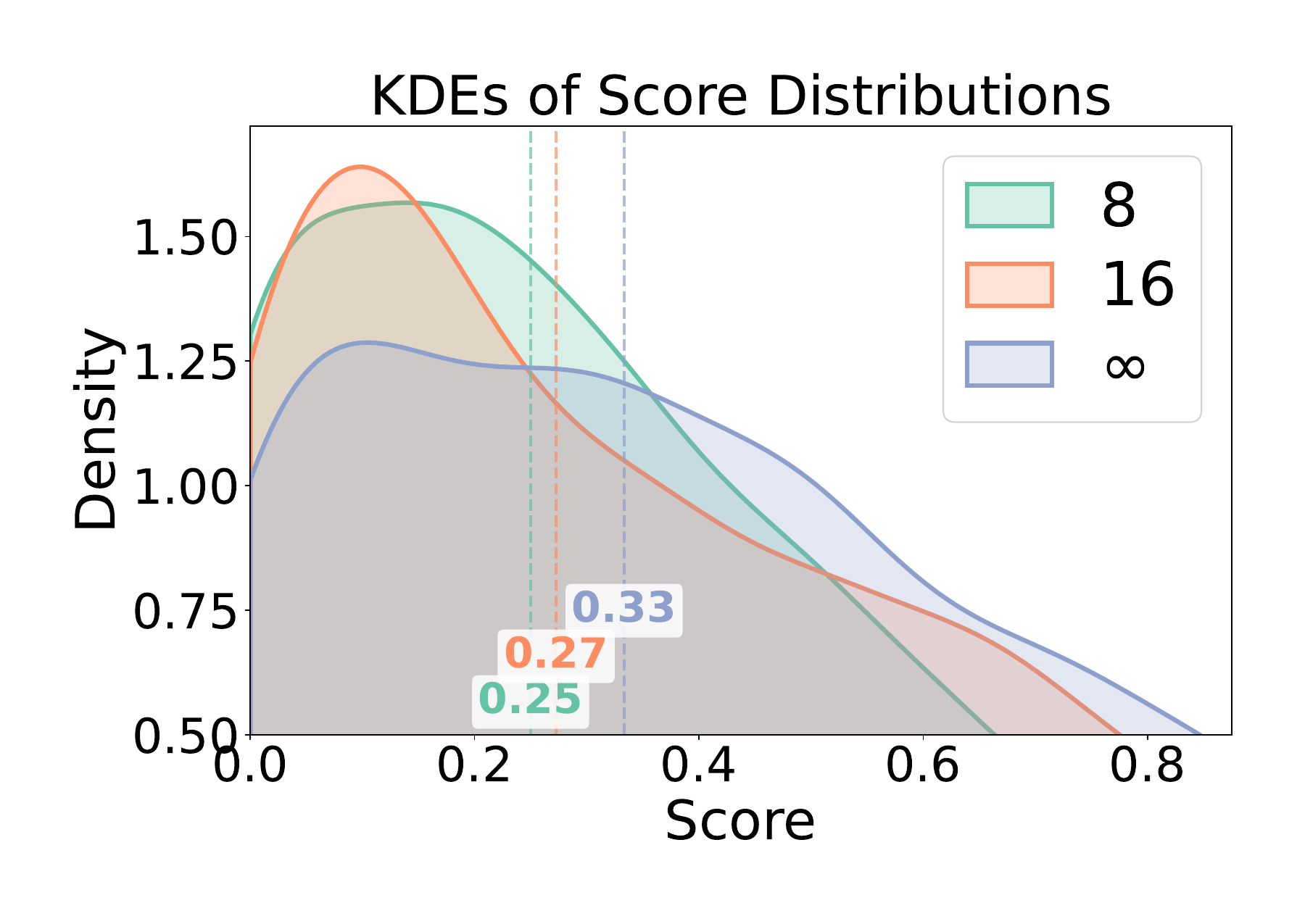}
        \caption{Wikipedia AI}
    \end{subfigure} \\
    \caption{Kernel density (KDE) plots comparing FactScore distributions when finetuning GPT-J 6B under different DP budgets. As the privacy budget becomes more stringent ($\epsilon$ from $\infty$  to $8$), the FactScore distribution shifts leftward, indicating that stronger privacy constraints result in less factually consistent outcomes.
    %\dd{The font sizes on the legends need to be bigger. I had left this note previously as well, please don't comment them out without addressing them or communicating about those.} \kr{Had to comment it out for the ARR submission. I didn't have the time to dig up the code and rerun the graphs.}
    }
    \label{fig:factscore_distributions}
\end{figure}

\begin{figure*}[t]
    \centering
    % --- Left: Image ---
    \begin{subfigure}{0.45\textwidth}
        \centering
        \includegraphics[width=\linewidth]{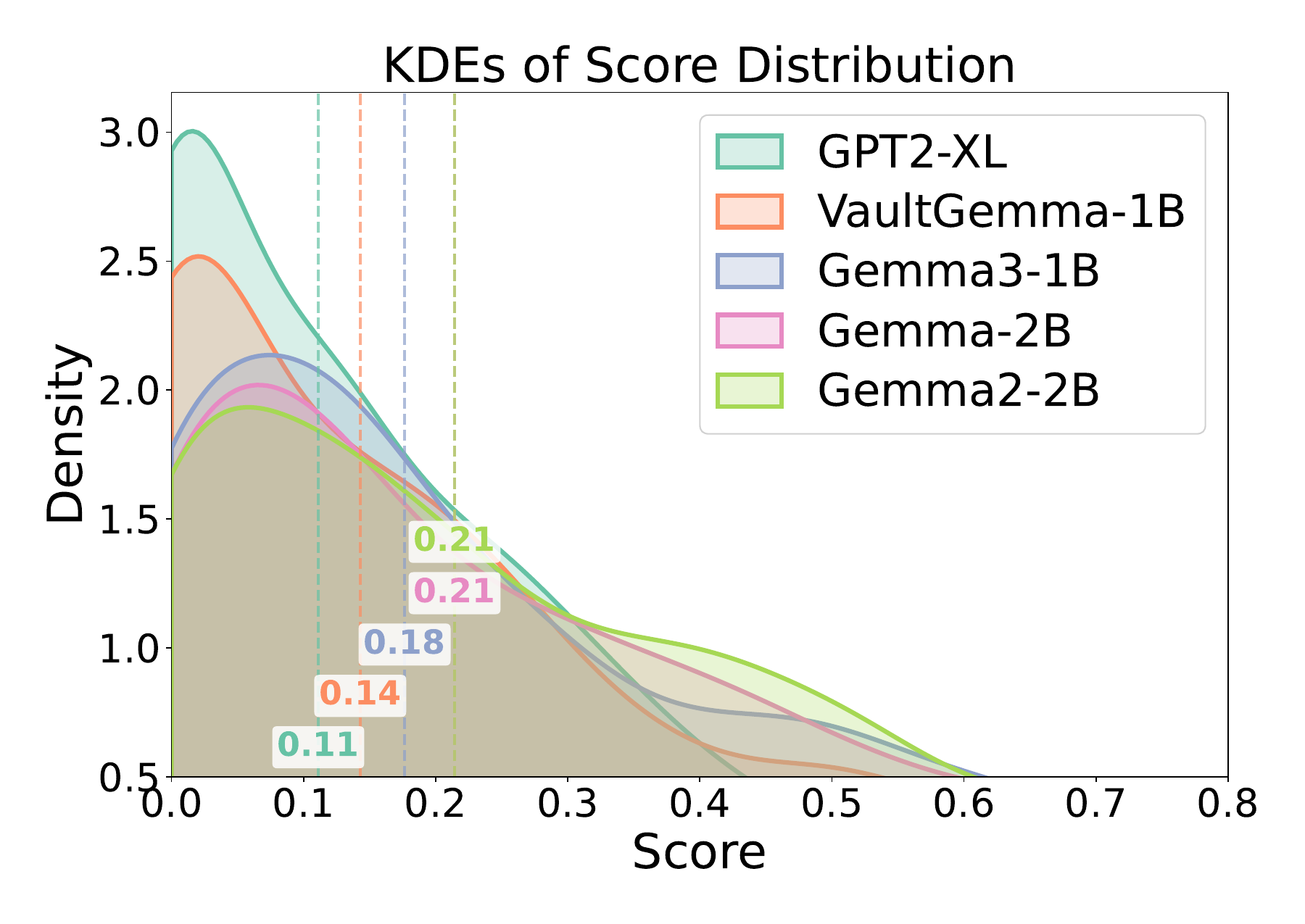}
        \caption{Wikipedia Pretrained}
    \end{subfigure}
    \begin{subfigure}{0.45\textwidth}
        \centering
        \includegraphics[width=\linewidth]{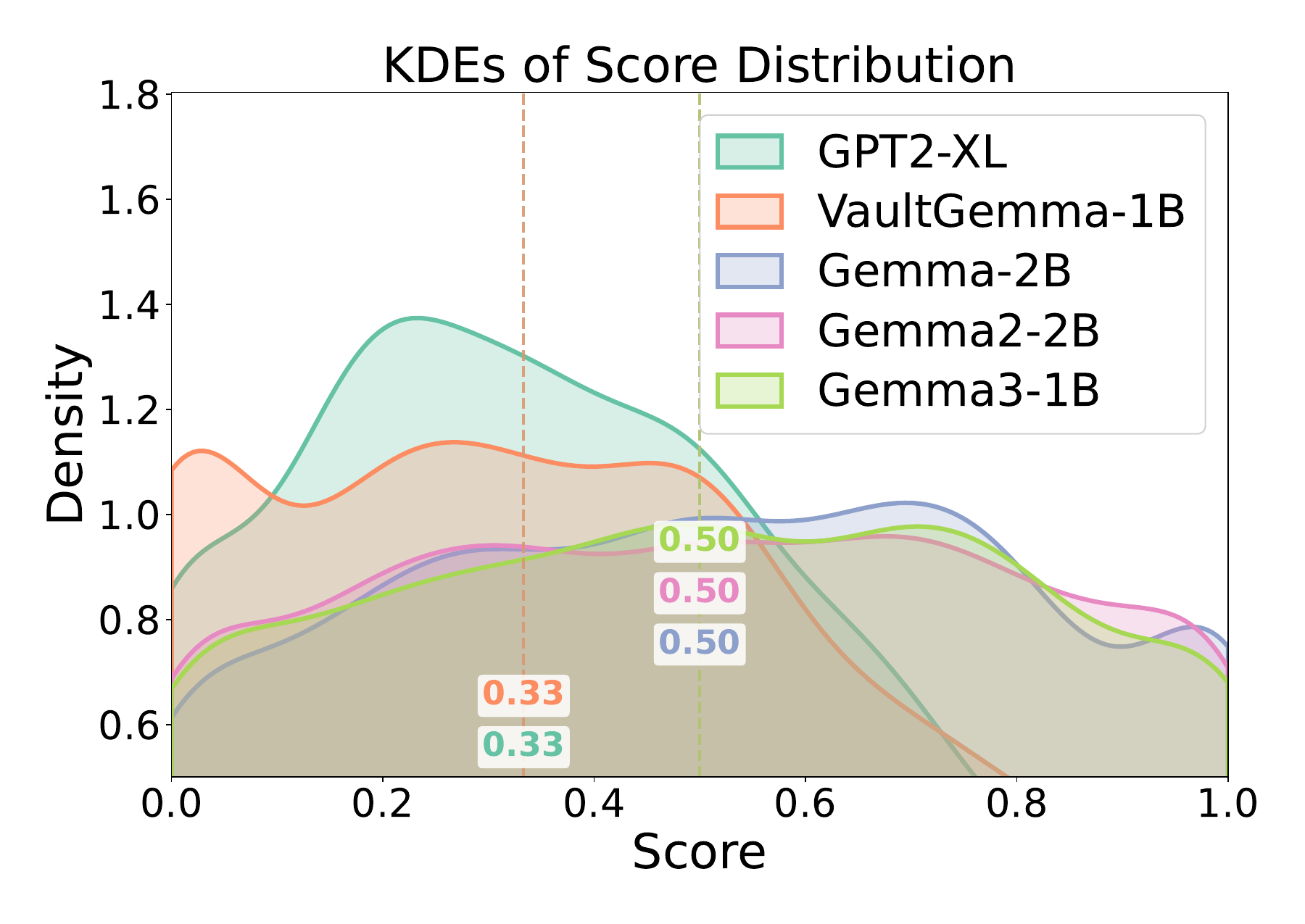}
        \caption{Wikipedia Science}
    \end{subfigure}
    \caption{\small 
    Kernel density (KDE) plots comparing FactScore distributions for models pre-trained with (VaultGemma $\epsilon = 2$) and without DP (Gemma models \& GPT2-XL). GPT2-XL, which has not seen the data during training, and VaultGemma (DP pre-trained) both lag behind the Gemma models in factual consistency.}
    \label{fig:kde_pre-train}
    \vspace{-0.3cm}
\end{figure*}

\begin{table}[t]
\centering
\renewcommand{\arraystretch}{0.95}
\setlength{\tabcolsep}{3pt}
\footnotesize
\begin{tabular}{@{}lcccc@{}}
\toprule
\multirow{2}{*}{\textbf{Wikipedia AI}} & 
\multicolumn{2}{c}{\textbf{FactScore $<$ 0.5}} & 
\multicolumn{2}{c}{\textbf{FactScore $\ge$ 0.5}} \\ 
\cmidrule(lr){2-3} \cmidrule(lr){4-5}
 & $\epsilon=\infty$ & $\epsilon=16$ & $\epsilon=\infty$ & $\epsilon=16$ \\ 
\midrule
Avg. Veracity Score  & 0.18  & 0.17  & 0.51  & 0.43  \\
Avg. Support Ratio   & 0.19  & 0.09  & 0.54  & 0.32  \\
Quality Issues      & 87    & 79    & 55.5  & 79    \\
\bottomrule
\end{tabular}
\caption{
Human annotation results on Wiki AI , stratified by response groups with FactScore  $<0.5$ and $\ge 0.5$. 
In both cases, hallucinations increase with DP ($\epsilon=16$) relative to no DP ($\epsilon=\infty$).
``Quality Issues'' is the number of claims annotated as having quality issues.
}
\vspace{-0.5cm}
\label{tab:wiki-ai-veracity}
\end{table}

\section{RQ1: What impact does DP training have on hallucinations in model outputs?}

\subsection{DP fine-tuning increases hallucinations}\label{sec:hallucinations_in_dp_finetuning}

\mypar{Automated Evaluation} We generate multiple articles per topic and report both the response-level statistics, where every generation is scored as an independent observation, and topic-level averages, where FactScores for generations are averaged within a topic first, so every topic is weighted equally, and the CIs are bootstrapped by resampling topics.
In addition to fine-tuned models, we also evaluate the ``base'' pre-trained model and the task-tuned model, which denotes the base model fine-tuned without DP on pre-training Wikipedia data only, excluding the unseen \textsc{Wiki\textsubscript{AI + Science}} articles. These baselines offer comparison between the DP models and models never trained on this data.

The automated factuality evaluation for unseen fine-tuning data in \Cref{tab:factscore_unseen} suggests that DP fine-tuning leads to more frequent hallucinations. 
There is a  consistent decrease in the average FactScore for the text generated from models fine-tuned with DP compared to those trained without DP: 
non-DP ($\epsilon=\infty$) $\succ$ DP ($\epsilon=16$) $\succ$  DP ($\epsilon = 8$) on both datasets. 
Performance decline is more pronounced for the stricter privacy budget setting $\varepsilon = 8$ on the Wikipedia AI data, and is further illustrated by the skew (toward lower factuality scores) in the distributions of FactScores for DP-trained models in \Cref{fig:factscore_distributions}. The differences in the FactScores of the non-DP and DP-finetuned models are significant at the response level and at the topic-level, except for $\varepsilon=16$ on Wikipedia AI. (cf.~\Sref{app:statistical_significance_testing}). 

As a sanity check, 
we observe that the model trained without DP outputs a higher percentage of correct claims than the base model  (32.9\% $\rightarrow$ 37.7\% for the AI articles and 52.1\% $\rightarrow$ 56.1\% for the Science articles), demonstrating successful fine-tuning.
Notably, on the Wikipedia AI dataset, the model fine-tuned with $\varepsilon = 8$ performs worse or comparably (mean: 31.6\%; median: 25.0\%), to the base model (mean: 32.9\%; median: 27.3\%), despite the training loss decreasing progressively when fine-tuning with this privacy budget (\Cref{app:training_loss_curves}).  While this marginal degradation in FactScore for Wikipedia AI articles could be statistical noise, it could also indicate the gradient signal from the AI articles being masked by the larger levels of noise introduced by DP-SGD when fine-tuning with lower privacy budgets ($\varepsilon = 8$). Under a more generous budget ($\varepsilon = 16$), FactScore does increase compared to the baseline (mean: 33.8\%), but still falls short of the model trained without DP (mean: 37.7\%).

FactScore is higher for the Science articles compared to AI articles; median scores are consistently $\ge$ 50\%. Furthermore, the  DP model with $\varepsilon = 8$  achieves a higher mean FactScore than the base model for the Science articles, suggesting that DP fine-tuning hallucinates less than the base model. The differences between the AI and Science datasets likely results from their differing overlap with pre-training data: concepts from Science articles are more likely to occur in pre-training data, even if exact articles are non-overlapping. While these experiments evaluate factuality on articles not seen during pre-training, in~\Cref{app:hallucination_pretraining_facts_dp}, we also test whether DP fine-tuning perturbs knowledge already encoded in the base model from pre-training.
% On evaluating the factuality of these models on articles from the pre-training corpus,
We find insignificant differences between the DP and non-DP models, which suggests that the noise injected during DP fine-tuning does not disrupt knowledge acquired in non-private pre-training.

\mypar{Human Evaluation} \Cref{tab:wiki-ai-veracity} reports the average ratings selected by human annotators. In lower-quality generations (FactScore $\le 0.5$), annotators rated both the non-DP and DP models with equally low veracity, while the DP model exhibited more unsupported facts. In higher-quality generations (FactScore $> 0.5$), annotators rated the DP model as outputting both lower veracity information and more unsupported facts.
There are not conclusive differences in counts of quality issues.
While agreement between human annotators was generally high, model-human agreement was not always high , and the model--annotator agreement on support was substantially lower ($\kappa = 0.23$--$0.49$; see Table~\ref{tab: human_claim_annotation_agreement}), indicating that automated evaluations of factuality are an imperfect proxy for human judgment of how factual a response is.
Regardless, overall trends are consistent between human and automated evaluations:  both indicate greater hallucination in the DP model, even the model with a more generous privacy budget.

\mypar{Additional diagnostics} We conduct further diagnostics to analyze the behavior of DP models, and report our findings in the appendix. First, we find that DP fine-tuning does not appear to disrupt knowledge already acquired during pre-training (\Sref{app:hallucination_pretraining_facts_dp}). Second, a natural concern is that the DP fine-tuned models revert to their pre-trained priors during generation, rather than acquiring knowledge from unseen fine-tuning data; our analyses in~\Sref{app:reverting_to_priors} demonstrates that this is not the case. Third, our experiments in~\Sref{sec:recurrent-claim} show that DP models tend to output the same repeated hallucinations, suggesting that there may be structured shifts in the  output distribution that increase the likelihood of certain incorrect alternatives. Finally, in addition to standard DP-SGD, we evaluate the adaptive noise allocation algorithm proposed in \citep{li-etal-2024-fine} and find that it does not help mitigate the privacy-hallucination tradeoff; we defer the results and the discussion of its privacy analysis to~\Sref{app:anadp_description}.

\subsection{DP pre-training and hallucinations}
% Krishna: we not have a strong enough causal link here, so I softened the language

We investigate the effect of DP pre-training, as opposed to fine-tuning, on hallucinations in \Cref{tab:factscore-pre-trained}. It is worth noting that the pre-training data mixtures for the Gemma models likely include \textit{all} the datasets used in our factual evaluations, whereas GPT-2 is not expected to have been trained on any of our factual evaluation datasets.

We find that the DP pre-trained model (VaultGemma) consistently outputs a greater percentage of inaccurate facts relative to every non-private Gemma baseline. On the Wikipedia pre-training data, Gemma3, the lowest-scoring Gemma baseline for this data, achieves an average FactScore of 26.6, compared to 22.0 for VaultGemma. Gemma-2B and Gemma2-2B score higher still (with average FactScores of 29.3 and 28.7), despite Gemma2-2B sharing VaultGemma's pre-training mixture. This gap in factual correctness is more pronounced for the domain-specific scientific and AI articles, where the difference between Gemma3-1B and VaultGemma reaches 10.9 points (50.4 vs. 39.5) and 8.7 points (51.9 vs. 43.2), respectively, with the other Gemma models ahead of VaultGemma by comparable margins. All the differences between the VaultGemma and Gemma models are statistically significant (\Sref{app:statistical_significance_testing}), and the FactScore distributions reflect this as well, (\Cref{fig:kde_pre-train}) where VaultGemma outputs have a higher density of low FactScores than any of the Gemma models.

\begin{table}[H]
\renewcommand{\arraystretch}{1.0}
\centering
\small
\setlength{\tabcolsep}{4pt}
\resizebox{\columnwidth}{!}{%
\begin{tabular}{@{}lrrrrc@{}}
\toprule
& \multicolumn{4}{c}{\textbf{Response-level}} & \multicolumn{1}{c}{\textbf{Topic-level}} \\
\cmidrule(lr){2-5} \cmidrule(l){6-6}
\textbf{Model} & \textbf{Avg FS} & \textbf{Med} & \textbf{Q1} & \textbf{Q3} & \textbf{Avg FS [95\% CI]} \\
\midrule
\multicolumn{6}{@{}l}{\textit{Wikipedia AI}} \\
{\footnotesize Gemma3}     & 51.9          & 55.6          & 25.0         & 79.6          & 51.6 {\scriptsize[46.4, 56.7]} \\
{\footnotesize VaultGemma} & 43.2          & 43.3          & 17.0         & 66.7          & 43.1 {\scriptsize[38.5, 47.7]} \\
{\footnotesize GPT-2 XL}   & \textbf{27.7} & \textbf{22.2} & \textbf{7.6} & \textbf{45.5} & \textbf{27.7} {\scriptsize[24.3, 31.5]} \\
{\footnotesize Gemma2}     & 49.4          & 50.0          & 21.4         & 77.8          & 49.4 {\scriptsize[44.2, 54.5]} \\
{\footnotesize Gemma}      & 49.6          & 50.0          & 23.1         & 77.1          & 49.4 {\scriptsize[44.3, 54.5]} \\
\midrule
\multicolumn{6}{@{}l}{\textit{Wikipedia Science}} \\
{\footnotesize Gemma3}     & 50.4          & 50.0          & 25.0          & 75.0          & 50.6 {\scriptsize[47.4, 53.7]} \\
{\footnotesize VaultGemma} & 39.5          & \textbf{33.3} & \textbf{14.6} & 60.0          & \textbf{39.1} {\scriptsize[36.4, 41.9]} \\
{\footnotesize GPT-2 XL}   & \textbf{39.1} & \textbf{33.3} & 16.9          & \textbf{57.1} & 39.3 {\scriptsize[36.7, 42.0]} \\
{\footnotesize Gemma2}     & 50.3          & 50.0          & 25.0          & 75.0          & 50.4 {\scriptsize[47.4, 53.5]} \\
{\footnotesize Gemma}      & 51.4          & 50.0          & 25.0          & 76.9          & 51.5 {\scriptsize[48.3, 54.8]} \\
\midrule
\multicolumn{6}{@{}l}{\textit{Wikipedia Pretrain}} \\
{\footnotesize Gemma3}     & 26.6          & 17.6          & 7.1          & 42.9          & 26.7 {\scriptsize[23.9, 29.6]} \\
{\footnotesize VaultGemma} & 22.0          & 14.3          & \textbf{0.0} & 33.3          & 22.0 {\scriptsize[19.6, 24.7]} \\
{\footnotesize GPT-2 XL}   & \textbf{17.6} & \textbf{11.1} & \textbf{0.0} & \textbf{25.0} & \textbf{18.0} {\scriptsize[15.9, 20.2]} \\
{\footnotesize Gemma2}     & 28.7          & 21.4          & 7.7          & 44.4          & 28.7 {\scriptsize[25.9, 31.6]} \\
{\footnotesize Gemma}      & 29.3          & 21.4          & 7.1          & 45.5          & 29.3 {\scriptsize[26.2, 32.7]} \\
\bottomrule
\end{tabular}}
\caption{
FactScore (FS; \%) of pre-trained models at $\tau = 0.3$. Response-level statistics pool all generations; topic-level averages within each entity first, with bootstrap 95\% CIs resampled over topics. Bolding indicates lower FactScore (increased hallucinations).\footnotesize{*Gemma3 = Gemma3-1B-PT, VaultGemma = VaultGemma-1B (DP, $\epsilon\!=\!2$), GPT-2 XL (1.5B), Gemma2 = Gemma-2-2B, Gemma = Gemma-2B. Significance tests for these differences against VaultGemma are reported in~\Sref{app:statistical_significance_testing}}}.
\label{tab:factscore-pre-trained}
\end{table}

\begin{figure*}[t]
        \centering
        \includegraphics[width=\linewidth]{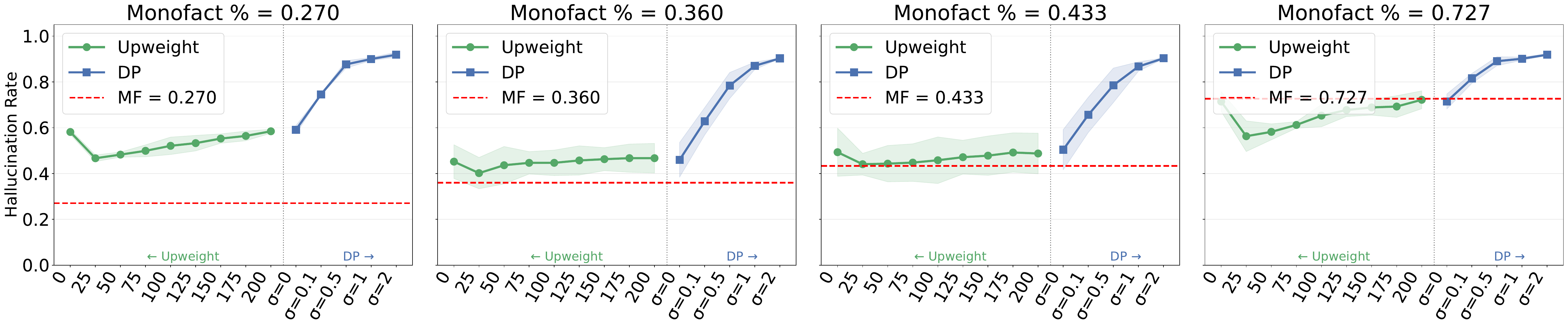}
        \caption{Hallucination rate of a bigram model under the selective upweighting strategy vs. DP Laplace noise. }
        \label{fig:miao-kearns-experiment}
\end{figure*}

Compared to GPT-2 XL, VaultGemma does have higher average FactScores for AI and pre-training articles.
As the cutoff date for GPT-2 XL's pre-training data was in 2019, this model had no exposure to most concepts in the AI articles, thus constituting an extremely low bar for factual correctness in this setting. The improved FactScores of VaultGemma over the Wikipedia pre-training may be a reflection of general improvements in LLM development over the last 5 years that are not undone by DP training. More surprisingly, VaultGemma fails to output more factually correct information than GPT-2 XL in the Wikipedia Science setting, even though these articles were all created after 2020, suggesting they were included in VaultGemma training data and not GPT-2 XL data (the difference between scores is not significant at the response level ($p=.752$), and the topic-level test favors GPT-2 XL ($p=.028$)). These results suggest DP-SGD pre-training can substantially weaken a model's ability to encode and output factually correct information, with DP model outputs sometimes as hallucinated as outputs from a model never directly exposed to the targeted information.

\section{RQ2: What impact does DP training have on model properties related to hallucination?}\label{rq2:what_impact_does_dp_training}

\subsection{DP miscalibration increases hallucinations}

Having established that DP training increases hallucinations, we investigate possible underlying mechanisms. We ground this investigation in recent theory \cite{10.1145/3618260.3649777} that shows that the hallucination rate of a model is governed by its calibration and the frequency of facts in its training data.

Formally, \citet{10.1145/3618260.3649777} represent this using the monofact rate $\widehat{\mathrm{MF}}$, the fraction of facts that appear exactly once in the training corpus, which is used to approximate the probability mass of facts the model has never seen. A calibrated model, whose predicted probabilities match the observed frequencies from the training data, thus spreads the probability mass over the \emph{unobserved} candidates from the unseen data, and \citep{10.1145/3618260.3649777} show that this yields a lower bound on the hallucination rate such that:

\begin{equation}
f_{\mathrm{gen}} \;\gtrapprox\; \widehat{\mathrm{MF}} \;-\; \mathrm{Mis}(g, p)
%\;-\; \frac{3e^{-m}}{\delta} \;-\; \sqrt{\frac{6\ln(6/\delta)}{n}},
\label{eq:monofact-bound}
\end{equation}
where $\widehat{\mathrm{MF}
}$ is the monofact rate, $\mathrm{Mis}({g, p})$ is the divergence of the model's learned distribution from the true distribution of factual frequencies. This implies that a calibrated model cannot do better than the monofact rate, but that miscalibration could decrease hallucinations.  

Building on this, \citet{doi:10.1073/pnas.2533582123} construct and validate an empirical analog of the theory of \citet{10.1145/3618260.3649777} and show that targeted miscalibration (via selective upweighting of training samples) that concentrates probability mass on high-confidence factual associations, can reduce hallucinations by up to 40\%. We investigate the effects of DP-induced miscalibration on hallucination rates by replicating the bigram model setup from \citet{doi:10.1073/pnas.2533582123}, as it allows us to study the effect of the upweighting and DP-based interventions on the model's hallucination rate using a controlled synthetic setup where the true distribution of facts is known.

\mypar{Setup} We replicate the bigram setup almost identically to the experiments from \citet{doi:10.1073/pnas.2533582123}, except that we use synthetic facts of the form $\texttt{Person}_{i} \; \texttt{Food}_{j} \; \texttt{Place}_{k} \;$, with the vocabulary constituting 40, 25 and 25 unique entities respectively, giving us a total of 25000 possible statements, out of which we designate $\mathcal{F} = 2000$ of these as ``true" facts. The frequency of each true fact is drawn from a Pareto distribution, which also allows us to modify the $\widehat{\mathrm{MF}}$ (a small $\alpha$ concentrates frequency on few facts, which lowers $\widehat{\mathrm{MF}}$ whereas a large $\alpha$ spreads it across many unique facts, raising $\widehat{\mathrm{MF}}$). From this, a training set of $|S| = 200$ samples is drawn with replacement, on which we fit a bigram model to predict the token that succeeds a given token. During the evaluation phase, the model generates 3000 tuples, over which we report the hallucination rate (defined as the proportion of generated tuples not belonging to $\mathcal{F}$) in \Cref{fig:miao-kearns-experiment}.

The upweighting strategy from \citet{doi:10.1073/pnas.2533582123} injects $k$ training facts, each of which we duplicate 10 times so as to make the model overconfident on those tuples when renormalizing the modified transition count tables into updated probabilities. For our DP comparison, we add Laplacian noise to every entry of all possible transition counts from the vocabulary space, with $L_1$ sensitivity of $3$, with all negative counts being clamped to zero. DP differs from strategic upweighting in that upweighting concentrates probability mass on observed facts, whereas DP is a non-targeted intervention that can disperse mass onto unseen transitions as well. We report the results in~\Cref{fig:miao-kearns-experiment}, where upweighting is found to produce a modest, non-monotonic decrease in hallucination, in contrast to DP noise that progressively inflates hallucination rates as the scale of noise added increases.

\subsection{Hallucinations in DP fine-tuned LLMs}

\begin{figure}[h]
    \centering
    \begin{subfigure}{0.23\textwidth}
        \centering
        \includegraphics[width=\linewidth]{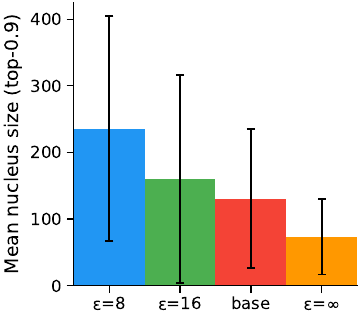}
        \caption{\footnotesize{Wikipedia Science}}
    \end{subfigure}
    \hfill
    \begin{subfigure}{0.23\textwidth}
        \centering
        \includegraphics[width=\linewidth]{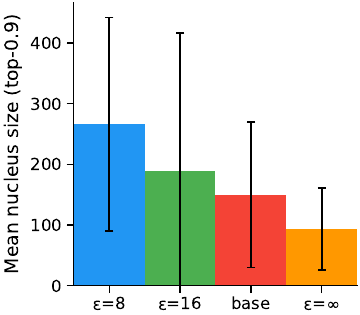}
        \caption{\footnotesize{Wikipedia AI}}
    \end{subfigure}
    \caption{Mean nucleus size (top-$p = 0.9$) for models fine-tuned under different DP budgets. DP models exhibit larger nucleus sizes, indicating greater dispersion of probability mass across tokens and increased next-token uncertainty.}
    \label{fig:nucleus sizes}
    \vspace{-0.5cm}
\end{figure}

\begin{figure}[h]
\centering
\begin{subfigure}{0.45\textwidth}
\centering
\includegraphics[width=\linewidth]{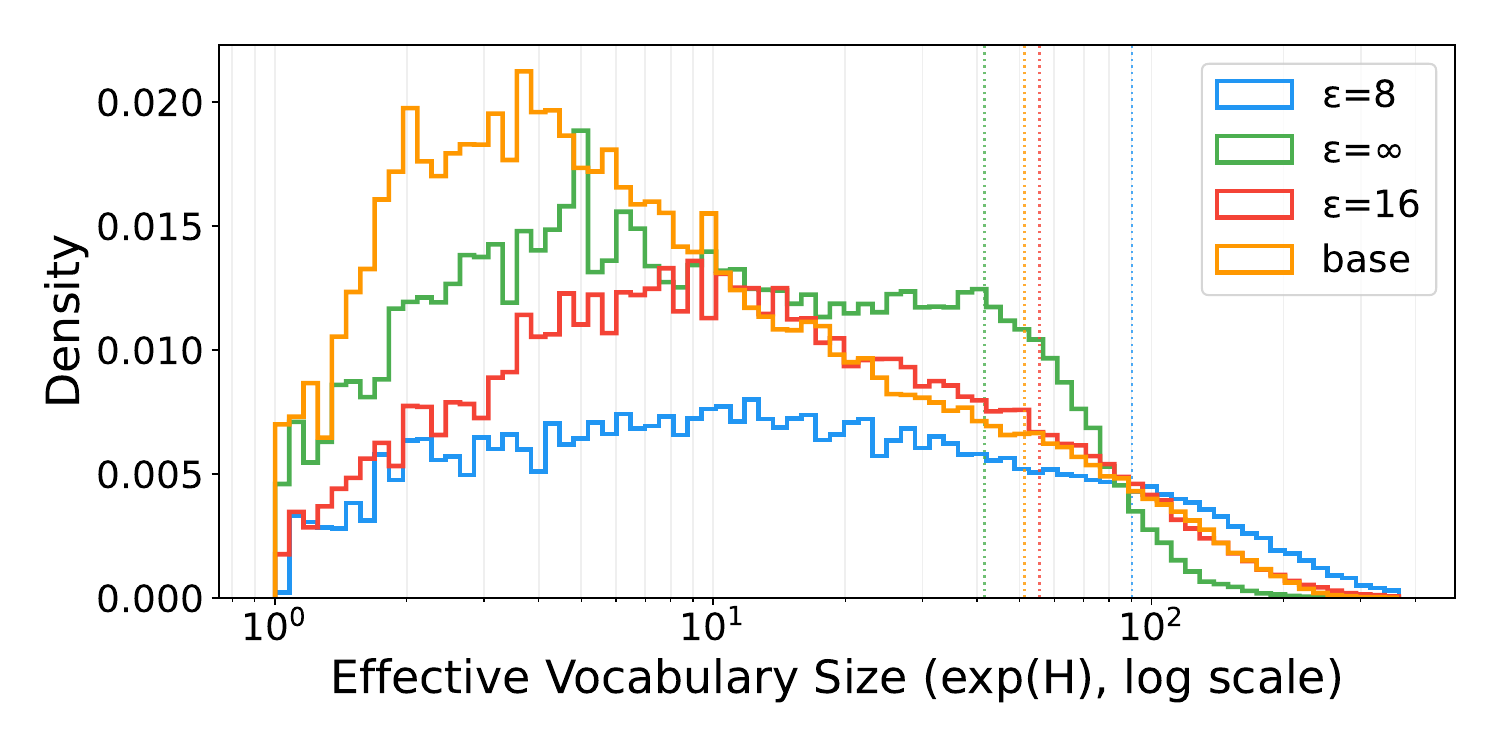}
\caption{\footnotesize{Wikipedia Science}}
\end{subfigure}
\\
\hspace{0.01\textwidth}
\begin{subfigure}{0.45\textwidth}
\centering
\includegraphics[width=\linewidth]{graphs/revisions/rq1/effective_vocab_histogram-wiki-ai.pdf}
\caption{\footnotesize{Wikipedia AI}}
%\caption{\footnotesize{Avg. FS of each topic for $\varepsilon = 16$ vs $\varepsilon = \infty$ }}
\end{subfigure}
\caption{Distribution of effective vocabulary size ($\exp(H)$), where higher values indicate probability mass spread across more tokens. DP models show higher effective vocabulary sizes, reflecting increased distributional uncertainty.}
\label{fig:eff-vocabulary-size}
\vspace{-0.5cm}
\end{figure}

The results over the bigram models raise the question: do we observe the same shifts in probability mass in DP fine-tuned LLMs? To study this, we quantify the model's uncertainty at each position in the sequences from the unseen evaluation data in two ways: i) using the nucleus size (the smallest set of tokens accounting for 90\% of the probability mass) (\Cref{fig:nucleus sizes}) and ii) the effective vocabulary size, which is the exponentiated entropy of the next-token distribution. It represents how many tokens the model's predictive uncertainty is spread across (\Cref{fig:eff-vocabulary-size}). Both metrics reveal the same distributional dispersion as in the bigram models, which grow more pronounced under stricter privacy budgets.

While higher entropy alone does not entail hallucination, it can foster conditions under which it becomes more likely. DP mechanisms produce effects on the next-token probability distribution that are analogous to increasing the sampling temperature, which has been shown to increase hallucination rates \cite{chang-etal-2025-real}. However, altering the temperature rescales the distribution in a manner that preserves the rank order of tokens. DP mechanisms do not flatten the distribution evenly and the results from the bigram experiments suggest that they can spread mass from plausible sequences toward incorrect alternatives, increasing the likelihood of hallucination. Our results in \Cref{app:flattening-pretraining} support this possibility; on pre-training data, the token rankings are preserved and hallucination rates are largely unaffected, whereas on the unseen data, these rankings shift and we observe an increase in hallucination rates.

%Intuitively, distribution flattening
%DP mechanisms do not flatten the distribution evenly and can reduce the probability mass on plausible sequences and redistribute it to incorrect alternatives, increasing the likelihood of hallucination. 

\begin{figure*}[th]
    \centering
        \centering
        \includegraphics[width=0.85\linewidth]{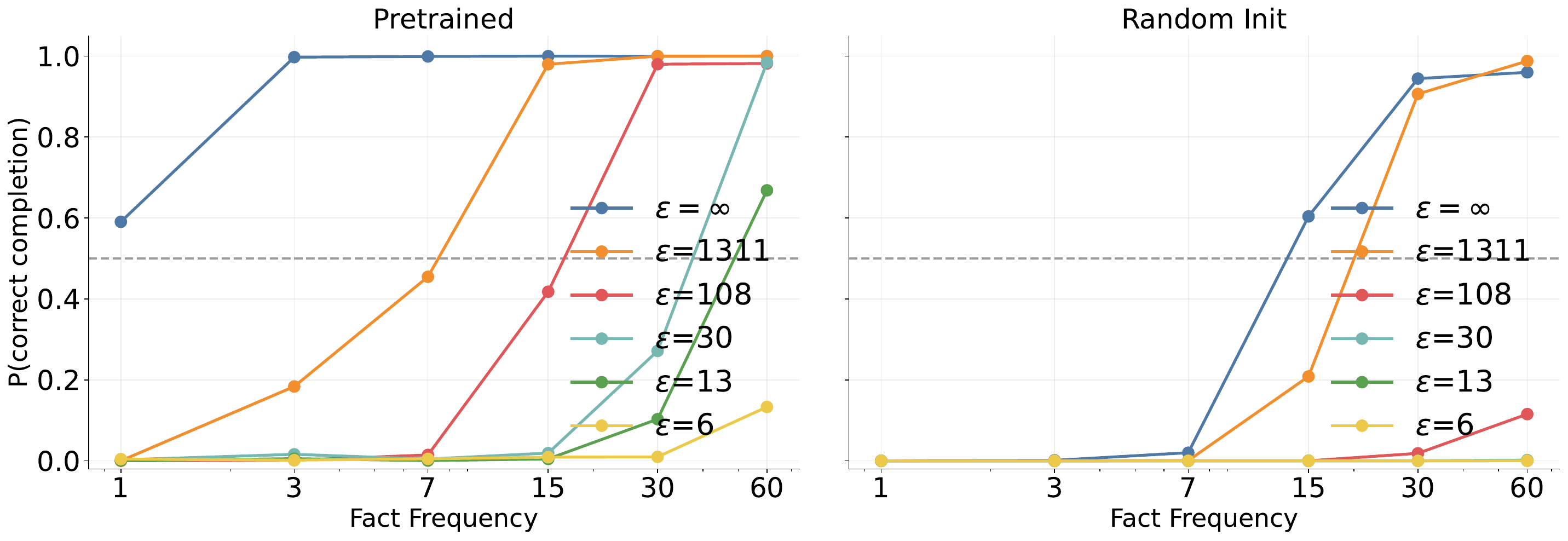}
        \caption{Factual recall vs.\ fact frequency under DP-SGD for pre-trained (left) and randomly initialized (right) GPT-2. Stricter privacy budgets ($\epsilon$) require increasingly frequent repetitions for learning.}
        \label{fig:rq3_synthetic}
\end{figure*}

\section{RQ3: Under what conditions could DP be usable without increasing hallucinations?}

Having shown that DP training increases hallucinations and investigated output distribution flattening as a possible underlying mechanism, we now ask a more fundamental question: under what conditions can a model trained with DP learn correct factual associations? The practical feasibility of DP training hinges on this question.

To study this, we construct a controlled memorization experiment using synthetic factual associations that take the form of entity triples (e.g.,~the capital of [subject] is [object]) sampled from a set of fictional subjects and objects to ensure that none of these facts can be answered from prior knowledge. Every fact appears $f$ times per epoch ($f \in \{1,3,7,15,30,60\}$), i.e.\
$15f$ times over the full training run, with five distinct facts per frequency tier. To approximate knowledge acquisition in fine-tuning and pre-training, we use the pre-trained version of GPT-2 and a randomly initialized version of GPT-2, both of which we train for 15 epochs on the synthesized facts with different privacy budgets. We fix the clipping norm at $C = 1.0$ and vary the amount of noise added during training, $\sigma \in \{0, 0.1, 0.3, 0.5, 0.7, 1.0\}$, where a higher $\sigma$ corresponds to a stronger privacy guarantee. Each setting corresponds to a privacy budget $\varepsilon \in \{\infty, 1311, 108, 30, 13, 6\}$. We report a detailed version 
% (which includes the noise multipliers corresponding to these privacy budgets)
of this setup in \Cref{app:controlled_mem_exp}. In this setup, we treat factual recall as the probability that the model assigns to the ground-truth answer given the fact prefix.

As $\varepsilon$ decreases (and the corresponding noise multiplier $\sigma$ increases), the minimum frequency required for the pre-trained model to learn an association rises sharply as well (\Cref{fig:rq3_synthetic}). Although at $\varepsilon=\infty$, the pre-trained model recalls facts that appear only once per epoch, at $\varepsilon = 1311$, this increases to $30$ repetitions, and to 60 repetitions at $\varepsilon = 30$. For stricter budgets (where $\varepsilon \leq 13$), no fact is learned even at 60 repetitions per epoch, indicating that DP noise does not simply slow down learning, but can prevent it entirely within the frequency range that we test. Furthermore, there is a stark gap between the pre-trained model and the randomly initialized setup. While the pre-trained model requires 7 repetitions per epoch at $\varepsilon = 1311$, the randomly initialized model requires 30 repetitions for the same $\varepsilon$. Beyond this, no fact is learned at any strict $\varepsilon$ within our frequency range.

This has implications for DP fine-tuning and DP pre-training. In practice, factual associations are unlikely to appear so frequently in a training corpus and will likely not be acquired by models fine-tuned over sensitive data with meaningful DP guarantees. For DP pre-training, which already entails high compute and memory costs, these frequency thresholds are considerably worse. Overall, our findings highlight the urgent need for methods that can balance meaningful privacy guarantees while still preserving the reliability of model outputs.

%This poses implications for DP pre-training, which already entails a high compute and memory cost, as it suggests that the pre-training corpus must be sufficiently representative of every concept that the model is expected to acquire. For a meaningful privacy guarantee, facts that are only sparsely represented in the corpus are unlikely to be learned at all, which raises questions about the practical feasibility of DP pre-training. 
% and the problem of data sparsity is often compounded by factors such as multilinguality or domain-specificity

\section{Discussion and Conclusions}

We reveal and systematically analyze a privacy–hallucination tradeoff in differentially private language models. Our results show that differential privacy significantly hinders the acquisition of new factual associations during pre-training and fine-tuning, leading to hallucinations. Our investigation into possible underlying mechanisms reveals that DP shifts model outputs toward factually incorrect content, and that moderate levels of noise can preclude the acquisition of a factual association.

Overall, our findings highlight how existing DP training methods can inadvertently compromise factual reliability, which is of particular concern in high-stakes applications. This underscores the need for refined privacy-preserving approaches that balance privacy guarantees with factual accuracy. Addressing this tradeoff is essential for deploying trustworthy AI systems in sensitive domains such as healthcare and law, where both privacy and factual consistency are non-negotiable.

\section*{Limitations}

Our experiments are limited to open-source models with known pre-training data cutoffs, to ensure the availability of sufficient unseen evaluation and training data for our experimental setup to adhere to differentially private constraints.

We do not investigate methods to decrease hallucinations, such as specialized post-training objectives to steer models toward expressing uncertainty or refusal over generating incorrect content. Given that fully mitigating hallucinations remains an unsolved problem even in non-DP settings, we do not expect such approaches to eliminate the risk of hallucination, although they could still reduce the tradeoff characterized in this work. Additionally, while this work focuses on DP training, we leave the investigation of this tradeoff in other DP approaches (such as DP retrieval-augmented generation \citep{koga2024privacy,grislain2025rag}) to future work, as maintaining privacy guarantees under multiple queries remains challenging, and more research is needed to assess if hallucinations persist in these settings.
\section*{Acknowledgements}

The authors would like to thank the reviewers for their helpful feedback. This work was supported in part by the AI2050 Fellowship program by Schmidt Sciences.

\bibliography{citations/anthology, citations/custom}

\appendix
\clearpage
\addtocontents{toc}{\protect\setcounter{tocdepth}{2}}
\tableofcontents
\vspace{1em}
\section{Background and Related Work}

\subsection{Related Work}
\label{app:related_work}

\mypar{Privacy in LLMs}
This risk of privacy leakage by language models has inspired work on provable privacy-preserving strategies such as differential privacy \citep{li2022large,miranda2025preservingprivacylargelanguage}, as well as heuristics such as knowledge unlearning to reduce the influence of sensitive data points on the model parameters \citep{jang-etal-2023-knowledge, zhang2024negative} and knowledge editing to locate and modify neurons containing private information \citep{wu-etal-2023-depn, wu-etal-2024-mitigating-privacy}.

Privacy-preserving methods typically (either directly or indirectly) involve changes to the model's parameters. For instance, DP training is typically accomplished through a modified version of SGD, which involves clipping and noising the gradient update \citep{Abadi2016DPSGD}. Knowledge unlearning methods trace and remove an approximate estimate of the influence of a training point on the model's parameters. All of these approaches contribute to some degradation in model utility: DP limits the information the model learns, while knowledge unlearning and knowledge editing affects useful non-private information the model has encoded.  We focus on the rigorous and future-proof guarantees of DP since heuristic privacy defenses can often be broken \cite{aerni2024evaluations,DBLP:journals/corr/abs-2406-13348}.

\mypar{Tradeoffs from DP} 
The classical no-free-lunch theorem of DP states that DP necessarily incurs a penalty on utility~\cite{kifer2011no}. In practice, this results in a privacy-utility-compute tradeoff~\cite{DBLP:conf/iclr/McMahanRT018,DBLP:journals/jair/PonomarevaHKXDMVCT23}.
Research has since established that privacy also comes at a cost to fairness in statistical estimation tasks~\cite{tran2021decision}, discriminative models~\citep{10.5555/3454287.3455674, farrand2020privatefairimpactdata,deoliveira2024empiricalanalysisfairnessnotions}, and LLMs \citep{lyu-etal-2020-differentially,matzken-etal-2023-trade, ramesh-etal-2024-evaluating,hansen-etal-2024-impact}.
\citet{ngong-etal-2025-differentially} explore the adverse effect of DP on elements such as grammatical correctness, fluency and the coherence of model-generated text.
% - considerations that are pertinent only to models trained on unstructured data. 
We note that it is possible to optimize for more balanced privacy-utility tradeoffs \citep{mireshghallah-etal-2021-privacy} or privacy-fairness tradeoffs \citep{pillutla2024federated,zhou2024differentially}.

Specifically in the text domain, prior work has empirically examined the privacy-utility tradeoffs with task-specific measures of utility, including classification accuracy, linguistic aspects such as fluency, grammatical correctness and lexical diversity. In contrast, we focus on tradeoffs between DP and hallucination. Minimizing  hallucinations is highly desirable, and is often distinct from other task-specific measures of utility surveyed above. Concurrent with our work, \citet{joo2026pearl} find that DP-decoding in RAG settings increases hallucinations in models, attributing this to a similar DP-induced flattening effect which amplifies knowledge conflict, which is measured as the gap between the retrieval-conditioned predictions and the model's parametric prior.

\mypar{Factuality in Language Models}
% Evaluating the factuality of LLMs  is challenging; 
No single factuality metric generalizes across settings, so it is common to use task- or domain-specific methods.
% language models has typically been approached with task or domain-specific methods, with no single metric generalizing effectively across all possible settings. 
Early research focused on measures such as the factual precision of cloze-style or short-form responses \citep{youssef-etal-2023-give, petroni-etal-2019-language}, and NLI-based methods to determine whether generated summaries are  consistent with their source document \citep{chen-etal-2021-nli-models, fabbri-etal-2022-qafacteval, laban-etal-2022-summac}. 
The rise of LLM applications with open-domain, free-form model-generated text, where there can be multiple plausible responses has led to the development of factuality metrics in these settings \citep{min-etal-2023-factscore, wei2024longformfactualitylargelanguage, song-etal-2024-veriscore}. This is significantly more challenging, as the decoding strategy  also influences model outputs \citep{wang-etal-2024-factuality}, and models can generate correct answers across multiple attempts \citep{tian2023finetuninglanguagemodelsfactuality}. Further, they may generate factually accurate content that contradicts or is unsupported by the training data \citep{cao-etal-2022-hallucinated}.\footnote{
We regard this as hallucinations for the purpose of this work as the training dataset is considered as the sole \emph{source of truth.}} 
We leverage existing state-of-the-art methods for fact-checking in open-ended generation along with human evaluations to analyze privacy-hallucination tradeoffs in LLMs \cite{min-etal-2023-factscore}.
% analyses to further advanceunderstanding of factuality in LLMs, specifically in relation to privacy. 

\subsection{Background: Differential Privacy}
\label{app:dp_background}

Differential privacy offers a formal privacy guarantee that ensures that any individual's data cannot be inferred from a query applied to a dataset \citep{Dwork2006}. In other words, the result of such a query is nearly indistinguishable from the result of the same query applied to a dataset that either includes a modified version of the individual's data or excludes the record entirely, thereby preserving the individual's privacy. In this case, the notion of adjacency specifies exactly how changing single record in the original dataset $D$ yields the modified dataset $D'$.

We review the definition of DP here; we refer to the textbooks \cite{dwork2014algorithmic,fioretto2025differential} for specific details and the guide \cite{DBLP:journals/jair/PonomarevaHKXDMVCT23} for details on DP-SGD.

Formally, differential privacy is defined as follows:

\textbf{Definition}: 
Two datasets $D$ and $D'$ are said to be neighboring (in the add-or-remove sense) if $D = D' \cup \{x\}$ or $D' = D \cup \{x\}$ for some training example $x$.

A randomized algorithm \(A\) is \((\epsilon, \delta)\)-private for  some \(\epsilon > 0\) and \(\delta \in [0, 1]\)if for \emph{any} two neighboring datasets \(D, D'\), the following holds true for all sets \(Y\)  in the range of $A$:
\[
\Pr[A(D) \in Y] \leq e^{\epsilon} \Pr[A(D') \in Y] + \delta .
\]
The value of \(\epsilon\) denotes the privacy budget, while \(\delta\) specifies the likelihood that the privacy guarantee may fail. If \(\delta\) is set to 0, this implies a purely differentially private setting with no probability of the guarantee being broken. The value of \(\epsilon\) constrains how similar the outputs of both distributions are; a higher \(\epsilon\) value indicates a greater privacy budget, meaning the algorithm is less private. DP guarantees that even if an adversary has access to any side-knowledge, the privacy leakage of \((\epsilon, \delta)\)-DP algorithms will not increase. Additionally, another property of DP is that it ensures that any post-processing on the outputs of \((\epsilon, \delta)\)-differentially private algorithms will remain \((\epsilon, \delta)\)-differentially private. 

We use DP-SGD \citep{Abadi2016DPSGD}, a modification to the stochastic gradient descent (SGD) algorithm, which is typically used to train neural networks. DP-SGD clips the gradients to limit the contribution of individual samples from the training data and subsequently adds noise from the Gaussian distribution to the sum of the clipped gradients across all samples.\footnote{
    The noise added is independent coordinate-wise and across time, although DP-SGD with temporal correlations has recently been of growing interest~\cite{pillutla2025correlated}. 
} 
DP-SGD thus provides a differentially private guarantee to obfuscate the gradient update, thereby ensuring that the contribution of any given sample in the training data is indistinguishable due to the aforementioned post-processing property. This process ensures \((\epsilon, \delta)\)-differential privacy for each model update. Given a privacy budget, number of epochs, and other training parameters, we can estimate the privacy parameters using standard privacy accounting algorithms, which implemented in common software.

%\clearpage
\subsection{FactScore and Recurring Claims Algorithms}\label{app:factscore}

We recall the pseudo-code of FactScore in \Cref{alg:factscore_detailed_exp} and describe our algorithm to cluster repeated claims in \Cref{algo:claim-cluster} (cf. \Sref{sec:recurrent-claim}).

\subsubsection{FactScore Algorithm}
\begin{algorithm}[h]
    \caption{FActScore: Atomic Fact Extraction and Verification}
    \label{alg:factscore_detailed_exp}
    \textbf{Input:} Generated texts $\mathcal{D} = {d_1, d_2 ... d_n}$, atomic fact extractor module $\mathcal{E}$, claim verification model $\mathcal{V}$, knowledge source $\mathcal{K}$ % \kpcomment{Prefer $E, V, K$ rather than their mathcal versions}
    \\
    \textbf{Output:} FactScore for each document in the generated corpus : $\mathcal{S}(\mathcal{E}, \mathcal{D})$
    \begin{algorithmic}[1]
    \For{each document $d_i \in \mathcal{D}$}
        \State Extract a candidate set of atomic claims: 
        % \kpcomment{Prefer ``$\mathrm{AF}_i$'' rather than ``$\mathcal{AF}_{d_i}$''. Also, consider inlining this equation}
        \[
        \mathcal{AF}_{d_i} = \mathcal{E}(d_i) %= \{ \alpha^{(d_i)}_j \}_{j=1}^{m}
        \]
        % \kpcomment{Avoid superscript $(d_i)$, consider just using $(i)$ instead}
        \For{each atomic claim $\alpha^{(d_i)}_j \in \mathcal{AF}_{d_i}$}
            \State Verify factuality if $\alpha^{(d_i)}_j$ is supported by knowledge source $\mathcal{K}$:
            \[
            \hat{y}^{(d_i)}_j = \mathcal{V}(\alpha^{(d_i)}_j, \mathcal{K}) \quad \text{where } \hat{y}^{(d_i)}_j \in \{0, 1\}
            \]
        \EndFor
        \State Compute per-document precision:
        \[
        \mathcal{S}(\mathcal{E}, g) = \frac{1}{|\mathcal{AF}_{d_i}|} \sum_{j=1} \mathbb{I}(\hat{y}^{(d_i)}_j = 1)
        \]
    \EndFor
    \end{algorithmic}
    \label{alg:factscore}
\end{algorithm}

Algorithm ~\ref{alg:factscore} describes how the FactScore is computed for a set of generated documents. For each document $d_i$, the atomic fact extractor $\mathcal{E}$ which is an instruction-tuned LLM, is prompted with in-context examples to decompose the text into a set of atomic claims, denoted $\mathcal{AF}_{d_i}$. 

Each atomic claim $\alpha^{(d_i)}_j \in \mathcal{AF}_{d_i}$ is then independently verified using an external knowledge source $\mathcal{K}$. The knowledge source is the reference data against which claims are verified, and in our setup, this consists of the relevant Wikipedia articles for the evaluation domain. For each atomic claim, evidence passages are retrieved (e.g. via BM25) from these articles which are then provided to the claim verification model. The claim verification model ($\mathcal{V}$) then judges whether the claim is supported by this knowledge, producing a binary label $\hat{y}^{(d_i)}_j \in \{0,1\}$ that indicates whether or not the claim is factually supported.

The FactScore for the document is computed as follows:
\[
\mathcal{S}(\mathcal{E}, d_i) = \frac{1}{|\mathcal{AF}_{d_i}|} \sum_{j} \mathbb{I}(\hat{y}^{(d_i)}_j = 1),
\]
which yields a scalar score corresponding to the factual correctness of the information in the generated document with respect to the knowledge source. A higher score translates to a greater proportion of verified claims, and conversely, a lower score is indicative of fewer supported claims.

We use Llama-3.1-8B-Instruct, to perform both the atomic fact extraction as well as the atomic claim verification. To demonstrate that our results remain consistent across different choices in the claim decomposition and claim verification modules used, we also report results over other instruction-tuned LLMs such as DeepSeek-R1-Distill-Qwen-7B and Llama 3.2-3B Instruct in Table ~\ref{tab:llama-3b-factscore-claim-eval} and Table ~\ref{tab:deepseek-factscore-claim-eval}.

\section{Definition of Hallucination}\label{app:define_hallucination}

The term ``hallucination'' typically refers to the generation of factually incorrect information by generative models. For this work, we focus on a class of facts supported by the model's data.

\noindent \textbf{Definition}: We say a model \emph{hallucinates} if it generates a claim that is not supported by the model's training data. Hallucinations w.r.t. finetuning are claims that are not supported by the finetuning data and hallucinations w.r.t. pretraining are claims not supported by the pretraining data.

Here are a few consequences of our definitions:
\begin{itemize}
    \item If the model generates a correct fact that is not grounded in its finetuning (resp. pretraining) data, we consider it a hallucination. 
    % This is the case even if it is true in an external knowledge base.
    \item If the model generates a correct fact that is grounded in the pretraining data but not the finetuning data, we consider it a finetuning hallucination.
    \item Conversely, a factually correct claim is not considered a hallucination provided it appears somewhere in the training data, even if it is absent from the specific source article being evaluated.
\end{itemize}

Our evaluation is tailored to this definition: we specifically target the model's ability to acquire and reproduce facts from previously unseen data. This definition allows us to directly assess how DP influences model learning. Failure to learn a fact is a different problem from hallucination: in the absence of knowledge acquisition, a model could output no content or generic content, which we might expect to be the case, as prior work has shown DP outputs tend to be shorter than non-DP outputs \cite{ngong-etal-2025-differentially, cano2025differentiallyprivatetextgenerationdegrades}. Instead our work highlights that DP increases hallucinations, which is not inherently implied by the definition of DP.

%\onecolumn
\section{Experimental Setup}

\subsection{Datasets}\label{app:dataset_description}
%\label{sec:methods_datasets}

Choosing a dataset for this study requires careful consideration of two factors. 
First, 
DP guarantees hinge on the assumption that the private fine-tuning data should not have appeared in the pre-training corpora of the LLM~\cite{DBLP:conf/icml/Tramer0C24,cummings2023advancing}. The importance of not violating this condition can be attributed to i) the potential for pre-training data to be adversarially extracted \citep{ishihara-2023-training}, and ii) evidence that pre-training and fine-tuning on the same data artificially inflates performance estimates \citep{igamberdiev-etal-2022-dp}. 
Second, the inclusion of factually verifiable information and statements in the fine-tuning data is essential to evaluate changes in the factual correctness of the model's outputs; domains and datasets (e.g., social media posts) without clear factual content cannot be assessed for factuality.

In view of these two factors, we focus on Wikipedia data for fine-tuning and evaluation, where content is constructed to contain verifiable facts rather than opinions or speculation, and automated fact-checking methods have been previously validated \citep{min-etal-2023-factscore}. Additionally, meta-data allows us to select articles created after 2020, ensuring they were not included in GPT-J 6B's pre-training corpora.
% Therefore, to isolate and study the effects of private fine-tuning on factual knowledge acquisition, we select and craft datasets that i) are explicitly not included in the pre-training corpus of the LLM ii) contain factually verifiable information.
% 
We expect our fine-tuning data to have some overlap with text in the pre-training data in terms of linguistic patterns, broad concepts, and topics (in some settings). This overlap is not inherently problematic, as it reflects natural language settings, where syntactic and semantic structures are rarely novel and learning dynamics are influenced by previously learned distributions.

\subsubsection{Evaluation Datasets}
We use three datasets for factuality evaluation as summarized in \Cref{tab:datasets} and described below. An exact list of topics included in these datasets is given in Appendix~\ref{app:list_of_topics}.

\mypar{Wikipedia Science}
We collect 231 Wikipedia articles on science topics that were created after the cutoff date for GPT-J 6B's pre-training data, where we use keyword searching of Wikipedia meta-data to identify science articles. We focus on science topics as they contain  detailed technical language, which is also common in sensitive data settings (e.g., clinical notes).
While Wikipedia articles on these topics did not exist before 2020, we do expect some concepts to exist in other pre-training data sources, which makes it feasible for a DP model to produce facts on these topics, even without memorizing individual data points.

\mypar{Wikipedia AI}
We collect 124 Wikipedia articles on AI topics, where we hand-curate products and models that did not exist before 2020, along with related articles we expect to mention them. Unlike the Wikipedia Science articles, GPT-J 6B cannot have any knowledge of most of these concepts without fine-tuning, as they could not have existed in pre-training data. However, by constructing our data to contain articles that mention overlapping topics, we ensure that it is feasible for a DP model to learn them. For example, if our dataset only contained \textit{DeepSeek (chatbot)}, DP would preclude learning of information isolated to one data point. By including  \textit{DeepSeek (chatbot)}, \textit{DeepSeek}, and \textit{DeepSeek (disambiguation)}, a DP model can hypothetically learn information about DeepSeek, as it is mentioned in multiple data points.

\mypar{Wikipedia pre-training}
To investigate effects of DP fine-tuning on knowledge acquired during pre-training, we randomly sample 250 Wikipedia articles from the GPT-J pretraining data that are not included in the fine-tuning data.

\subsubsection{Fine-tuning Dataset}
Fine-tuning with differential privacy generally requires large enough datasets and large batch sizes (e.g. $\Omega(10^3)$ or more)~\cite{DBLP:conf/iclr/McMahanRT018,DBLP:journals/jair/PonomarevaHKXDMVCT23}.
% requires large batch sizes for the noise multipliers associated with user-defined ($\epsilon$, $\delta$)-DP budgets. 
However, since our curated evaluation sets are insufficient to meet these batch size specifications, we intersperse our collected articles with an additional 20,000 randomly sampled Wikipedia articles that likely occurred in the pre-training data. We ensure that these samples do not overlap with the Wikipedia pre-training dataset used for factuality evaluation.

The data is divided into sequences of 512 tokens, which is the unit of privacy protection. We fine-tune the model to produce an article when prompted on the article title (e.g., a topic). 
For evaluations, we similarly prompt the model with article titles and evaluate the factual accuracy of the generated text. 

\subsection{Experimental Setup: Methods and Hyperparameters}\label{app:exp_setup}

DP fine-tuning is achieved using a stochastic gradient optimization approach known as DP-SGD~\cite{Abadi2016DPSGD}. This algorithm bounds the information learned from each sample by clipping the per-sample gradients to a fixed $\ell_2$ norm bound, and perturbs them (for DP) with white Gaussian noise. The scale of the Gaussian noise is calibrated to the desired $(\epsilon, \delta)$-DP guarantee. 
The DP guarantees are provided with respect to the add-or-remove adjacency at the sequence-level, i.e., the model outputs should be nearly indistinguishable if a new sequence of $k$ tokens is added to or removed from the training dataset. We set privacy budgets of $\epsilon \in \{8, 16\}$, and $\delta= {n^{-1.1}}$, where $n$ is the dataset size (in terms of number of sequences). We use the DP-SGD implementation from Opacus \cite{yousefpour2022opacus} and measure the privacy budget consumed using the PLD accountant (with amplification by sampling) ~\cite{doroshenko2022connect}. 

All experiments were conducted on NVIDIA A100 GPUs. DP fine-tuning was performed using 1$\times$A100 GPUs, with each DP fine-tuned model requiring approximately 24 GPU hours.

\begin{table*}[htbp]
\renewcommand{\arraystretch}{1.2}
\small
\centering
\begin{tabular}{lcccccccc}
\toprule
\textbf{Setting} & \textbf{Batch Size} & \textbf{Epochs} & \textbf{LR} & \textbf{Clip} & \textbf{LoRA $r$} & \textbf{LoRA $\alpha$} & \textbf{Seq. Len.} & \textbf{DP/PLD} \\
\midrule
Vanilla        & 8    & 15  & $1 \times 10^{-4}$ & 1.0 & 4 & 32 & 512 & -- \\
DP, $\epsilon \in \{8, 16\}$   & 4096 & 20  & $1 \times 10^{-4}$ & 1.0 & 4 & 512 & 512 & $\delta=1/n^{1.1}$ \\
\bottomrule
\end{tabular}
\caption{Summary of major hyperparameters in vanilla and private training settings.}
\label{tab:hyperparams}
\end{table*}

The PLD accounting algorithm proposed in \citet{doroshenko2022connect} provides us with tighter estimates of the privacy loss as compared to alternate accounting techniques. This in turn allows us to more accurately determine the noise multiplier required to satisfy the specified privacy budget for our fine-tuning setups. 

While VaultGemma's DP pre-training uses a unit of privacy of 1024 tokens (which includes separating and merging documents based on this fixed sequence length), our Wikipedia-based fine-tuning experiments use a single document as the privacy unit, which is reflective of settings where each document corresponds to a distinct individual.

\paragraph{Why VaultGemma is not data-matched.} 

We evaluate against the two baselines used in the VaultGemma technical report~\citep{vaultgemma}, Gemma3 (1B) and GPT-2 (1.5B), together with Gemma2 (2B) and Gemma (2B). No exactly matched non-private counterpart exists, but VaultGemma does share the same Gemma2 pretraining mixture as the Gemma2 (2B) model, and VaultGemma is approximately the same size as Gemma3 (1B) (although they differ in mixture). However, this does not affect our results as i) every Gemma model likely includes the Wikipedia data in our evaluation setup in its pre-training mixture, and ii) GPT-2 (1.5B) matches VaultGemma on Wikipedia Science despite being a 2019 model trained on vastly less data, and despite WebText excluding Wikipedia outright.

Additionally, data-matching would require DP pre-training from scratch, which is infeasible at the required scale. The utility in using DP-SGD's to train language models hinges on very large batch sizes. For reference, VaultGemma was trained with an expected batch size of ${\approx}500$k, and even smaller models like BertTiny require a batch size of ${\approx}283$k \citep{mckenna2025scaling}. \citet{mckenna2025scaling} make the same point via DP scaling laws, where at $10^{22}$ FLOPs, ${\sim}10^{8}$ parameters are compute-optimal under privacy versus ${\sim}10^{10}$ non-privately. Due to the intense computational demands for pre-training these models, we use VaultGemma as a reference point for what DP pre-training yields at comparable scale, not as a controlled ablation of pre-training data. We leave further exploration of this problem through the use of controlled re-training experiments to future work.

The hyperparameters from our DP fine-tuning setup are summarized as follows:

%in Table~\ref{tab:hyperparams}.

\paragraph{LoRA Gradient Update}

We apply Low-Rank Adaptation (LoRA) \cite{Hu2022} in all experiments. The adapted weight is parameterized as
\[
    W = W_0 + \frac{\alpha}{r} AB
\]
where the LoRA matrices are $A \in \mathbb{R}^{d \times r}$, $B \in \mathbb{R}^{r \times k}$, and $W_0$ is the frozen pre-trained weight. During training, the LoRA parameters $A, B$ are updated via:
\begin{multline*}
(A^{(t+1)}, B^{(t+1)}) = (A^{(t)}, B^{(t)}) \\
- \eta \cdot \mathrm{clip}\left(\nabla_{A,B} L,\, c\right)
\end{multline*}
where $\eta$ is the learning rate, $c = 1.0$ is the clipping norm, and $L$ is the loss function. In the DP setting, noise is added to the clipped gradient.

During training, the update step for $(A, B)$ in SGD or DP-SGD is:
\[
    (A^{(t+1)}, B^{(t+1)}) = (A^{(t)}, B^{(t)}) - \eta \cdot \widetilde{\nabla} L
\]
where
\[
    \widetilde{\nabla} L = \mathrm{clip}\left(\nabla L, c\right) + \mathcal{N}(0,\,\sigma^2)
\]

 \[
    \nabla_{A} L = \frac{\alpha}{r} \ \nabla_W L \cdot B^\top
 \]

Instead of using a large learning rate in the DP fine-tuning, we use a larger value of ($\alpha = 512$) in the DP setting (compared to ($\alpha = 32$) in the standard fine-tuning setting) to amplify the contribution of the adapted weights without increasing the noise magnitude (which is applied to the gradients directly). This prevents the gradient signal from being obscured by the DP noise.

\paragraph{Why LoRA rather than full fine-tuning.}
DP-SGD requires per-example gradients, and the resulting memory overhead makes full fine-tuning at this scale computationally infeasible for most academic labs, including our own. Additionally, given that LoRA-based DP fine-tuning has already been shown to be competitive with full fine tuning \cite{kurakin2024harnessinglargelanguagemodelsgenerate}, and parameter-efficient adaptation is a standard practice in related work \cite{yu2022differentially, tan2025synthesizing}, our setup is reflective of common practices in this area.

\clearpage
\onecolumn
\subsection{Dataset : Examples of Input}
\label{app:examples_input}

\begin{longtable}{p{3cm} p{12cm}}
\caption{Examples of excerpts of the input and output pairs from our Wikipedia AI and Science articles.} \label{tab:eval-examples} \\

\toprule
\textbf{Input} & \textbf{Output} \\
\midrule
\endfirsthead

\toprule
\textbf{Input} & \textbf{Output} \\
\midrule
\endhead

\bottomrule
\endfoot

Right to Know &
Right To Know is a non profit support project for those who discover via genealogical genetic testing that their lineage is not what they had supposed it to be due to family secrets and misattributed parentage, thus raising existential issues of adoption, race, ethnicity, culture, rape, etc. == See also == Genealogy Genetic testing == External links == Right To Know - Your Genetic Identity. \\

Neurosemiotics &
Neurosemiotics is an area of science which studies the neural aspects of meaning making. It interconnects neurobiology, biosemiotics and cognitive semiotics. Neurolinguistics, neuropsychology and neurosemantics can be seen as parts of neurosemiotics. == Description == The pioneers of neurosemiotics include Jakob von Uexküll, Kurt Goldstein, Friedrich Rothschild, and others. The first graduate courses on neurosemiotics were taught in some American and Canadian universities since 1970s. The term 'neurosemiotics' is also not much older. Neurosemiotics demonstrates which are the necessary conditions and processes responsible for semiosis in the neural tissue. It also describes the differences in the complexity of meaning making in animals of different complexity of the nervous system and the brain. == See also == Semiotics Zoosemiotics. \\

Cyclosiloxanes &
Cyclosiloxanes are a class of silicone material. They are volatile and often used as a solvent. The three main commercial varies are octamethylcyclotetrasiloxane (D4), decamethylcyclopentasiloxane (D5) and dodecamethylcyclohexasiloxane (D6). They evaporate and degrade in air under sunlight. == Octamethylcyclotetrasiloxane (D4) == The octamethylcyclotetrasiloxane silicone liquid has no odor and consists of four repeating units of silicon (Si) and oxygen (O) atoms in a closed loop giving it a circular structure. Each silicon atom has two methyl groups attached (CH3). == Decamethylcyclopentasiloxane (D5) == Decamethylcyclopentasiloxane silicone liquid has no odor and consists of five repeating units of silicon (Si) and oxygen (O) atoms in a closed loop giving it a circular structure. Each silicon atom has two methyl groups attached (CH3). Typically it is used as an ingredient in antiperspirant, skin cream, sun protection lotion and make-up. With a low surface tension of 18 mN/m this material has good spreading properties. \\

Cancer exodus hypothesis &
The cancer exodus hypothesis establishes that circulating tumor cell clusters (CTC clusters) maintain their multicellular structure throughout the metastatic process. It was previously thought that these clusters must dissociate into single cells during metastasis. According to the hypothesis, CTC clusters intravasate (enter the bloodstream), travel through circulation as a cohesive unit, and extravasate (exit the bloodstream) at distant sites without disaggregating, significantly enhancing their metastatic potential. This concept is considered a key advancement in understanding of cancer biology and CTCs role in cancer metastasis. == Mechanism == Traditionally, it was believed that CTC clusters needed to dissociate into individual cells during their journey through the bloodstream to seed secondary tumors. However, recent studies show that CTC clusters can travel through the bloodstream intact, enabling them to perform every step of metastasis while maintaining their group/cluster structure. \\

Generative pre-trained transformer &
Generative Pre-trained Transformer 1 (GPT-1) was the first of OpenAI's large language models following Google's invention of the transformer architecture in 2017. In June 2018, OpenAI released a paper entitled "Improving Language Understanding by Generative Pre-Training", in which they introduced that initial model along with the general concept of a generative pre-trained transformer. Up to that point, the best-performing neural NLP models primarily employed supervised learning from large amounts of manually labeled data. This reliance on supervised learning limited their use of datasets that were not well-annotated, in addition to making it prohibitively expensive and time-consuming to train extremely large models; many languages (such as Swahili or Haitian Creole) are difficult to translate and interpret using such models due to a lack of available text for corpus-building. In contrast, a GPT's "semi-supervised" approach involved two stages: an unsupervised generative "pre-training" stage in which a language modeling objective was used to set initial parameters, and a supervised discriminative "fine-tuning" stage in which these parameters were adapted to a target task. \\

GPTZero &
GPTZero is an artificial intelligence detection software developed to identify artificially generated text, such as those produced by large language models. While GPTZero was praised for its efforts to prevent academic dishonesty, many news outlets criticized the tool's false positive rate, which can be especially harmful in academic settings. == History == GPTZero was developed by Edward Tian, a Princeton University undergraduate student, and launched online in January 2023 in response to concerns about AI-generated usage in academic plagiarism. GPTZero said in May 2023 it raised over 3.5 million dollars in seed funding. In the first week of its release, the GPTZero experienced 30,000 uses, which led to a crash. It was supported by the web app company Streamlit, who allocated more server resources in response. In July 2024, it had 4 million users, compared to 1 million one year earlier. In summer 2024, GPTZero raised \$10 million in Series A round funding. In September 2024, GPTZero announced an authorship tracking software that enables "to compile and share data about their writing process such as their copy/paste history, the number of editors they had, and how long editing took", in an effort "to move away from an all-or-nothing paradigm around AI writing towards a more nuanced one." \\

GPT 4.5 &
GPT-4.5 (codenamed "Orion") is a large language model developed by OpenAI as part of the GPT series. Officially released on February 27, 2025, GPT-4.5 is available to users subscribed to the ChatGPT Plus and Pro plans across web, mobile, and desktop platforms. Access is also provided through the OpenAI API and the OpenAI Developer Playground. == Overview == It was primarily trained using unsupervised learning, which improves its ability to recognize patterns, draw connections, and generate creative insights without reasoning. This method was combined with supervised fine-tuning and reinforcement learning from human feedback. The computational resources needed for training were provided by Microsoft Azure. Sam Altman described GPT-4.5 as a "giant, expensive model". \\

Claude &
Claude is a family of large language models developed by Anthropic. The first model was released in March 2023. The Claude 3 family, released in March 2024, consists of three models: Haiku, optimized for speed; Sonnet, which balances capability and performance; and Opus, designed for complex reasoning tasks. These models can process both text and images, with Claude 3 Opus demonstrating enhanced capabilities in areas like mathematics, programming, and logical reasoning compared to previous versions. Claude 4, which includes Opus and Sonnet, was released in May 2025. == Training == Claude models are generative pre-trained transformers. They have been pre-trained to predict the next word in large amounts of text. Then, they have been fine-tuned, notably using constitutional AI and reinforcement learning from human feedback (RLHF). \\

\end{longtable}

\clearpage
\twocolumn
\subsection{List of topics in the datasets}\label{app:list_of_topics}

\subsubsection{Wikipedia AI}

\textbf{124 articles}: DALL-E; OpenAI; Midjourney; Imagen (text-to-image model); Text-to-image model; Recraft; DeepSeek; DeepSeek (chatbot); Liang Wenfeng; High-Flyer; 2025 in artificial intelligence; DeepSeek (disambiguation); Six Little Dragons; R1; Ideogram (text-to-image model); Stable Diffusion; Automatic1111; ComfyUI; Stability AI; Emad Mostaque; Artificial intelligence and copyright; Fooocus; LAION; Sai; BLOOM (language model); Gemini; Gemini (chatbot); Gemini (language model); Gemini Robotics; Gemini Home Entertainment; Jet Force Gemini; Pixel 9; NotebookLM; Large language model; AlphaEvolve; Anthropic; Google Lens; Google AI Studio; Android XR; Chris Welty; Large language models in government; ChatGPT; Generative pre-trained transformer; GPT-4; GPT; GPT-4o; GPT-3; GPT-2; GPT-4.1; GPT-4.5; GPT-1; AutoGPT; Microsoft Copilot; GPTs; GPT-J; GPT Store; OpenAI o1; GPT4-Chan; GPTZero; Sora (text-to-video model); EleutherAI; YandexGPT; Writesonic; ChatGPT in education; Pause Giant AI Experiments: An Open Letter; Deep Learning (South Park); PauseAI; Chinchilla (language model); Artificial intelligence content detection; General-purpose technology; The Last Screenwriter; Wu Dao; Microsoft Recall; Alice and Sparkle; Amazon Q; Connor Leahy; Multimodal learning; OpenAI o4-mini; 2022 in artificial intelligence; Death of an Author (novella); GigaChat; P(doom); XLNet; Boyfriend Maker; 2023 in artificial intelligence; LLMs in higher education; Perceiver; NovelAI; Supremacy (book); Rabbit r1; Preamble (company); BookCorpus; Omneky; Machine unlearning; Artificial empathy; Llama (language model); Llama.cpp; DBRX; Llama (disambiguation); Alpaca (disambiguation); Qwen; Brave Leo; B65; Mistral; Mistral AI; Arthur Mensch; General Catalyst; Cédric O; Le Chat (disambiguation); PaLM; List of large language models; Prompt engineering; Foundation model; BERT (language model); LaMDA; T5 (language model); Alibaba Group; Claude (language model); Grok (chatbot); XAI (company); Colossus (supercomputer); X Corp.; Explainable artificial intelligence; Google DeepMind

\subsubsection{Wikipedia Science}

\textbf{231 articles}: Eurotrac; Scienticide; 505(b)(2) regulatory pathway; Anti-asthmatic agent; Breastmilk medicine; Cancer exodus hypothesis; Confocal endoscopy; Diabetes self-management; Dorsal pancreatic agenesis; Drone-Enhanced Emergency Medical Services; Electronic health record (Germany); Follicular drug delivery; LAMA2 related congenital muscular dystrophy; Most Favored Nation Drug Pricing; Musicians' Medicine; Poison exon; RNU2-2 syndrome; RNU4-2 syndrome; Synthetic Cannabinoid Use Disorder; Urinary anti-infective agent; Vestibular paroxysmia; Antarlide; Bioliteracy; Cancer exodus hypothesis; Dermestarium; Functional information; Interdigitation; Plasmagene; Poison exon; Polylecty; Spatial biology; Edge states; Electrostatic solitary wave; Frenesy (physics); History of the LED; HUN-REN Wigner Research Centre for Physics; Joaquim da Costa Ribeiro; Missile lofting; Nottingham effect; Physics of Life; Quasi-isodynamic stellarator; Riccardo D'Auria (theoretical physicist); Shockwave cosmology; Synchronous lateral excitation; Toroidal solenoid; Wohlfarth Lectureship; Compliance constants; Cononsolvency; Corrosion inhibitors for the petroleum industry; Cyclosiloxane; Dark oxygen; Direct reduction; Energy-rich species; Grupo Fertiberia; Intrinsic DNA fluorescence; Krupp–Renn process; Mental gland; Probico; School of Molecular Sciences; Shape of the atomic nucleus; Stable phosphorus radicals; Superelectrophilic anion; TOP Assay; Mathematical oncology; Mathethon; The Math(s) Fix; Conductivity cell; Generalized renewal process; Glossary of engineering: M–Z; Marine construction; Museum of Engines and Mechanisms; Northern Technical College; Safer end of engineering life; Synchronous lateral excitation; The Clark Collection of Mechanical Movements; Third medium contact method; UNESCO World Engineering Day for Sustainable Development; Positive health; Bell's mania; Chialvo map; Dysfunctome; Femoral nerve dysfunction; Fiber photometry; Fork cell; High Price (book); Hyper-empathy; Large dense core vesicles; Lateral olfactory tract usher substance; Malaria therapy; Max Planck Institute for Biological Intelligence; Nerve glide; Neural synchrony; Neurosemiotics; Neurotrophin mimetics; Optogenetic methods to record cellular activity; Personality neuroscience; Representational drift; Single-particle trajectory; Smell training; Spongy degeneration of the central nervous system; Walk Again Project; Amoeboflagellate; Borg (microbiology); Chrompodellid; Dissimilatory iron reducing bacteria; Garrod Lecture and Medal; Hydrocarbonoclastic bacteria; Laboratory-acquired infection; Matground; Microbial pathogenesis; Milnesium alpigenum; Mitochondrion-related organelle; Phageome; Phytoplankton microbiome; Virivore; Virome analysis; Zodletone Mountain; Glossary of cellular and molecular biology (M–Z); Agricultural weed syndrome; Cell autonomous sex identity; Codon reassignment; De novo domestication; Endemixit; Genetic map function; Hovlinc; Integrative and conjugative element; Jena Declaration; Macrosatellite; Museomics; Poison exon; Polydactyly-myopia syndrome; Red cell genotyping; Right To Know; Selection limits; Shadow effect; Transcriptome-wide association study; Tumor mutational burden; Allogeneic processed thymus tissue; Cellular anastasis; COVID-19 passports in the United Kingdom; History of phagocytosis; Immunocapitalism; Macrophage-activating lipopeptide 2; Metal allergy; Milk immunity; Myocarditis-myositis-myasthenia gravis overlap syndrome; Oligoclonal antibody; P-i mechanism; Pathogen avoidance; Peripheral ulcerative keratitis; Post-acute infection syndrome; RVT-802; T memory stem cell; Thymic mimetic cells; TMEM61; Type 2 inflammation; Vaccine passports during the COVID-19 pandemic; Vaccine resistance; Zigakibart; 2022–2023 pediatric care crisis; Acoustic epidemiology; Causal pie model; Connecting Organizations for Regional Disease Surveillance; Elimination of tuberculosis; Epidemics Act; Epidemiology in Relation to Air Travel; Epidemiology of gonorrhoea; European Society of Health and Medical Sociology; Harvard Six Cities study; Hyperendemic; Loneliness epidemic; Microbial pathogenesis; Origin tracing; Pathogenic microorganisms in frozen environments; SARS-CoV-2 in white-tailed deer; Source attribution; Sporadic disease; Outline of public health; Alcohol tax; Autobesity; Biomedical Research Center; CalOptima; Care Group approach; Christian Health Association of Malawi; Commercial determinants of health; Connecting Organizations for Regional Disease Surveillance; COVID-19 lockdowns by country; Epidemics Act; History of public health in Australia; History of public health in Canada; History of public health in Chicago; History of public health in New York City; History of public health in the United Kingdom; History of public health in the United States; Intermittent water supply; International Association for Cannabinoid Medicines; Langya virus; LGBT life expectancy; User:Lguzmang06/sandbox; Loneliness epidemic; Malawi Network of AIDS Services; Mass. and Cass; Medical officer of environmental health; Motonormativity; National Association for People living with HIV/AIDS in Malawi; North Karelia Project; Nuisance ordinance; Origin tracing; Preventive and social medicine; Responsibility Deal; SaTScan; Sleeping Sickness Commission; Slug gate; Social determinants of mental health; Special Programme of Research, Development and Research Training in Human Reproduction- HRP; Telemedicine in Nepal; Vaccine equity; Vaccine line jumping; Vaccine storage; WHO Hub for Pandemic and Epidemic Intelligence; WHO public health prizes and awards; Additive effect; Antica Farmacia Sant'Anna; FK962; Institute for Safe Medication Practices; Model-Informed Precision Dosing; P-i mechanism; Penetration enhancer; Pharmacological cardiotoxicity; Pullulan bioconjugate; Reversible Hill equation

\subsection{Training Loss Curves for Wikipedia Fine-tuning}
\label{app:training_loss_curves}
\begin{figure*}[]
    \centering
    \includegraphics[width=0.9\linewidth]{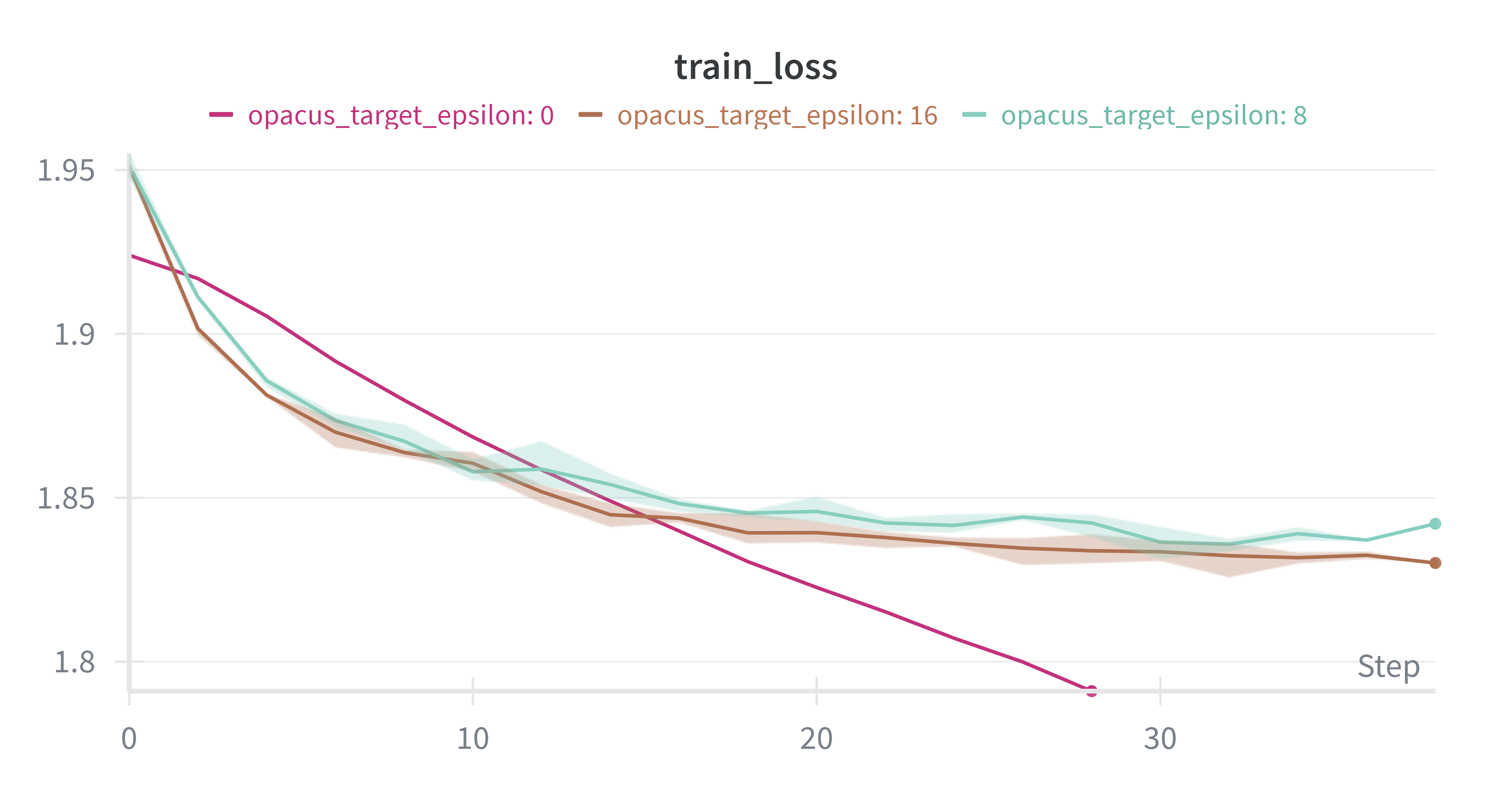}
    \caption{Training loss curve for the models fine-tuned on the large Wikipedia dataset,}
    \label{fig:wiki-large-ft-loss-full}
\end{figure*}

\begin{figure*}[]
    \centering
    \includegraphics[width=0.9\linewidth]{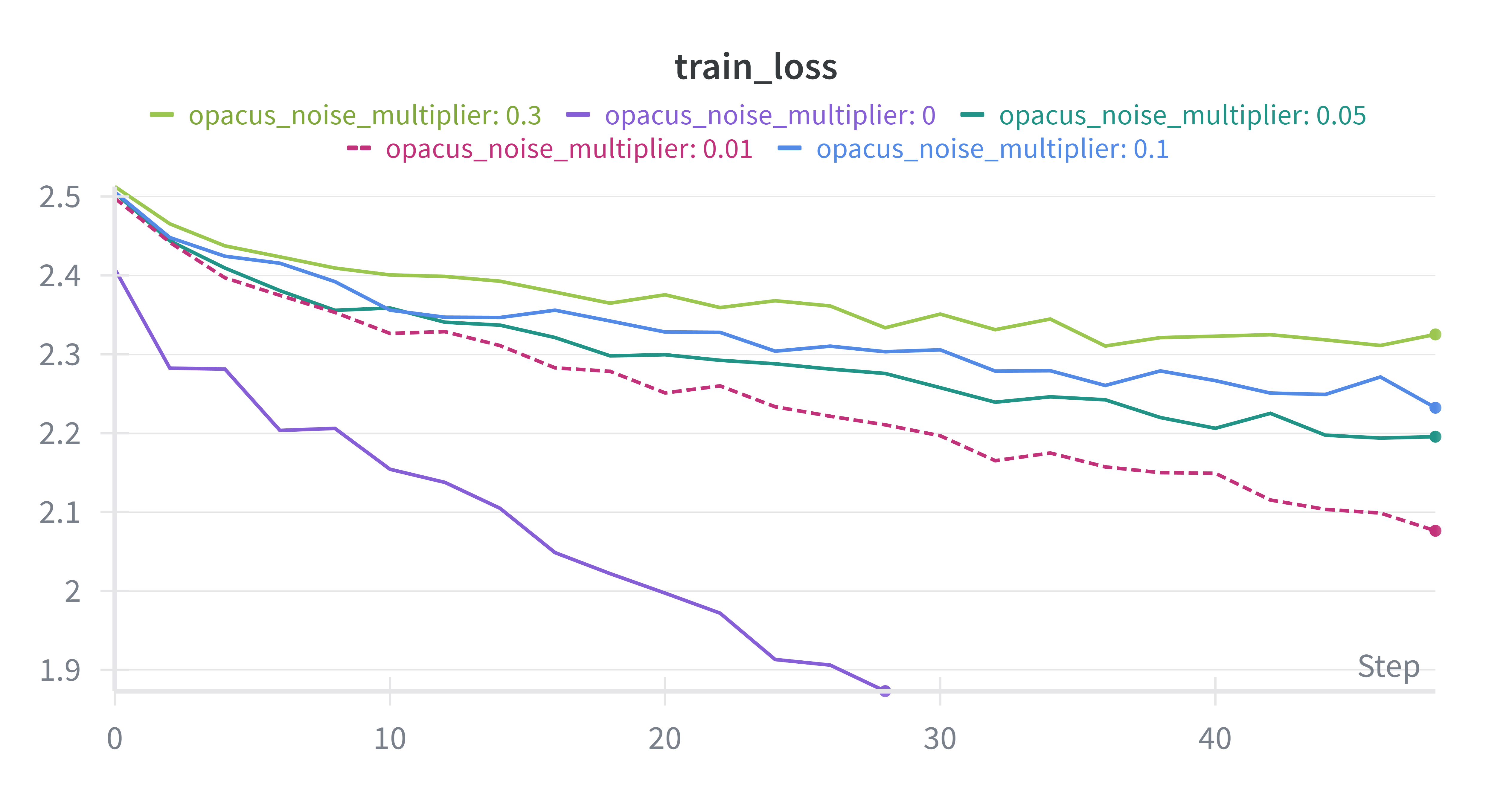}
    \caption{Training loss curve for the models fine-tuned only on the private Wikipedia articles.}
    \label{fig:wiki-large-ft-loss-pvt}
\end{figure*}

We report the training loss curves for both the non-DP and DP models trained on the Wikipedia datasets in \Cref{fig:wiki-large-ft-loss-full} and \Cref{fig:wiki-large-ft-loss-pvt}. We include results from two fine-tuning settings: (1) the setup where the unseen Wikipedia articles are interspersed with pre-training Wikipedia data, and privacy is specified through a target privacy budget $\varepsilon$; and (2) fine-tuning only over the unseen Wikipedia dataset, where we directly set the noise multiplier due to the instability in computations of noise multipliers for large $\varepsilon$.

%\vspace*{\fill}
\clearpage
%\newpage

\section{Detailed Experimental Results}
\subsection{Analysis of Recurring Hallucinations}
\label{sec:recurrent-claim}

%\textbf{[Placeholder:} Explain the approach for identifying recurring claims (e.g., clustering techniques, semantic similarity, etc.). \textbf{]}

The metrics in \Sref{sec:methods_fact_eval} can capture if models hallucinate incorrect facts, but they do not distinguish between models that output a range of incorrect information (suggesting general noisiness) or if they repeatedly output the same incorrect information across generations (suggesting encoding of inaccurate facts). To identify recurring factual claims across model generations, we propose a multi-stage clustering algorithm. We give a high-level summary below, with precise pseudo-code in \Cref{algo:claim-cluster}.

Consider the generated synthetic documents grouped by the the topic $t \in \mathcal{T}$ used to prompt their generation.
% The input is a set of topics $\mathcal{T}$ (e.g., Wikipedia article titles), where each topic $t \in \mathcal{T}$ has an associated set of synthetically generated documents. 
Each document is decomposed into atomic claims and these claims are aggregated into a claim set $C_t = {c_i}$ for each topic $t$.

We first index all extracted claims to their source (synthetic) documents, then cluster them using sentence-embedding–based agglomerative clustering to group semantically similar claims. To refine boundaries, we apply DBSCAN with Jaccard similarity, ensuring clusters are both semantically coherent and lexically consistent. This reduces cases where semantically related but factually distinct claims are grouped together.

Finally, we retain only recurring claim clusters, defined as clusters containing claims from at least two distinct documents for the same topic. This step isolates claims that recur across different generations, highlighting factual patterns the model consistently produces rather than one-off statements. 
We then analyze these clusters of claims to identify potential recurrent hallucinations.
\begin{algorithm}[t]
\caption{Recurring Claim Cluster Algorithm}
\label{algo:claim-cluster}
\begin{algorithmic}[1]
\Require Set of topics $\mathcal{T}$, where each $t$ $\in$ $\mathcal{T}$ has $\mathcal{S}$ corresponding generated documents about the topic. $C = \{C_t\}$: Claims per topic $t$ $\in$ $\mathcal{T}$, where each $C_t = \{c_i\}$ with atomic facts (supported or unsupported).
\State $\mathcal{I} \gets \textsc{IndexClaimsByText}(C, \mathcal{I})$ \Comment{Index Claims to their Source Document}
\State $\mathcal{K} \gets \textsc{ClusterAssignmentOfClaims}(\mathcal{I})$ \Comment{Sentence Embedding-based Agglomerative Clustering}
\State $\mathcal{K} \gets \textsc{DBSCANJaccardClustering}(\mathcal{K})$ \Comment{DBSCAN Clustering over clusters to ensure their Jaccard Distance is low}
\For{$t \in \mathcal{T}$}
    \State $\mathcal{K}'[t] \gets \{\}$ \Comment{Initialize $\mathcal{K}'$ to contain clusters of recurring claims}
\EndFor
\For{topic $t \in \mathcal{T}$}
    \For{Cluster $k \in \mathcal{K}(t)$}
        \If{$\textsc{COUNT}(S)$ for any $c_i \in k$ $\ge 2$}
            \State Append $k$ to $\mathcal{K}'[t]$ \Comment{Append a cluster of claims if the claims contain at least two supporting documents}
        \EndIf
    \EndFor
\EndFor
\end{algorithmic}
\end{algorithm}

\subsubsection{Recurring Hallucination Analysis}
\Cref{tab:claim-cluster} reports results from the claim clustering analysis, as described in \Sref{sec:recurrent-claim}. In both datasets, the DP model with $\epsilon=8$ outputs fewer recurring supported claims (i.e. factually correct statements) than other models. DP models also output more recurring unsupported claims than non-DP models, with $\epsilon=8$ highest for Wikipedia Science and $\epsilon=16$ highest for Wikipedia AI. For both datasets, the ratio of supported recurring claims to unsupported recurring claims is consistently lower with stricter privacy budgets. This indicates that the increased hallucinations under DP are not random. Given that the same unsupported claims recur across generations, it suggests that DP systematically shifts the model's output distribution toward factually incorrect content. 

\Cref{tab:claim-examples} shows an example of a cluster of recurring hallucinations for each model, where the prompt was ``AlphaEvolve.'' The non-DP model correctly outputs that AlphaEvolve is a model, but incorrectly describes model use. In contrast, the DP models both hallucinate that AlphaEvolve is a video game, with repeated fabricated information about the development and game play.

\begin{table*}[]
\centering
\renewcommand{\arraystretch}{1.4}
\small
\begin{tabular}{rccccccccc}
\toprule
\multicolumn{1}{l}{}                                                         & \multicolumn{1}{l}{} & \multicolumn{3}{c}{\textbf{\textit{Supported}}}                                                                                                                                                                             & \multicolumn{1}{l}{} & \multicolumn{3}{c}{\textbf{\textit{Unsupported}}}                                                                                                                                                                           & \multicolumn{1}{l}{}                                                      \\ \cline{3-5} \cline{7-9}
\textbf{}                                                                    & \textbf{DP Setting}  & \textbf{\begin{tabular}[c]{@{}c@{}}Recur\\ Count\end{tabular}} & \textbf{\begin{tabular}[c]{@{}c@{}}Total\\ Count\end{tabular}} & \textbf{\begin{tabular}[c]{@{}c@{}}\% Avg\\ Recur\end{tabular}} &                      & \textbf{\begin{tabular}[c]{@{}c@{}}Recur\\ Count\end{tabular}} & \textbf{\begin{tabular}[c]{@{}c@{}}Total\\ Count\end{tabular}} & \textbf{\begin{tabular}[c]{@{}c@{}}\% Avg\\ Recur\end{tabular}} & \textbf{\begin{tabular}[c]{@{}c@{}}Supported:\\ Unsupported\end{tabular}} \\
\midrule
\multirow{3}{*}{\begin{tabular}[c]{@{}r@{}}Wikipedia\\ AI\end{tabular}}      & $\epsilon=\infty$ & 242 & 1198 & 20.20 & & 494 & 1992 & 24.80 & 0.490 \\
& $\epsilon=16$ & 200 & 1103 & 18.13 & & \textbf{523} & 2002 & \textbf{26.12} & 0.382 \\
& $\epsilon=8$ & 174 & 1019 & \textbf{17.08} & & 509 & 2011 & 25.31 & 0.342 \\
 & $\textsc{AnaDP}_{\epsilon=16}$ & 207 & 1025 & 20.20 & & 452 & 1915 & 23.60 & 0.458 \\
& $\textsc{AnaDP}_{\epsilon=8}$ & \textbf{161} & 932 & 17.27 & & 499 & 1922 & 25.96 & \textbf{0.323} \\
 & Task-tuned & \textbf{161} & 920 & 17.50 & & 460 & 1945 & 23.65 & 0.350 \\ \midrule
\multirow{3}{*}{\begin{tabular}[c]{@{}r@{}}Wikipedia\\ Science\end{tabular}} & $\epsilon=\infty$ & 705 & 2896 & 24.34 & & 560 & 2417 & 23.17 & 1.259 \\
& $\epsilon=16$ & 582 & 2467 & 23.59 & & 556 & 2128 & 26.13 & 1.047 \\
& $\epsilon=8$ & 556 & 2378 & 23.38 & & \textbf{633} & 2222 & \textbf{28.49} & \textbf{0.878} \\
& $\textsc{AnaDP}_{\epsilon=16}$ & 567 & 2372 & 23.90 & & 536 & 2065 & 25.96 & 1.058 \\
& $\textsc{AnaDP}_{\epsilon=8}$ & 544 & 2287 & 23.79 & & 509 & 2128 & 23.92 & 1.069 \\
& Task-tuned & \textbf{474} & 2258 & \textbf{20.99} & & 535 & 2002 & 26.72 & 0.886 \\
\bottomrule
\end{tabular}
\caption{\small
Analysis of recurrent claims and hallucinations (\Sref{sec:recurrent-claim})
% Claim clustering analysis 
for temperature $\tau=0.3$. ``Total Count'' reports the total number of generated clusters, ``Recur Count'' reports the number of those clusters with $\ge2$ supporting documents, and \%Avg Recur is ``Recur Count''/``Total Count''. The far right column reports  ``Recur Count'' of supported claims / ``Recur Count'' of unsupported claims. DP models output a lower ratio of repeated supported claims  to unsupported claims, suggesting increased repeated hallucinations. Bolding indicates most hallucinations (i.e. least supported or most unsupported). Task-tuned indicates the base model trained on pretraining data for 1 epoch.}
\label{tab:claim-cluster}
\end{table*}

\begin{table*}[]
    \centering
    \small
    \begin{tabular}{lp{5.75in}}
    \toprule
        $\epsilon=\infty$ &  `AlphaEvolve is used for generating the molecular structures of organic molecules.', `AlphaEvolve is used for generating molecular structures.' \\
        \midrule
        $\epsilon=16$  &  `There are two types of enemies in the game.', `The game features two types of enemies.' \\
        \midrule
        $\epsilon=8$ & `Black Hole Interactive is a game development company.', `Black Hole Interactive is a video game development company.'\\
        \bottomrule
    \end{tabular}
    \caption{\small Example unsupported claim clusters (hallucinations) for each model. We provide additional examples in \autoref{app:example_claims}.}
    \label{tab:claim-examples}
\end{table*}

%\newpage
\clearpage
\subsection{Statistically Significant Differences in FactScore}\label{app:statistical_significance_testing}

We report the statistical significance of the FactScore differences in this section. We compare responses only for topics that have a usable generation with a corresponding FactScore. A topic is only taken into consideration if the configurations being compared cover it. For the DP fine-tuning comparisons, we take one intersection across $\varepsilon\in\{8,16,\infty\}$, which results in $221/222$ Science, $118/124$ AI topics for the topic-level differences. For the pre-trained models, we take the pairwise intersection of topics (between VaultGemma and the pre-trained model being compared against) for which there are valid generations, which results in $198$--$213$ (Science) and $103$--$118$ (AI) topics. The response-level tests use all responses from the model configuration, irrespective of topics.

\subsubsection{Pre-trained models}

\begin{table}[H]
\centering
\small
\setlength{\tabcolsep}{3pt}
\resizebox{\columnwidth}{!}{%
\begin{tabular}{lcccccc}
\toprule
& \multicolumn{3}{c}{Wiki-Science} & \multicolumn{3}{c}{Wiki-AI} \\
\cmidrule(lr){2-4}\cmidrule(lr){5-7}
Model & $\Delta$ & $p_{W}$ & $p_{t}$ & $\Delta$ & $p_{W}$ & $p_{t}$ \\
\midrule
Gemma-2-2B & $+.111$ & $<.001^{***}$ & $<.001^{***}$ & $+.055$ & $.035^{*}$ & $.025^{*}$ \\
Gemma-2B   & $+.138$ & $<.001^{***}$ & $<.001^{***}$ & $+.084$ & $<.001^{***}$ & $<.001^{***}$ \\
Gemma-3    & $+.143$ & $<.001^{***}$ & $<.001^{***}$ & $+.100$ & $<.001^{***}$ & $<.001^{***}$ \\
GPT2-XL    & $+.038$ & $.028^{*}$ & $.033^{*}$ & $-.140$ & $<.001^{***}$ & $<.001^{***}$ \\
\bottomrule
\end{tabular}}
\caption{Topic-level paired tests against VaultGemma, on the worst
(minimum) response per topic. $\Delta$: mean paired FActScore difference
(model $-$ VaultGemma); positive favors the model. $p_W$: Wilcoxon
signed-rank; $p_t$: paired $t$-test. Topics are excluded where either model produced no scorable generation (empty or degenerate output yielding no atomic claims), with varying $n$ of the range $198$--$201$ topics (Science), $103$--$116$ (AI). $^{*}p<.05$, $^{***}p<.001$.}
\label{tab:pretrain-min-wilcoxon-ttest}
\end{table}

\begin{table}[H]
\centering
\small
\setlength{\tabcolsep}{4pt}
\begin{tabular}{lcccc}
\toprule
& \multicolumn{2}{c}{Wiki-Science} & \multicolumn{2}{c}{Wiki-AI} \\
\cmidrule(lr){2-3}\cmidrule(lr){4-5}
Model & $\Delta$ & $p_{t}$ & $\Delta$ & $p_{t}$ \\
\midrule
Gemma-2-2B  & $+.108$ & $3.5$e${-}14^{***}$ & $+.062$ & $6.0$e${-}4^{***}$ \\
Gemma-2B    & $+.119$ & $5.2$e${-}17^{***}$ & $+.064$ & $3.1$e${-}4^{***}$ \\
Gemma-3     & $+.109$ & $3.1$e${-}12^{***}$ & $+.087$ & $1.4$e${-}5^{***}$ \\
GPT2-XL     & $-.005$ & $.752$ & $-.154$ & $8.0$e${-}17^{***}$ \\
\bottomrule
\end{tabular}
\caption{Welch two-sample $t$-tests against VaultGemma for the pretrained
baselines, over every generated response. $\Delta$: mean response-level
FActScore difference (model $-$ VaultGemma); positive favors the model.}
\label{tab:response-all-welch-pretrain}
\end{table}

\subsubsection{Fine-tuned models}

\begin{table}[H]
\centering
\small
\setlength{\tabcolsep}{3pt}
\begin{tabular}{lcccccc}
\toprule
& \multicolumn{3}{c}{Wiki-Science} & \multicolumn{3}{c}{Wiki-AI} \\
\cmidrule(lr){2-4}\cmidrule(lr){5-7}
$\varepsilon$ & $\Delta$ & $p_{W}$ & $p_{t}$ & $\Delta$ & $p_{W}$ & $p_{t}$ \\
\midrule
$8$  & $-.032$ & $.026^{*}$ & $.030^{*}$ & $-.048$ & $.009^{**}$ & $.007^{**}$ \\
$16$ & $-.035$ & $.005^{**}$ & $.018^{*}$ & $-.033$ & $.032^{*}$ & $.079$ \\
\bottomrule
\end{tabular}
\caption{Topic-level paired tests against the non-private baseline
($\varepsilon=\infty$), on the worst
(minimum) response per topic. $\Delta$: mean paired FActScore difference
(DP model $-$ non-private); positive favors the model. $p_W$: Wilcoxon
signed-rank; $p_t$: paired $t$-test. Topics are excluded where either model produced no scorable generation (empty or degenerate output yielding no atomic claims), with $n=221$ (Science), $n=118$ (AI). $^{*}p<.05$, $^{**}p<.01$, $^{***}p<.001$.}
\label{tab:wilcoxon-ttest-mean-finetuned}
\end{table}

\begin{table}[H]
\centering
\small
\setlength{\tabcolsep}{4pt}
\begin{tabular}{lcccc}
\toprule
& \multicolumn{2}{c}{Wiki-Science} & \multicolumn{2}{c}{Wiki-AI} \\
\cmidrule(lr){2-3}\cmidrule(lr){4-5}
$\varepsilon$ & $\Delta$ & $p_{t}$ & $\Delta$ & $p_{t}$ \\
\midrule
$8$         & $-.026$ & $.024^{*}$ & $-.061$ & $2.4$e${-}5^{***}$ \\
$16$        & $-.025$ & $.033^{*}$ & $-.038$ & $.009^{**}$ \\
\bottomrule
\end{tabular}
\caption{Response-level tests against $\varepsilon=\infty$, treating each
generation as an observation. $\Delta$: mean FActScore difference
(model $-$ non-private); negative favors non-private. $p_t$: Welch two-sample
$t$-test. $^{*}p<.05$, $^{**}p<.01$, $^{***}p<.001$.}
\label{tab:response-welch-test}
\end{table}

\subsection{Generation Length Statistics}

\begin{table}[h]
\centering
\small
\setlength{\tabcolsep}{4pt}
\begin{tabular}{@{}llcrr@{}}
\toprule
Category & $\varepsilon$ & Facts & \multicolumn{2}{c}{Length (words)} \\
\cmidrule(lr){5-5}\cmidrule(lr){4-5}
 & & /Resp. & Mean & Std. \\
\midrule
\multirow{3}{*}{Wiki.\ AI}
  & 8        & 10.24 & 80.41 & 21.16 \\
  & 16       & 10.60 & 82.00 & 21.42 \\
  & $\infty$ & 11.12 & 86.64 & 17.47 \\
\midrule
\multirow{3}{*}{Wiki.\ Science}
  & 8        &  8.78 & 79.14 & 19.56 \\
  & 16       &  8.66 & 79.88 & 18.91 \\
  & $\infty$ &  9.75 & 88.75 & 14.79 \\
\bottomrule
\end{tabular}
\caption{Average number of facts and response length by category and privacy budget $\varepsilon$.}
\label{tab:facts-length}
\end{table}

We report the generation-length statistics to rule out any potential length-related artifacts in our FactScore evaluations in Table~\ref{tab:facts-length}. While DP-generated text is slightly shorter on average than the text from the non-private models, consistent with prior work \cite{cano2025differentiallyprivatetextgenerationdegrades}, this difference is not large enough to warrant a significant shift in the FactScore (which already accounts for length by normalizing by the number of claims). Likewise, we show that the density of claims is stable across privacy budgets, so our reported difference in FactScores are not attributable to the length of the generation.

\subsection{Lower Perplexity Does Not Correlate With Factual Reliability}
\label{app:hallucinations_vs_perplexity}

\begin{figure}[h]
    \centering
    \begin{subfigure}{0.45\textwidth}
        \centering
        \includegraphics[width=\linewidth]{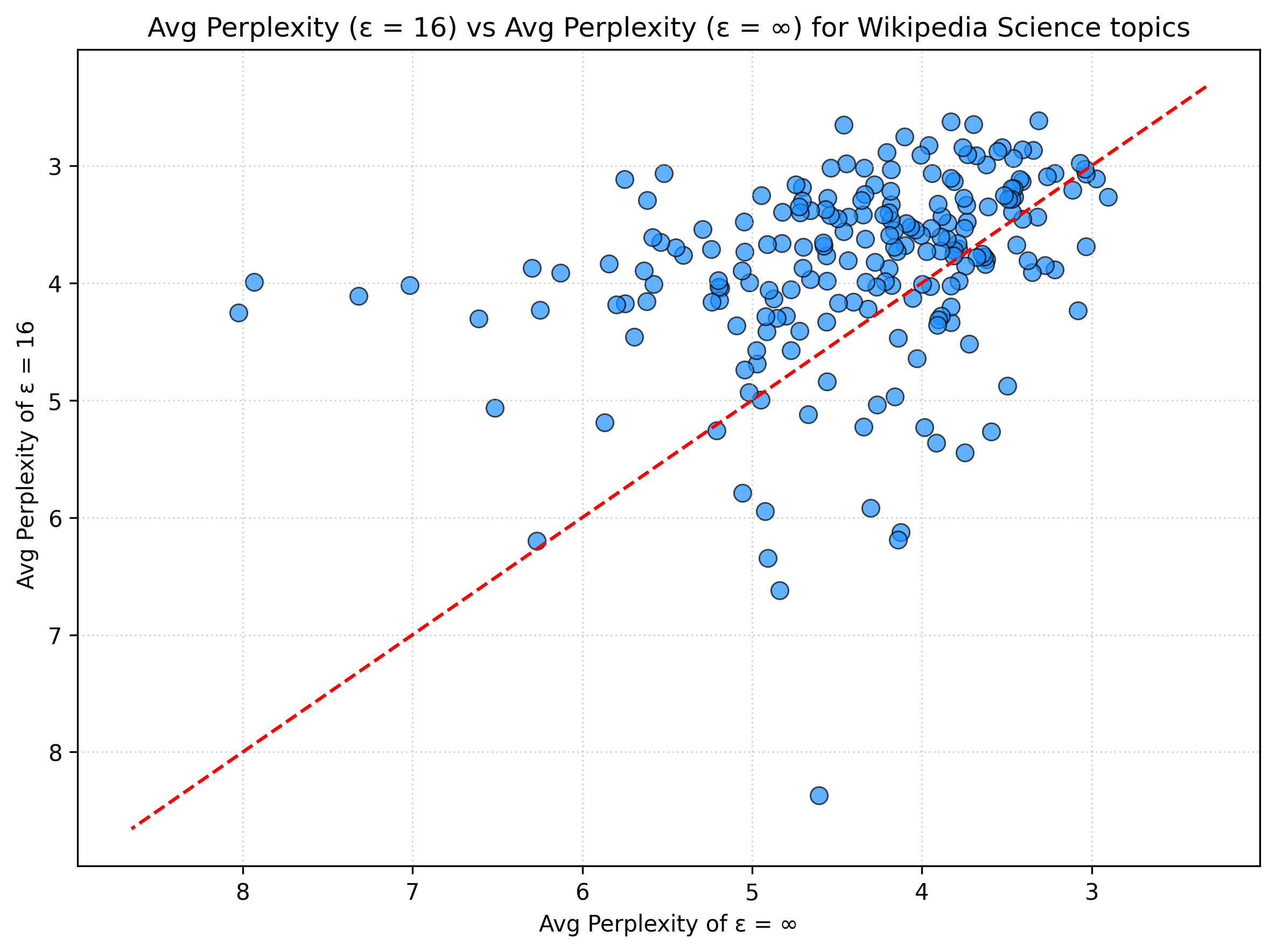}
        \caption{Avg perplexity of each topic for  $\varepsilon = 16$ vs $\varepsilon = \infty$.}
    \end{subfigure}
    \\
    \hspace{0.01\textwidth}
    \begin{subfigure}{0.45\textwidth}
        \centering
        \includegraphics[width=\linewidth]{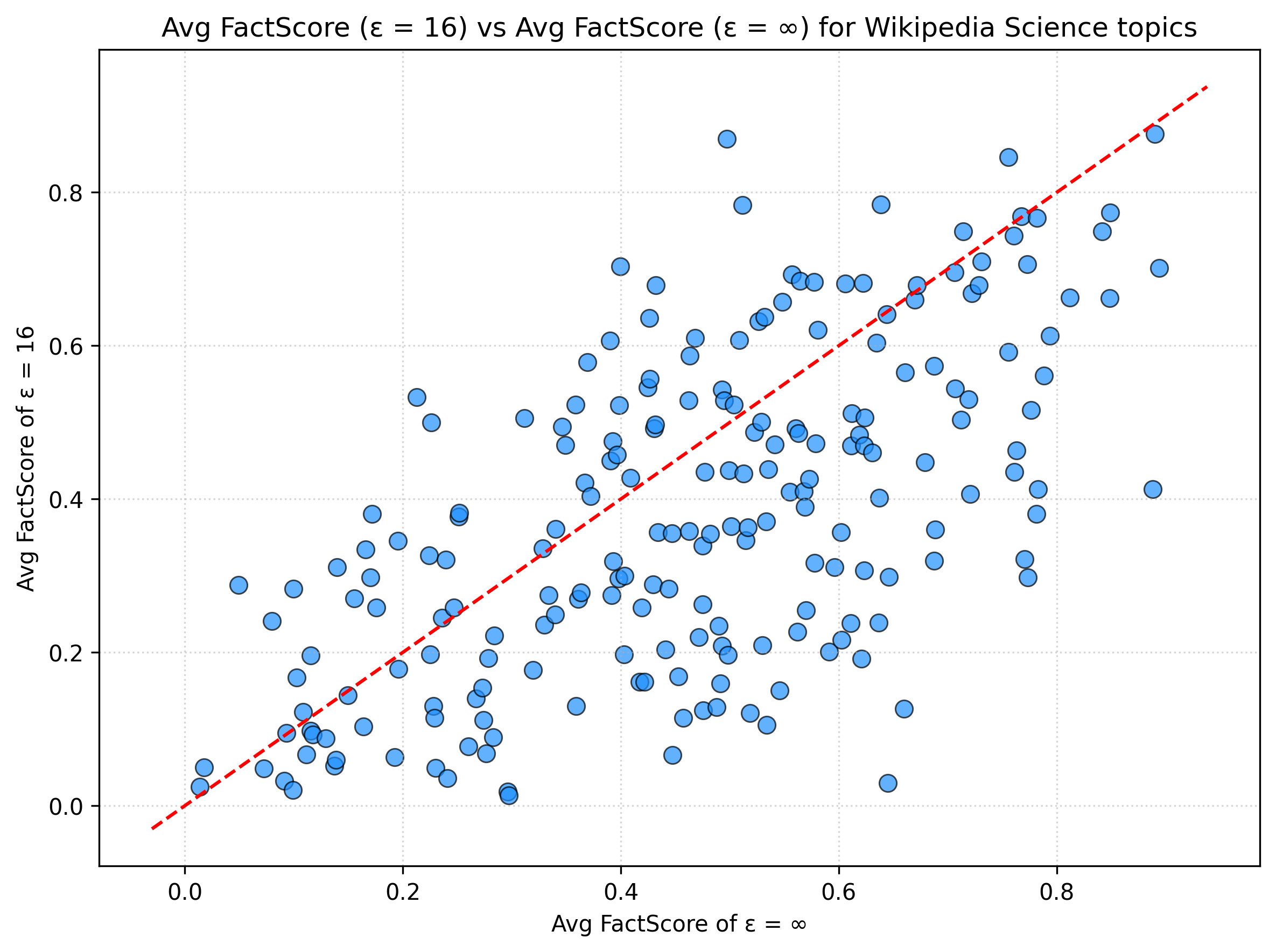}
        \caption{Avg FactScore of each topic for  $\varepsilon = 16$ vs $\varepsilon = \infty$ }
    \end{subfigure}
    \caption{Average perplexity and average FactScore per topic across models. DP-finetuned models often achieve lower average perplexity  (points above the y = x line), yet the non-DP model attains higher average FactScores across topics—showing that lower perplexity does not imply better factuality.}
    \label{fig:factscore_and_perplexity_graphs}
\end{figure}

\begin{figure}[h]
    \centering
    \begin{subfigure}{0.45\textwidth}
        \centering
        \includegraphics[width=\linewidth]{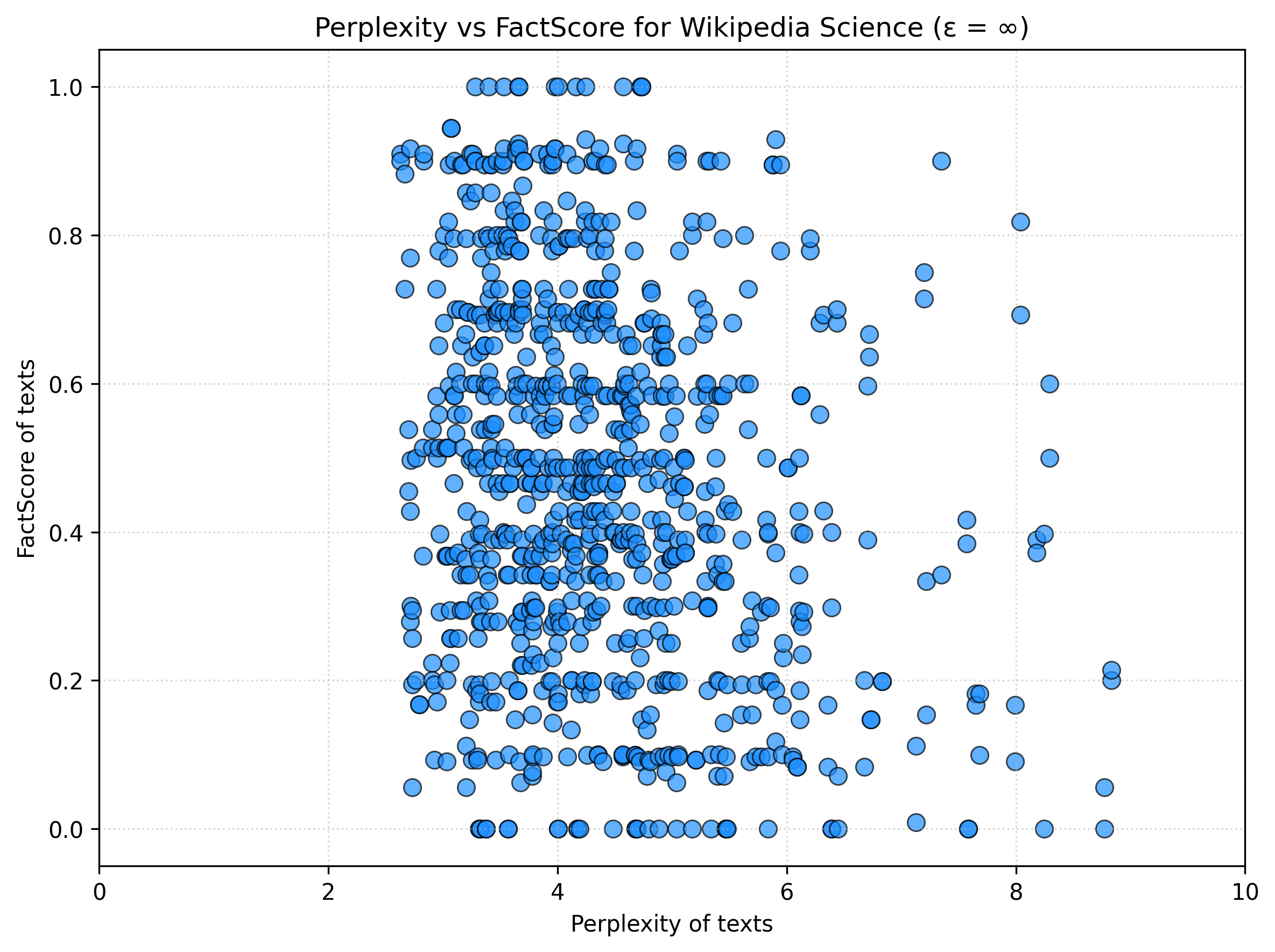}
        \caption{$\varepsilon = \infty$}
       % \label{fig:kde1}
    \end{subfigure}
    \\
    \hspace{0.01\textwidth}
    \begin{subfigure}{0.45\textwidth}
        \centering
        \includegraphics[width=\linewidth]{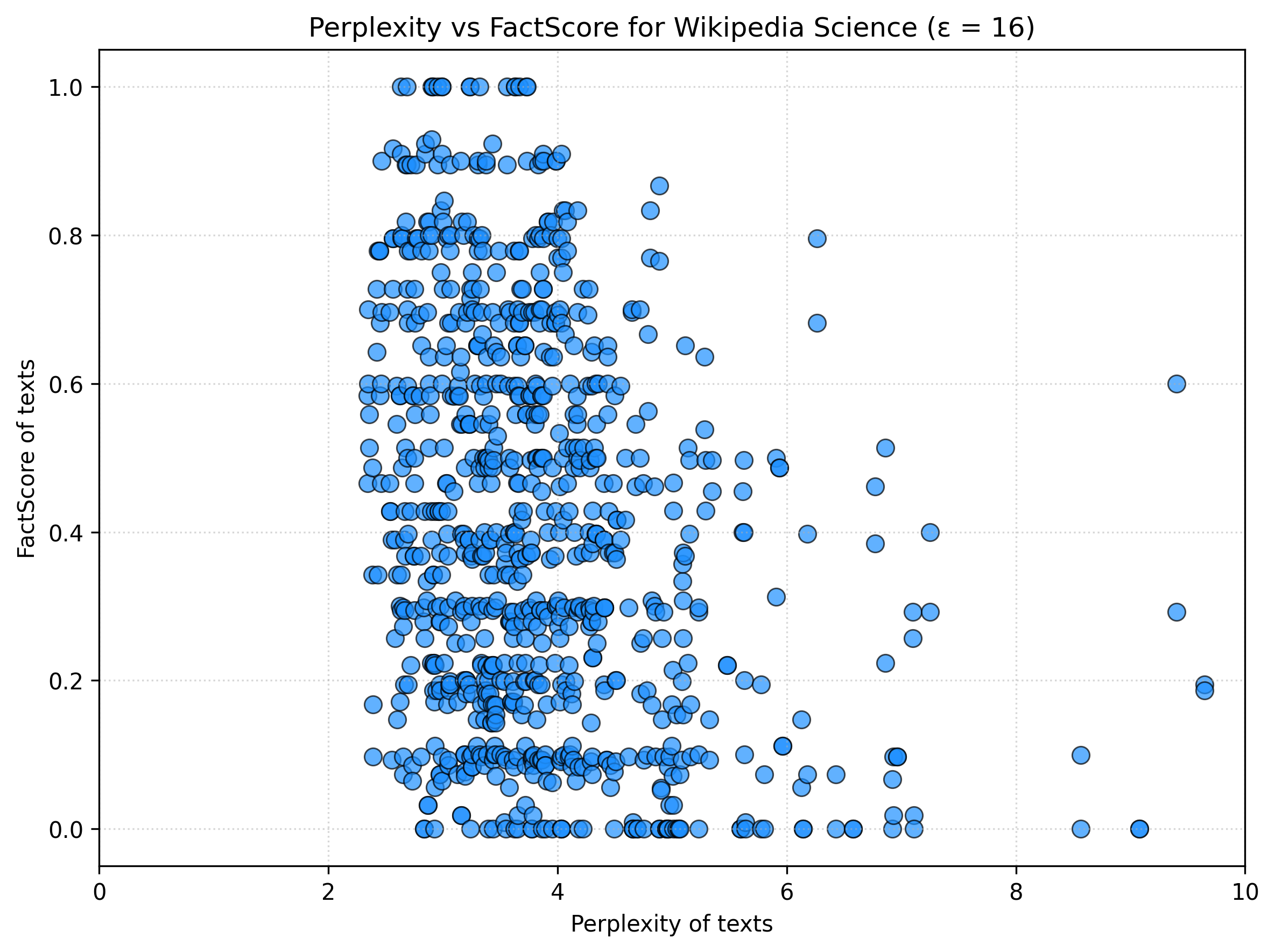}
        \caption{$\varepsilon = 16$}
       % \label{fig:kde2}
    \end{subfigure}
    \\
    \hspace{0.01\textwidth}
    \begin{subfigure}{0.45\textwidth}
        \centering
        \includegraphics[width=\linewidth]{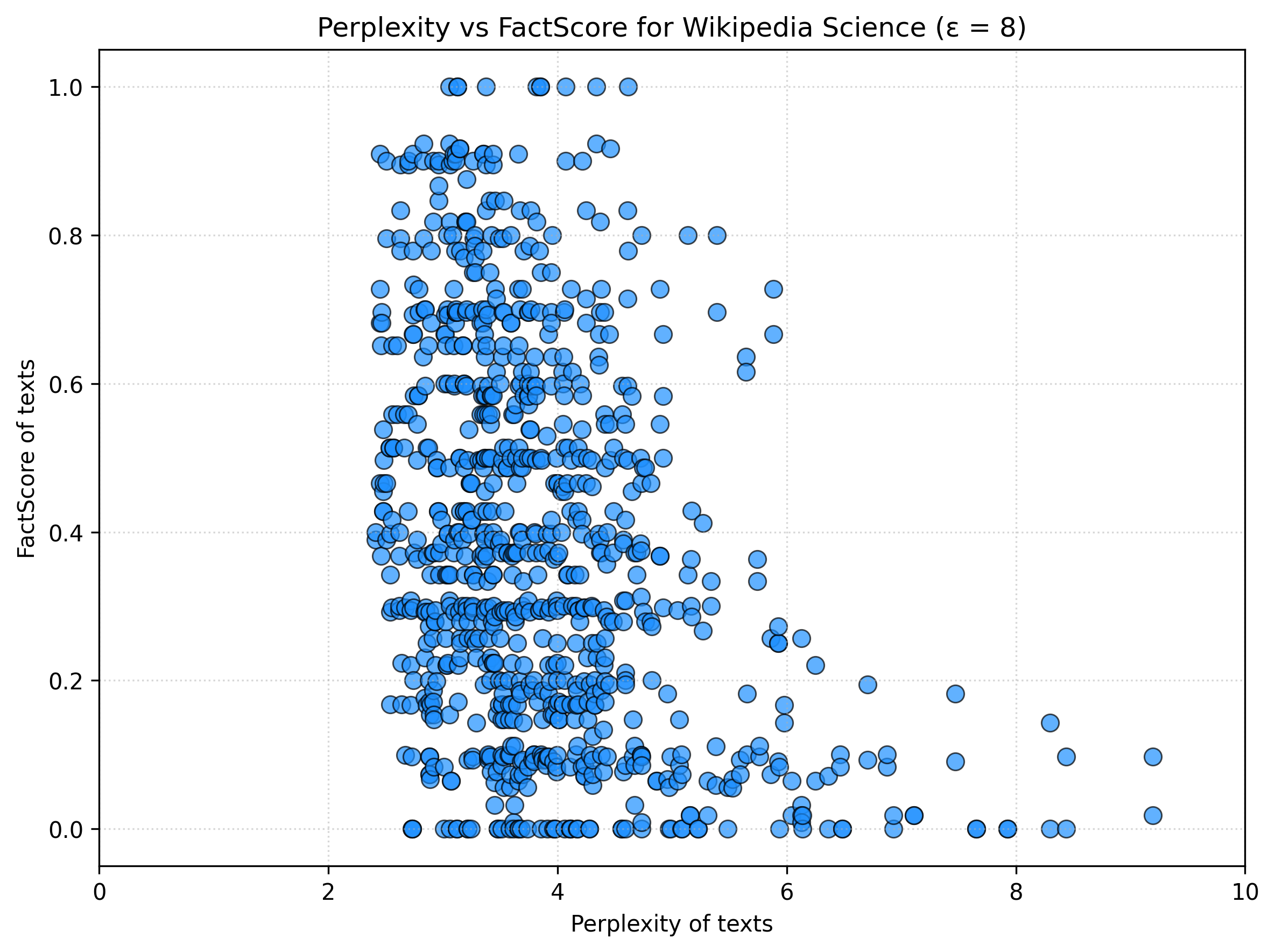}
        \caption{$\varepsilon = 8$}
        % \label{fig:kde3}
    \end{subfigure}
    \hspace{0.01\textwidth}
    \caption{Relationship between model perplexity and FactScore for all texts generated for topics from the Wikipedia Science. Lower perplexity is not predictive of higher factual accuracy.
    }
    \label{fig:perp_vs_factscore}
\end{figure}

Our analysis of hallucinations from language model outputs is a construct that is distinct from and cannot be conflated with generation notions of utility, as depicted in  \Cref{fig:perp_vs_factscore} and \Cref{fig:factscore_and_perplexity_graphs}, which demonstrates the lack of correlation between perplexity and factual correctness. Several generations with low perplexity receive low FactScores, and vice versa. The DP fine-tuned models often produce text with lower perplexity on average, further underscoring that other utility metrics do not necessarily capture hallucinations, that is to say, a model can generate highly fluent text (low perplexity) and also hallucinate. We have also included training loss curves in \Cref{app:training_loss_curves}, which demonstrates that the models we evaluate do learn to fit the training data.

\clearpage
\newpage
\subsection{Hallucination of pre-training facts in DP fine-tuned models}
\label{app:hallucination_pretraining_facts_dp}

Our experiments also look to address whether DP fine-tuning degrades the knowledge already encoded during standard pre-training. \Cref{tab:factscore_seen} reports FactScores over Wikipedia pre-training. The differences between DP and non-DP models are marginal, suggesting DP finetuning does not disrupt factual knowledge acquired from pre-training data.
This trend is consistent across temperatures (\Cref{fig:kde_factscore_wiki_pretrain}) and stands in stark contrast to the evaluations on previously unseen data, where stronger privacy constraints correlate with lower factual accuracy.

\subsection{Distributional flattening for pre-training data}\label{app:flattening-pretraining}

\begin{figure}[H]
    \centering
    \begin{subfigure}{0.23\textwidth}
        \centering
        \includegraphics[width=\linewidth]{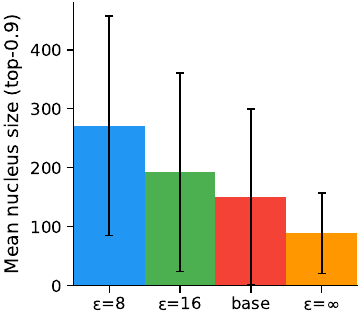}
        \caption{\footnotesize{Wikipedia Pretraining}}
    \end{subfigure}
    \caption{Mean nucleus size (top-$p = 0.9$) for models fine-tuned under different DP budgets. DP models exhibit larger nucleus sizes, indicating greater dispersion of probability mass across tokens and increased next-token uncertainty.}
    \label{fig:nucleus sizes-wiki-pretrain}
\end{figure}

\begin{figure}[H]
\centering
\hspace{0.01\textwidth}
\begin{subfigure}{0.45\textwidth}
\centering
\includegraphics[width=\linewidth]{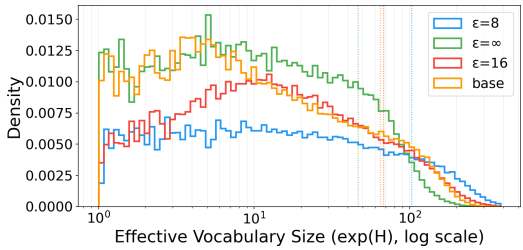}
\caption{\footnotesize{Wikipedia Pretraining}}
%\caption{\footnotesize{Avg. FS of each topic for $\varepsilon = 16$ vs $\varepsilon = \infty$ }}
\end{subfigure}
\caption{Distribution of effective vocabulary size ($\exp(H)$), where higher values indicate probability mass spread across more tokens. DP models show higher effective vocabulary sizes, reflecting increased distributional uncertainty.}
\label{fig:eff-vocabulary-size-wiki-pretrain}
\end{figure}

Although DP models exhibit higher entropy for both the unseen fine-tuning and the Wikipedia pre-training data (\Cref{fig:nucleus sizes-wiki-pretrain} and \Cref{fig:eff-vocabulary-size-wiki-pretrain}), the distributional flattening does not affect hallucinations in the same way across settings. An analysis of the nucleus threshold overlap, where we compare the top-$p$ nucleus sets of the DP and non-DP models, indicates that the ranking of tokens is preserved for the seen data despite the distribution flattening. On the nucleus-overlap metric, the $\varepsilon=16$ model attains a mean Jaccard similarity of $0.707 \pm 0.071$ on seen data and $0.623 \pm 0.051$ on unseen data, while the $\varepsilon=8$ model achieves $0.704 \pm 0.069$ (seen) and $0.619 \pm 0.051$ (unseen). The models share greater token overlap on seen pre-training data than on unseen data. These results suggest that on data seen during pre-training, the increased entropy reflects a more diffused but still structurally intact next-token probability distribution. On the other hand, for the unseen data, the redistributed probability mass falls on incorrect alternatives, leading to the observed increase in hallucinations.

\begin{table*}[]
\centering
\renewcommand{\arraystretch}{1.2}
\small
\begin{tabular}{@{}lrrrrrrrr@{}}
\toprule
 \multicolumn{1}{c}{\textbf{\begin{tabular}[c]{@{}c@{}}DP Budget \end{tabular}}} & \multicolumn{1}{c}{\textbf{\begin{tabular}[c]{@{}c@{}}Average \\ FS\end{tabular}}} & \multicolumn{1}{c}{\textbf{\begin{tabular}[c]{@{}c@{}}Median \\ FS\end{tabular}}} & \multicolumn{1}{c}{\textbf{\begin{tabular}[c]{@{}c@{}}Q1 \\ FS\end{tabular}}} & \multicolumn{1}{c}{\textbf{\begin{tabular}[c]{@{}c@{}}Q3 \\ FS\end{tabular}}} & \multicolumn{1}{c}{\textbf{\begin{tabular}[c]{@{}c@{}}Avg Max\\ FS / Topic\end{tabular}}} & \multicolumn{1}{c}{\textbf{\begin{tabular}[c]{@{}c@{}}Avg Min \\ FS / Topic\end{tabular}}} & \multicolumn{1}{c}{\textbf{\begin{tabular}[c]{@{}c@{}}\# of \\ FS $\ge$0.5 \end{tabular}}} & \multicolumn{1}{c}{\textbf{\begin{tabular}[c]{@{}c@{}}\# of \\ FS $<$0.5 \end{tabular}}} \\ \midrule
$\epsilon=\infty$ & 30.4 & 23.5 & 9.1 & 46.2 & 42.5 & 18.8 & 131 & \textbf{841} \\
$\epsilon=16$ & \textbf{29.5} & \textbf{23.1} & 9.1 & \textbf{44.4} & 41.8 & \textbf{18.1} & \textbf{125} & 829 \\
$\epsilon=8$ & 30.6 & 25.0 & \textbf{8.3} & 50.0 & 43.2 & 18.8 & 144 & 824 \\
Base & 29.7 & 25.0 & 9.1 & \textbf{44.4} & \textbf{41.0} & 19.6 & 139 & 837 \\
\bottomrule
\end{tabular}
\caption{\small
FactScores  (FS; reported in \%) for GPT-J evaluated with temperature $\tau=0.3$ over Wikipedia pre-training, which contains articles likely to be in pre-training data, but not included in fine-tuning. DP models perform similarly as non-DP models, suggesting no disruption to facts learned in pre-training. Bolding indicates worse FactScores (i.e. higher hallucinations).}
\label{tab:factscore_seen}
\end{table*}

\begin{figure}[h]
  \centering
  \includegraphics[width=\linewidth]{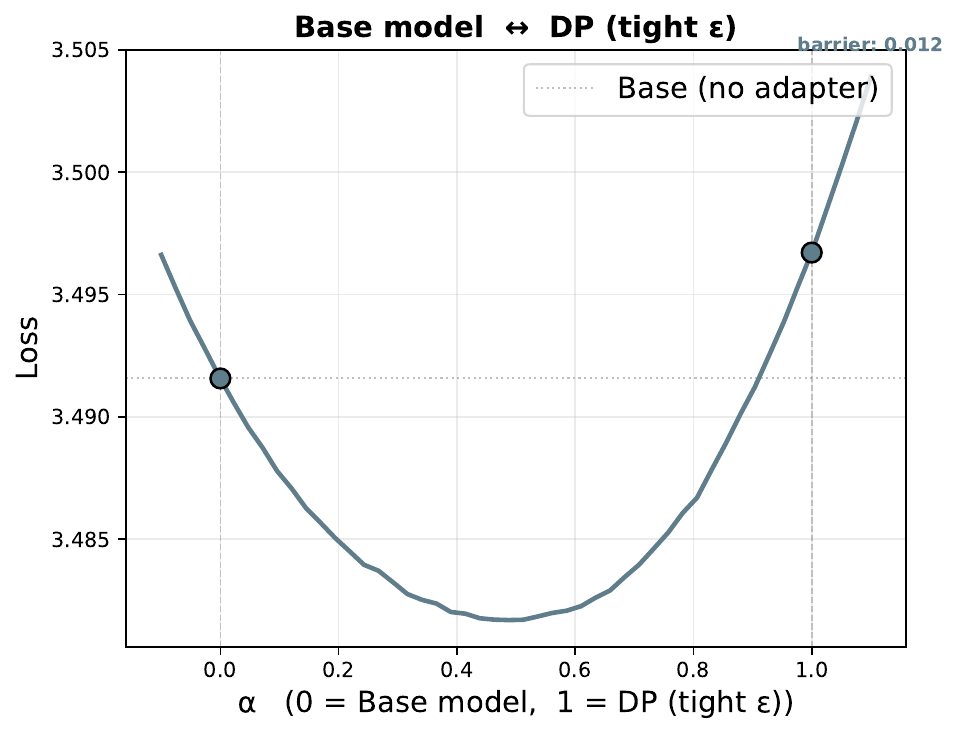}\\[6pt]
  \includegraphics[width=\linewidth]{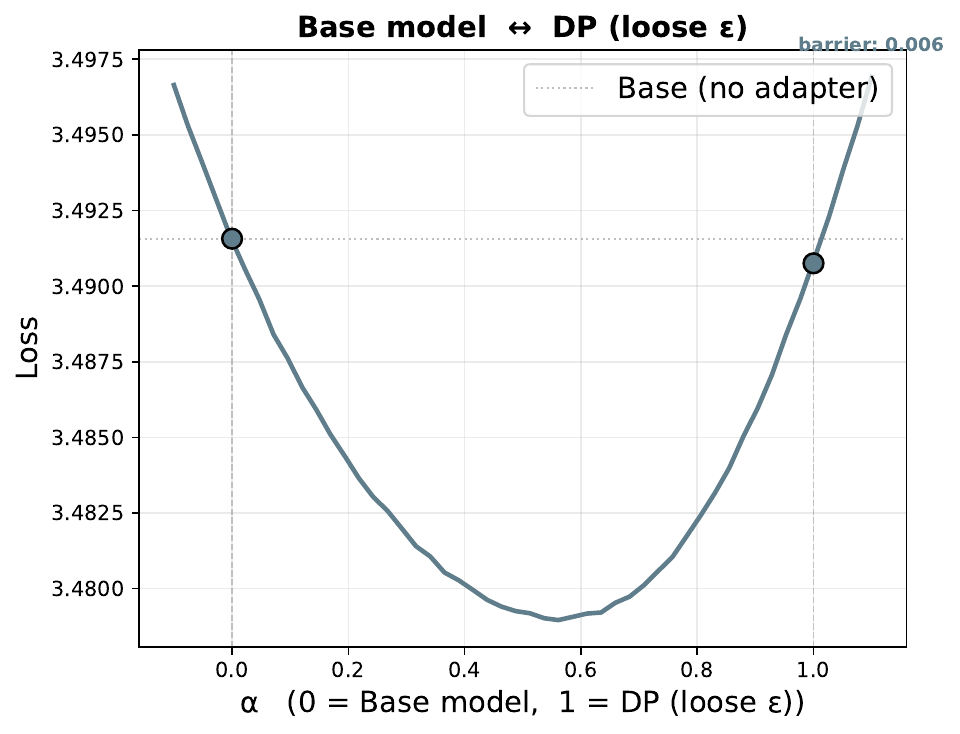}\\[6pt]
  \includegraphics[width=\linewidth]{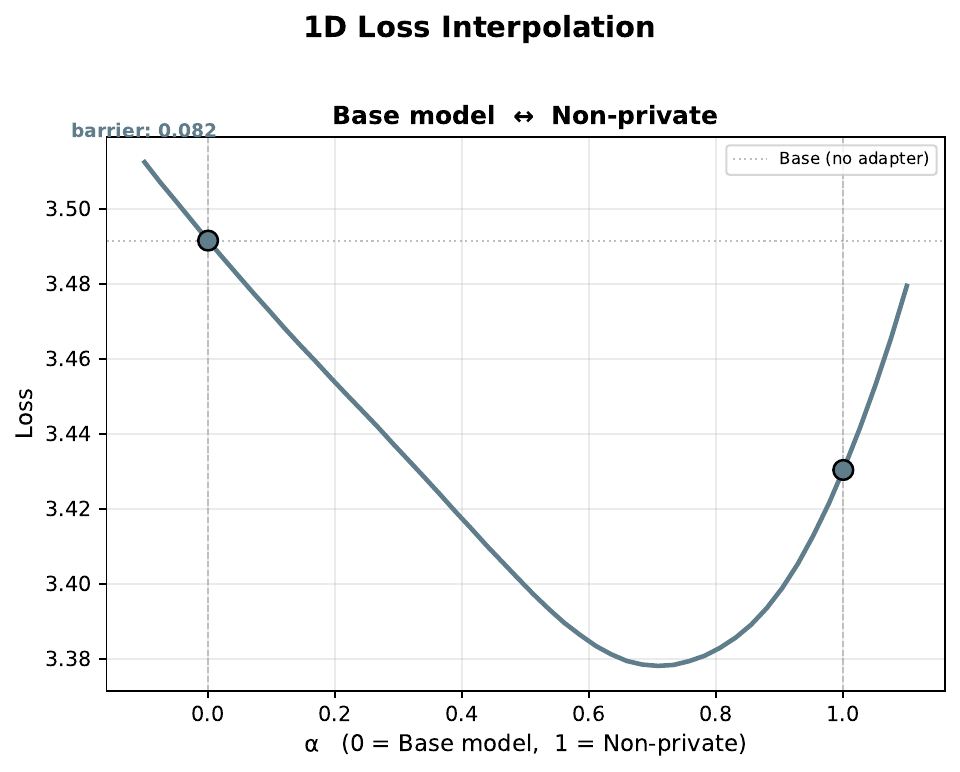}
  \caption{1D loss interpolation between the base GPT-J 6B model ($\alpha = 0$) and fine-tuned models ($\alpha = 1$) under different privacy settings. The dashed line indicates the loss of the base model without an adapter. DP models ($\varepsilon=8$ and $\varepsilon=16$) remain relatively close to the base model in parameter space, with small loss barriers (0.012 and 0.006, respectively), while the non-private model shifts significantly further (barrier of 0.082), reflecting a stronger gradient signal during fine-tuning. Despite remaining closer to the base model, the DP models sit in regions of lower loss, indicating that they learn from the fine-tuning data rather than reverting to pre-training priors.}
  \label{fig:loss_landscape}
\end{figure}

\subsection{Is the model reverting to its pre-training priors?}\label{app:reverting_to_priors}
\begin{figure*}[]
    \centering
    \begin{subfigure}{\textwidth}
        \centering
        \includegraphics[width=\linewidth]{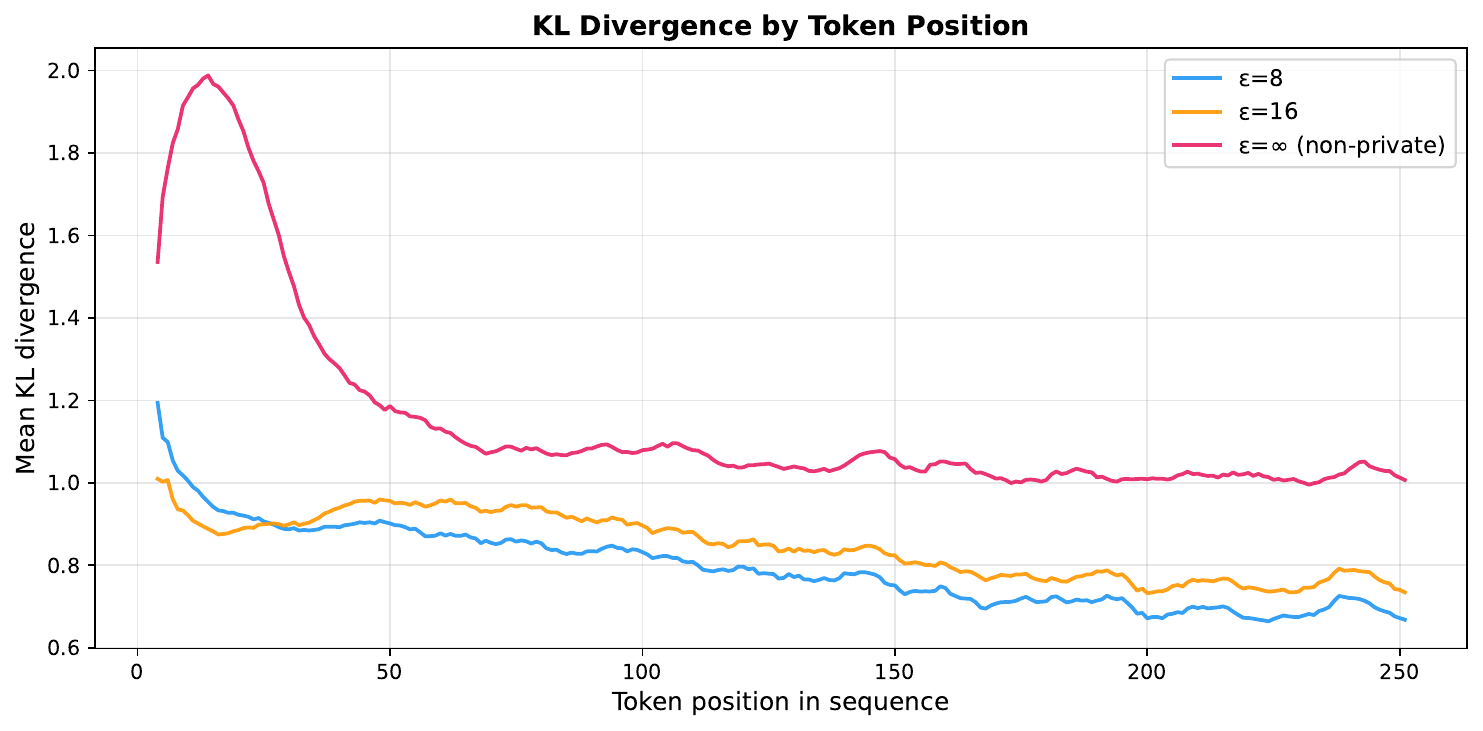}
        \caption{Wikipedia science and AI}
    \end{subfigure}
    \\
    \hspace{0.01\textwidth}
    \begin{subfigure}{\textwidth}
        \centering
        \includegraphics[width=\linewidth]{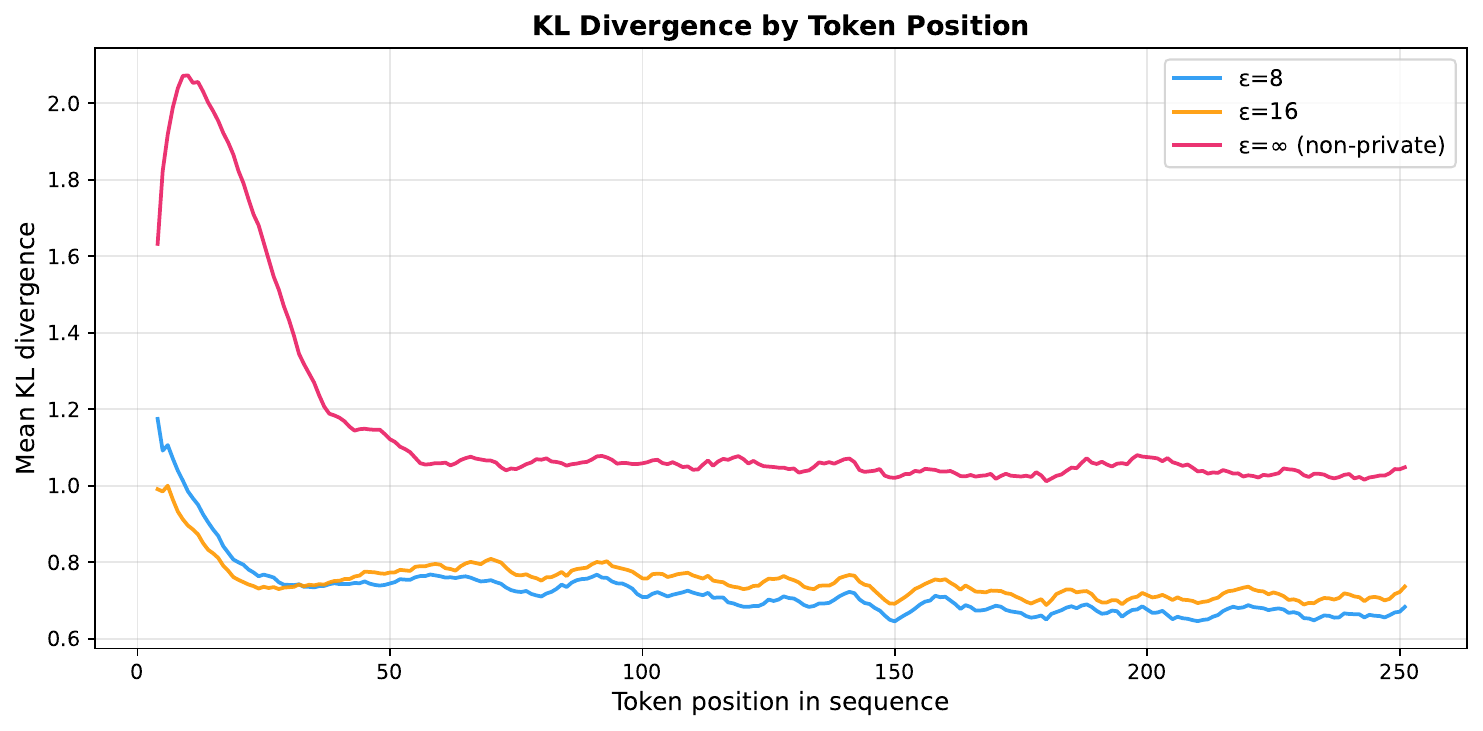}
        \caption{Wikipedia pretraining}
        %\caption{\footnotesize{Avg. FS of each topic for  $\varepsilon = 16$ vs $\varepsilon = \infty$ }}
    \end{subfigure}
    \caption{KL divergence between each fine-tuned model and the base GPT-J 6B model, measured at each token position in generated sequences. KL divergence quantifies how much the fine-tuned model's next-token predictions differ from the base model's: higher values indicate that fine-tuning has shifted the model's predictions further from the base model's behavior. (a)~On unseen fine-tuning data (Wikipedia Science and AI), all fine-tuned models---including DP models---exhibit higher KL divergence from the base model than (b)~on data seen during pre-training. If DP models were simply reverting to their pre-training priors, we would expect low KL divergence on the unseen data, as the models would behave similarly to the base model on topics it was never fine-tuned on. The higher divergence on unseen data indicates the opposite: DP models do acquire new information from fine-tuning and diverge from their pre-trained behavior, even under strict privacy budgets.}
    \label{fig:rq2_kl_div}
\end{figure*}

A natural concern is that the DP fine-tuned models revert to their pre-trained priors during generation, rather than acquiring knowledge from the unseen fine-tuning data. Our evidence points to the contrary. A model reverting to its priors would be expected to resemble the base model more closely on the unseen fine-tuning data. However, as shown in \Cref{fig:rq2_kl_div} for both DP models, the KL divergence from the base model is higher on the unseen data (which the base model was never fine-tuned on) than on the seen data.

This is further corroborated by the 1D loss interpolation (\Cref{fig:loss_landscape}) in the model's subspace. Although the DP models remain relatively close to the base model in the parameter space compared to the non-private model, it sits in a region of the space associated with a lower loss. The non-private model, on the other hand, has shifted away significantly from the base model, which is reflective of a much more precise signal during fine-tuning. Additionally, as discussed in \Cref{sec:hallucinations_in_dp_finetuning} the DP fine-tuned models hallucinate at a rate higher than the base model on unseen topics. If the model was relying on its priors, we would expect hallucination rates to be similar to that of the base model. Together, these results provide clear evidence that DP models learn from the fine-tuning data and diverge from their pre-trained priors across both the unseen and seen evaluation sets.

\clearpage
\twocolumn
\subsection{Manual Annotations: Claim Annotation Interface}\label{app:claim-annotations}

\begin{figure*}[]
    \centering
    \includegraphics[width=0.6\textwidth]{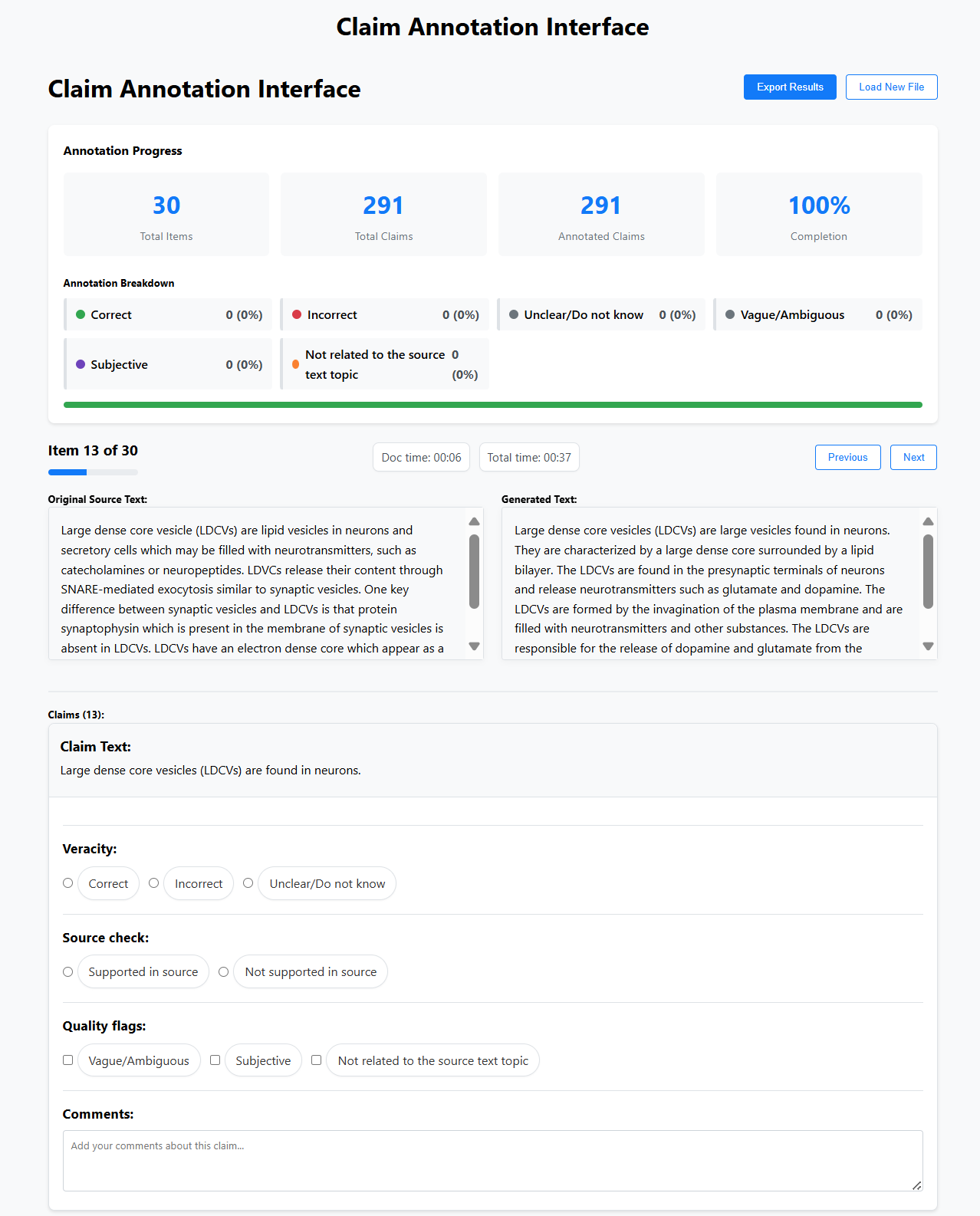}
    \caption{The interactive claim annotation tool used by participants to evaluate the factual correctness of claims from the model-generated text.}
    \label{fig:claim_annotation_interface}
\end{figure*}

\begin{table*}[h]
\centering
\begin{tabular}{llcc}
\hline
\textbf{Comparison} &                & \textbf{DP-inf} & \textbf{DP-16} \\
\hline
\multirow{2}{*}{Inter-Annotator}     
          & Veracity                & 0.57 & 0.84 \\
          & Support                 & 0.57 & 0.73 \\
\multirow{2}{*}{Model vs Annotator}  
          & Support (Annotator 1)      & 0.49 & 0.23 \\
          & Support (Annotator 2)      & 0.48 & 0.34 \\
\hline
\end{tabular}
\caption{Cohen's Kappa scores for DP-inf and DP-16 settings for human annotation of Wikipedia AI article claims.}
\label{tab: human_claim_annotation_agreement}
\end{table*}

\subsubsection{Annotation Guidelines}

In our study, annotators were tasked with evaluating the quality of factual claims from a model-generated text, with respect to a given source text. Each annotation instance consists of: (i) a source text corresponding to the original document for a given topic from Wikipedia; (ii) the text generated by the DP/non-DP fine-tuned models for the given topic; and (iii) a set of atomic claims extracted from the model-generated text. Annotators were required to assess each claim along multiple dimensions such as its factual accuracy, support from the source text, in addition to other quality flags.

\subsubsection{Annotation Dimensions}

\paragraph{1. Veracity}
Annotators evaluate the factual correctness of each claim based on real-world knowledge, independent of whether it appears in the source text.
\begin{itemize}
    \item \textbf{Correct}: The claim is factually accurate and verifiable through reliable sources.
    \item \textbf{Incorrect}: The claim is factually inaccurate or false.
    \item \textbf{Unclear/Do Not Know}: The claim is too vague, ambiguous, or requires some specialized expertise to verify.
\end{itemize}

\paragraph{2. Source Support}
This dimension captures whether the claim is explicitly supported by the content in the source text.
\begin{itemize}
    \item \textbf{Supported in Source}: The claim is directly mentioned or implied by the content in the source text.
    \item \textbf{Not Supported in Source}: The claim is either absent from the source or contradicted by the information in the source.
\end{itemize}

\paragraph{3. Quality Flags}
Annotators may also flag issues related to clarity or subjectivity.
\begin{itemize}
    \item \textbf{Vague / Ambiguous}: The claim does not communicate a precise meaning or is difficult to interpret.
    \item \textbf{Subjective}: The claim expresses an opinion or value judgment, as opposed to an objective fact.
\end{itemize}

\subsubsection{Annotation Procedure}

For each claim, annotators perform the following steps:
\begin{enumerate}
    \item They go through the source text and model-generated text to understand the context.
    \item Label each claim along the three dimensions described above.
    \item Optionally, provide free-text comments to clarify ambiguous cases or provide additional context. Additionally, they were permitted to make use of search engines or external judgements to annotate each of the claims.
\end{enumerate}

\clearpage
\subsection{Expanded FactScore Evaluations}\label{app:additional_exp}

\begin{figure*}[!htbp]
    \centering
    \begin{subfigure}{0.4\textwidth}
        \centering
        \includegraphics[width=\linewidth]{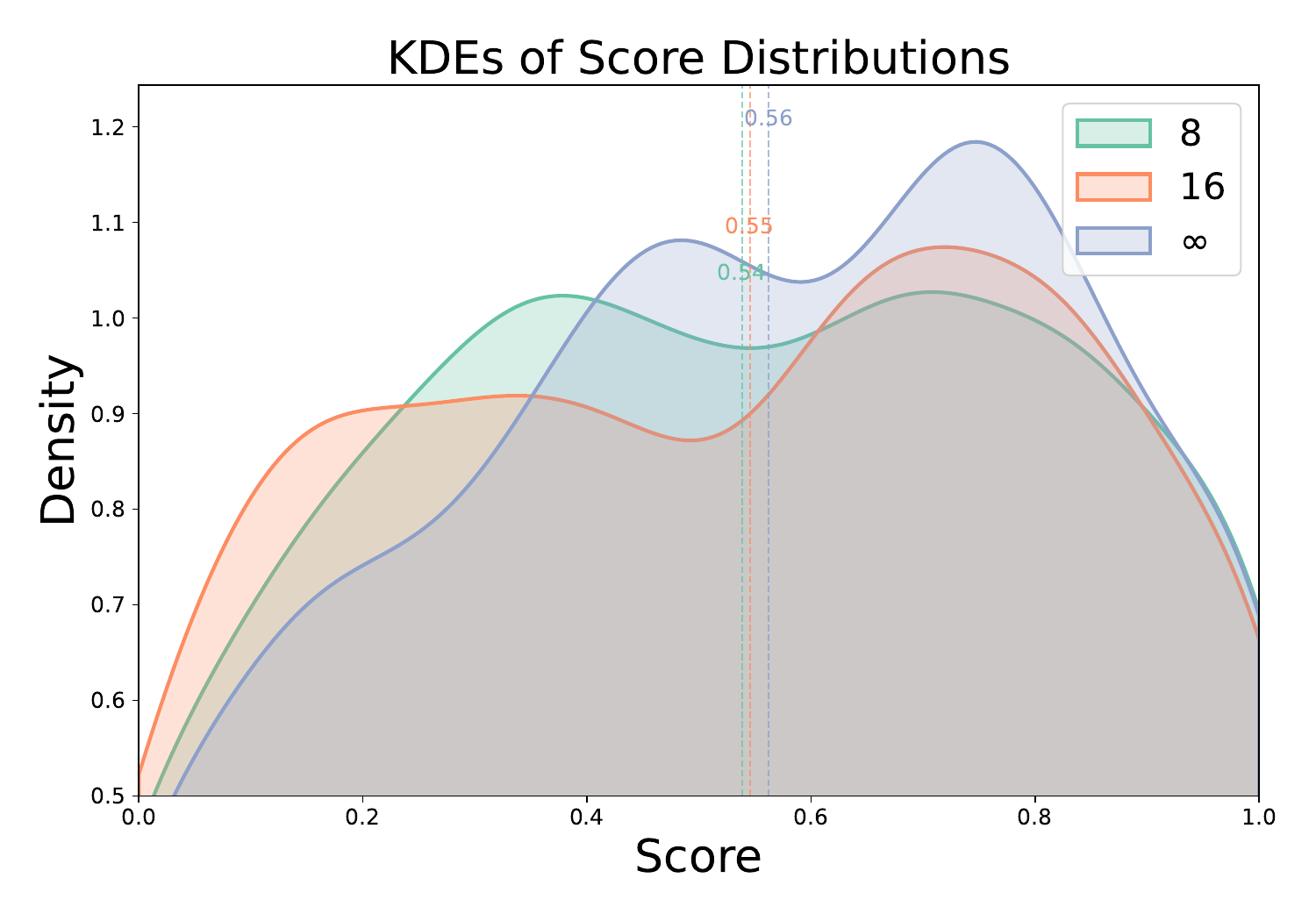}
        \caption{Temperature = 0.5}
        % \label{fig:kde2}
    \end{subfigure}
    \hspace{0.01\textwidth}
    \begin{subfigure}{0.4\textwidth}
        \centering
        \includegraphics[width=\linewidth]{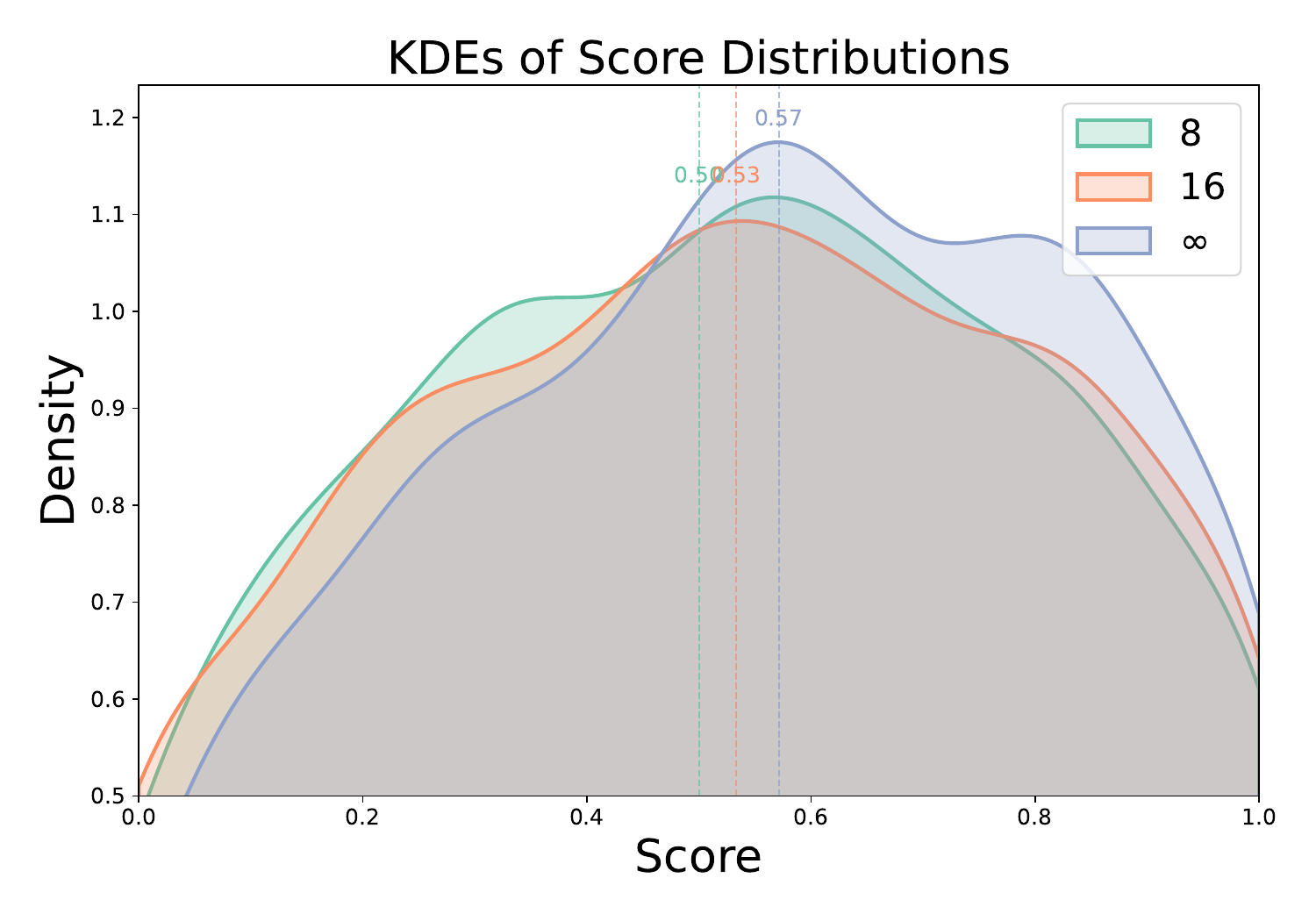}
        \caption{Temperature = 0.7}
        % \label{fig:kde3}
    \end{subfigure}
    \hspace{0.01\textwidth}
    \begin{subfigure}{0.4\textwidth}
        \centering
        \includegraphics[width=\linewidth]{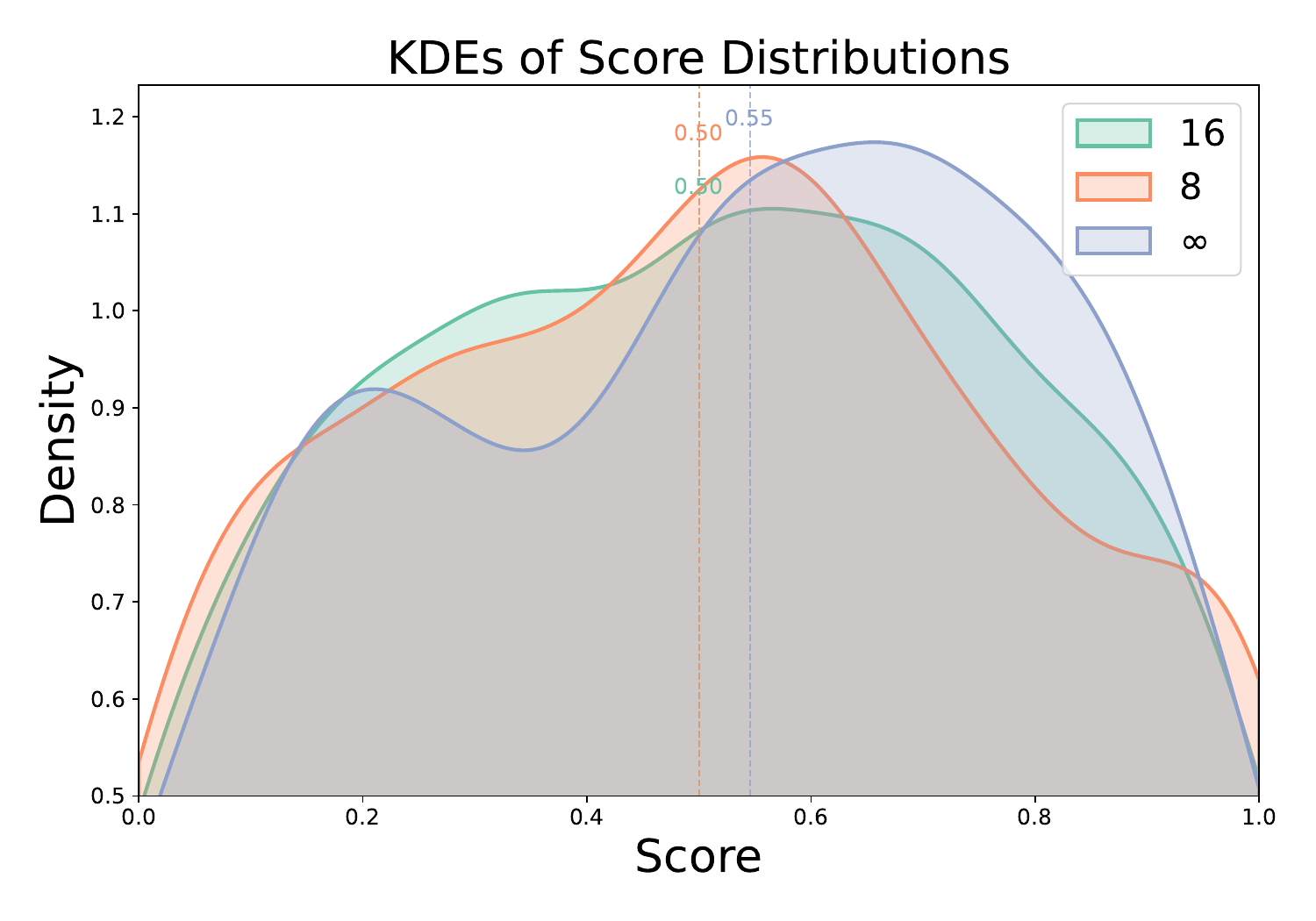}
        \caption{Temperature = 1.0}
       %  \label{fig:kde4}
    \end{subfigure}
    \caption{
        KDE plots of FactScore distributions of texts generated from topics in the Wikipedia Science data for under different temperature settings for GPT-J 6B.
    }
    \label{fig:kde_factscore_wiki_science}
\end{figure*}

\begin{figure*}[!htbp]
    \centering
    \hspace{0.01\textwidth}
    \begin{subfigure}{0.4\textwidth}
        \centering
        \includegraphics[width=\linewidth]{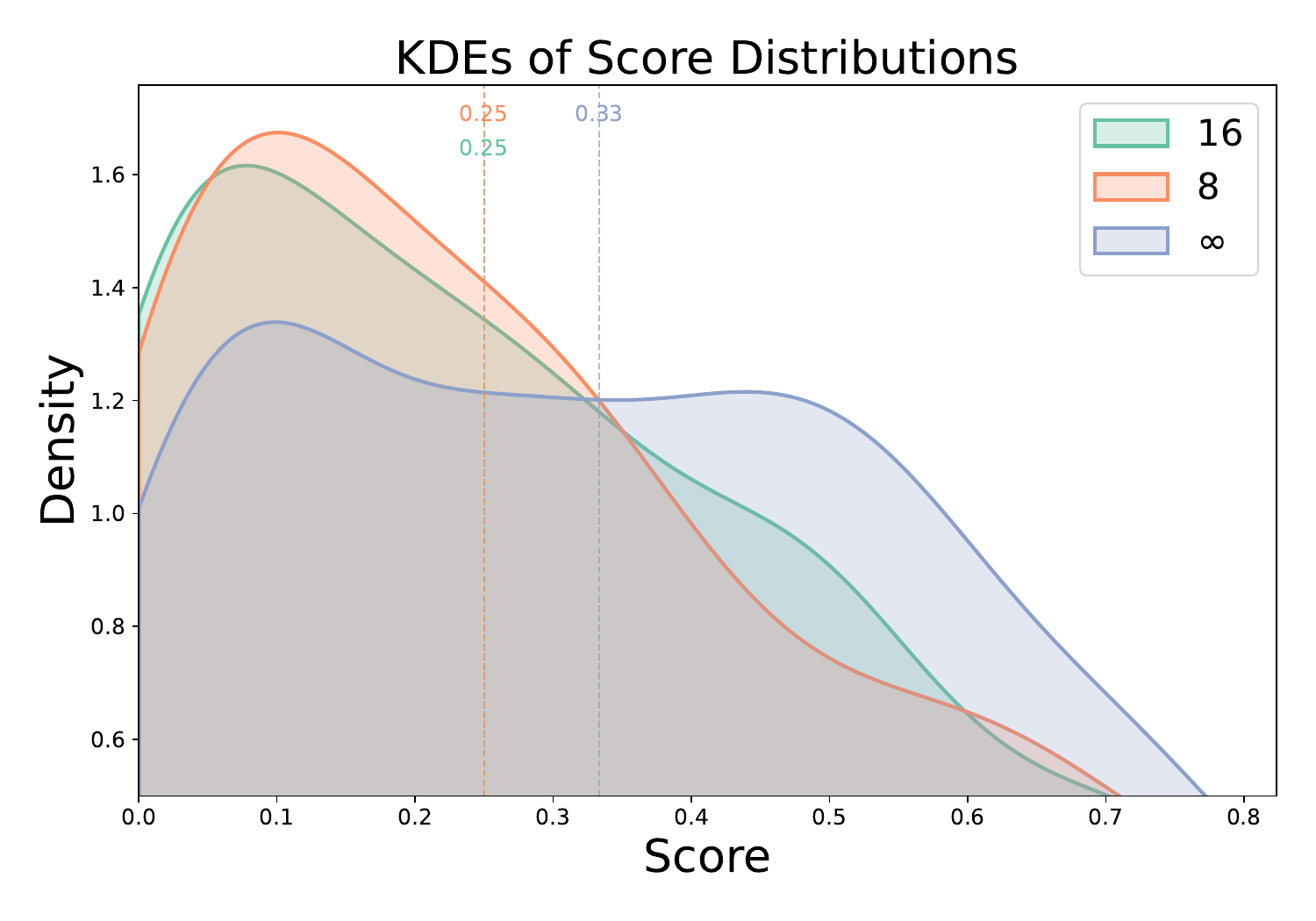}
        \caption{Temperature = 0.5}
        % \label{fig:kde2}
    \end{subfigure}
    \hspace{0.01\textwidth}
    \begin{subfigure}{0.4\textwidth}
        \centering
        \includegraphics[width=\linewidth]{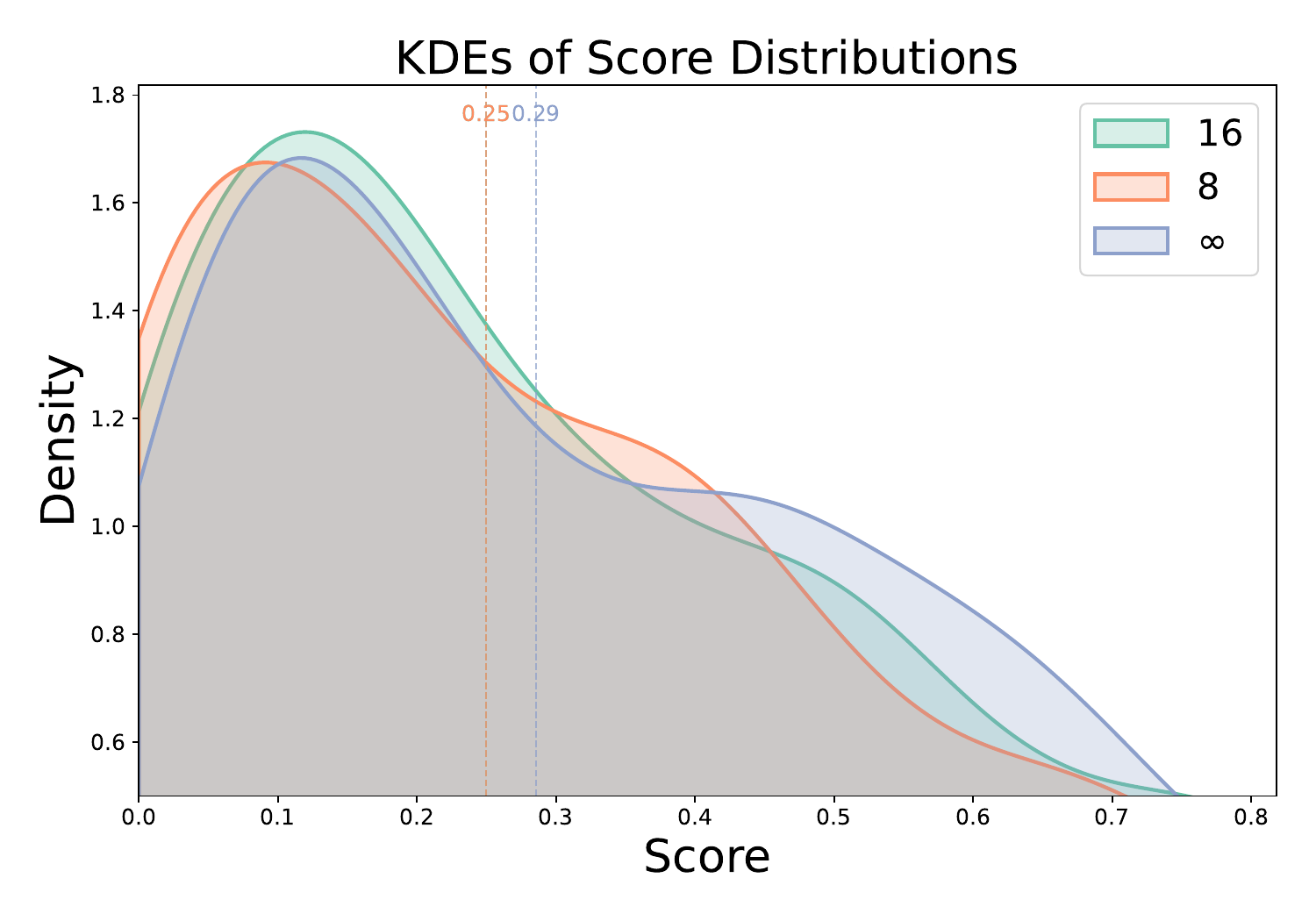}
        \caption{Temperature = 0.7}
        % \label{fig:kde3}
    \end{subfigure}
    \hspace{0.01\textwidth}
    \begin{subfigure}{0.4\textwidth}
        \centering
        \includegraphics[width=\linewidth]{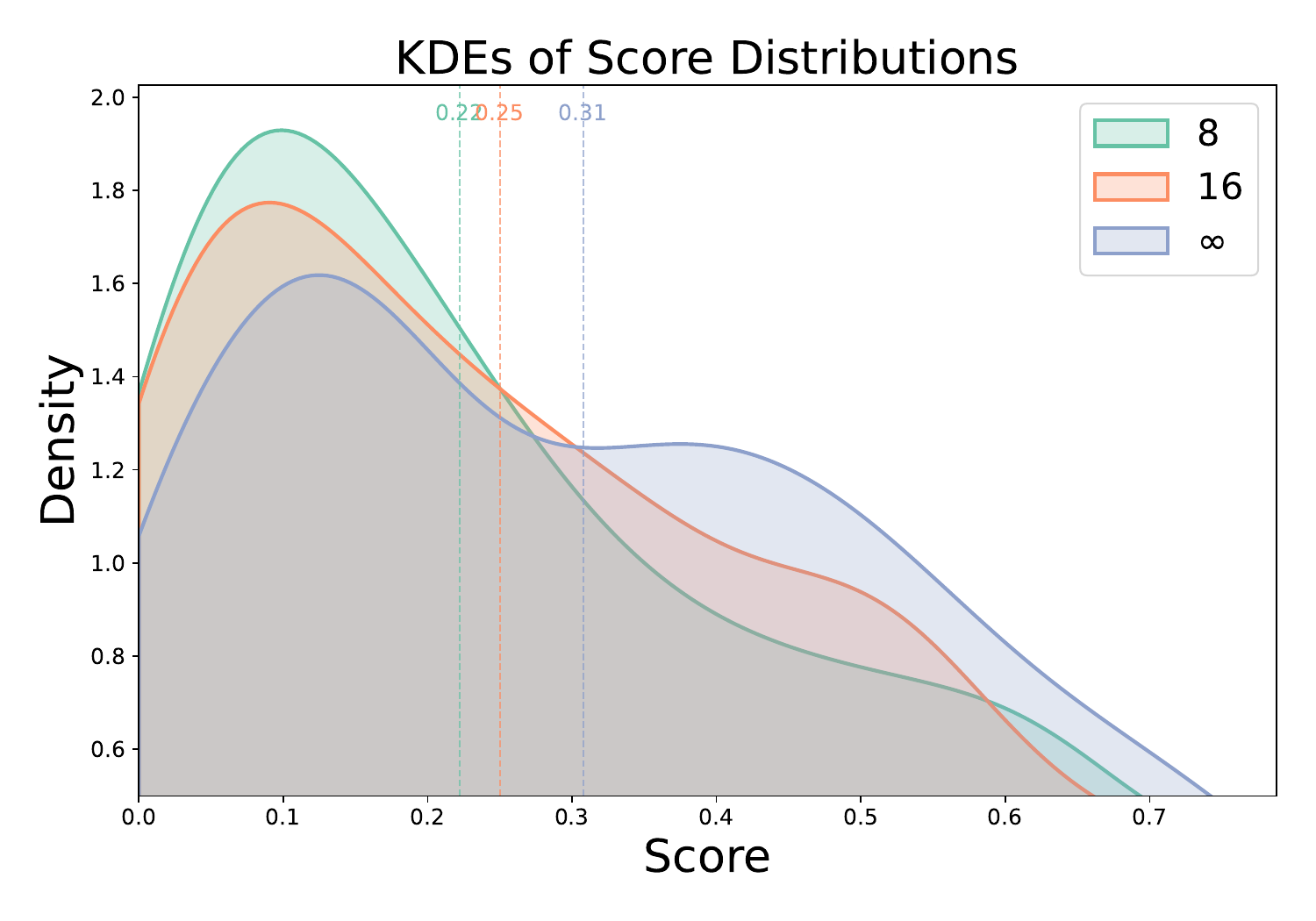}
        \caption{Temperature = 1.0}
        % \label{fig:kde4}
    \end{subfigure}
    \caption{
        KDE plots of FactScore distributions of texts generated from topics in the Wikipedia AI data for under different temperature settings for GPT-J 6B.
    }
    \label{fig:kde_factscore_wiki_ai}
\end{figure*}

\begin{figure*}[!htbp]
    \centering
    \hspace{0.01\textwidth}
    \begin{subfigure}{0.4\textwidth}
        \centering
        \includegraphics[width=\linewidth]{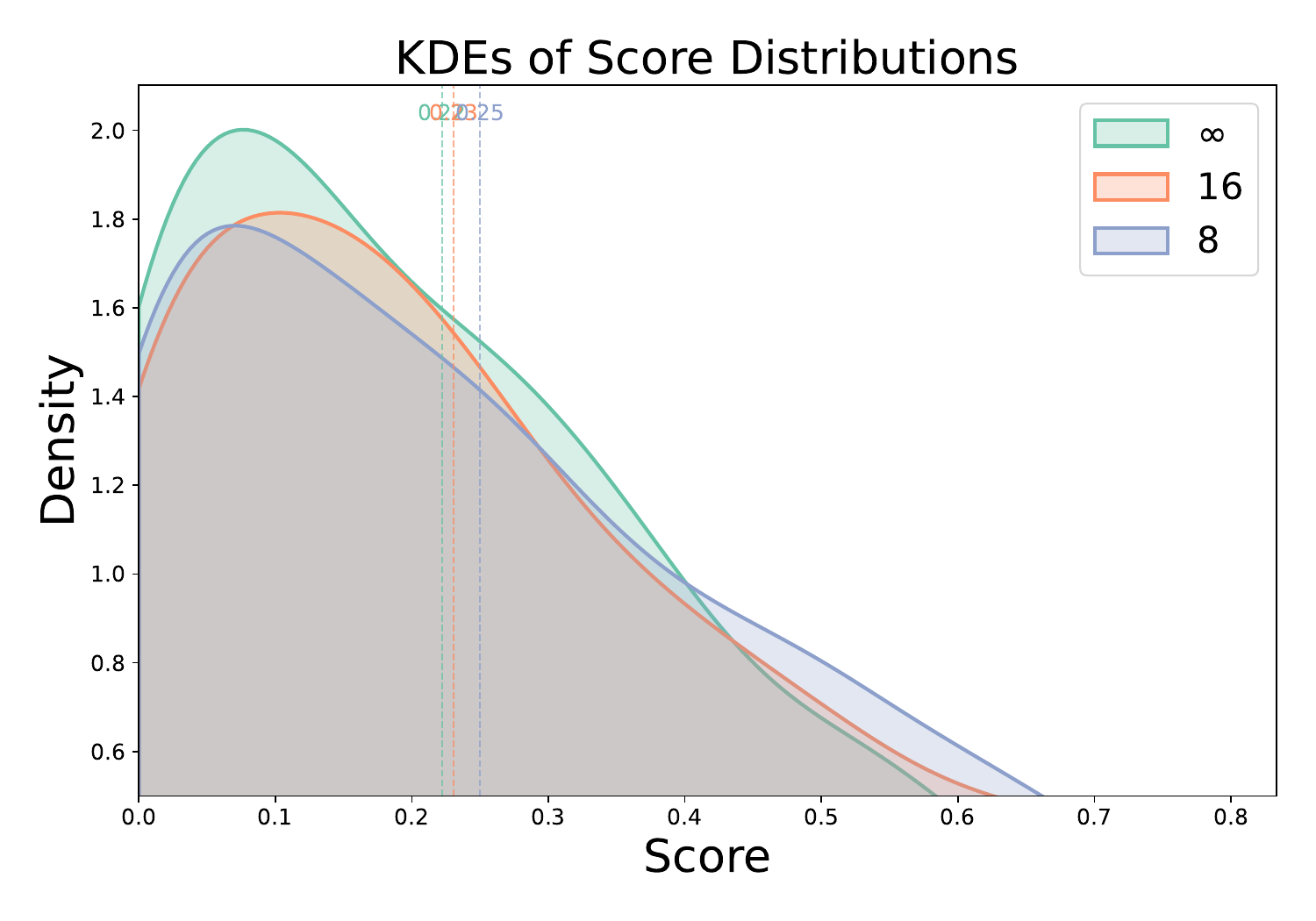}
        \caption{Temperature = 0.5}
        % \label{fig:kde2}
    \end{subfigure}
    \hspace{0.01\textwidth}
    \begin{subfigure}{0.4\textwidth}
        \centering
        \includegraphics[width=\linewidth]{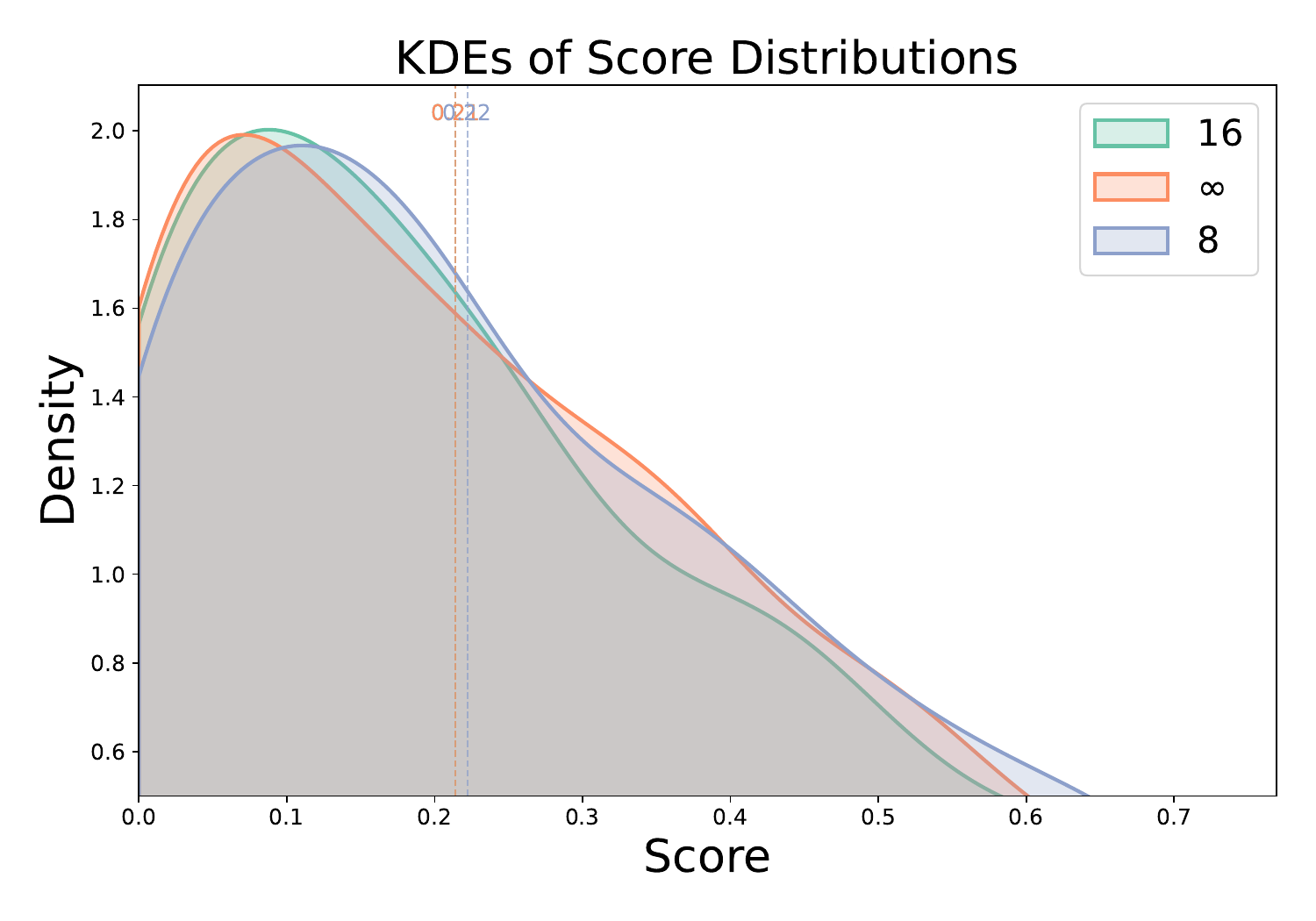}
        \caption{Temperature = 0.7}
       %  \label{fig:kde3}
    \end{subfigure}
    \hspace{0.01\textwidth}
    \begin{subfigure}{0.4\textwidth}
        \centering
        \includegraphics[width=\linewidth]{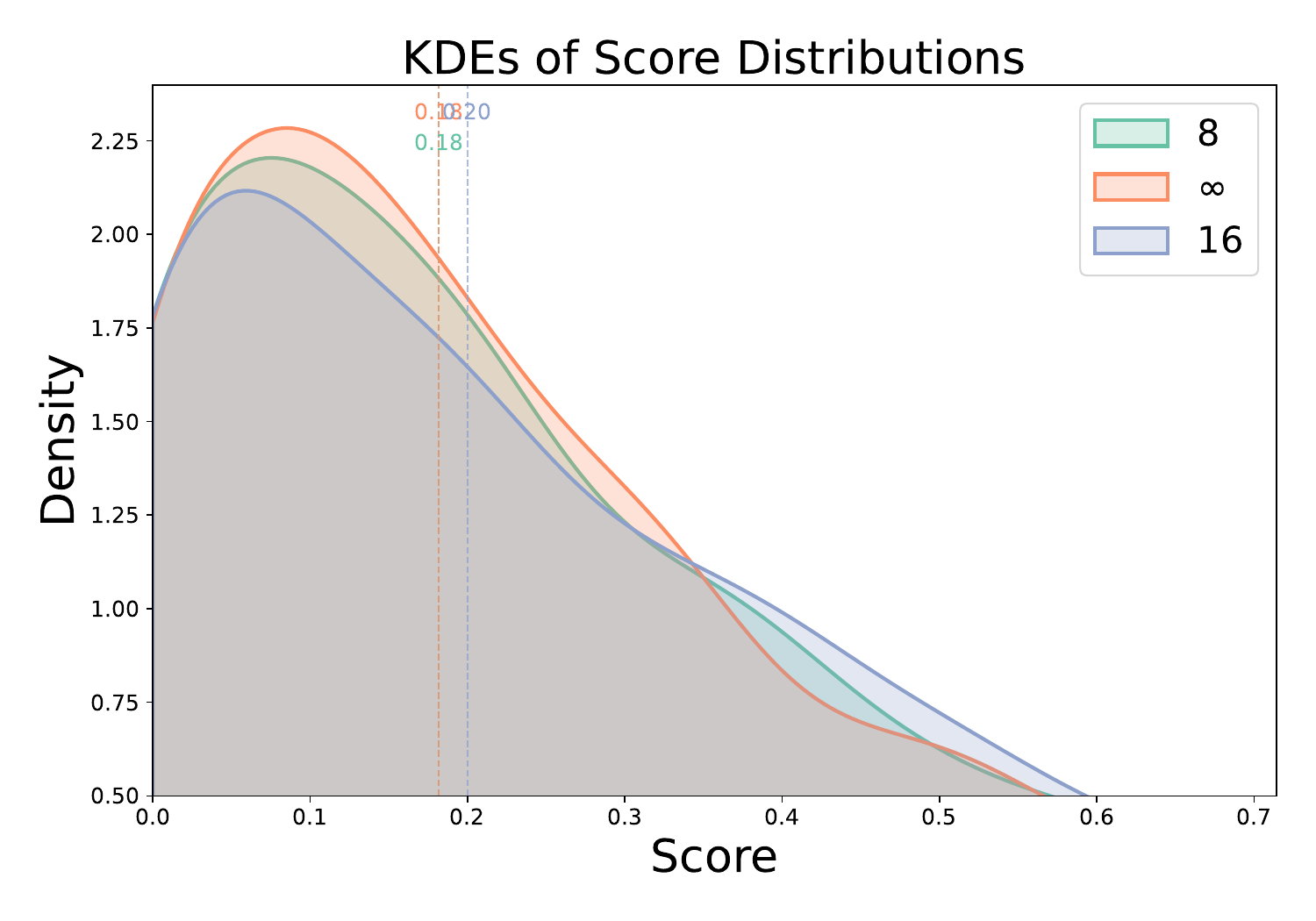}
        \caption{Temperature = 1.0}
       % \label{fig:kde4}
    \end{subfigure}
    \caption{
        KDE plots of FactScore distributions of texts generated from topics in the Wikipedia Pretraining data for under different temperature settings for GPT-J 6B.
    }
    \label{fig:kde_factscore_wiki_pretrain}
\end{figure*}

\begin{figure*}[ht]
    \centering
    \begin{subfigure}{0.3\textwidth}
        \centering
        \includegraphics[width=\linewidth]{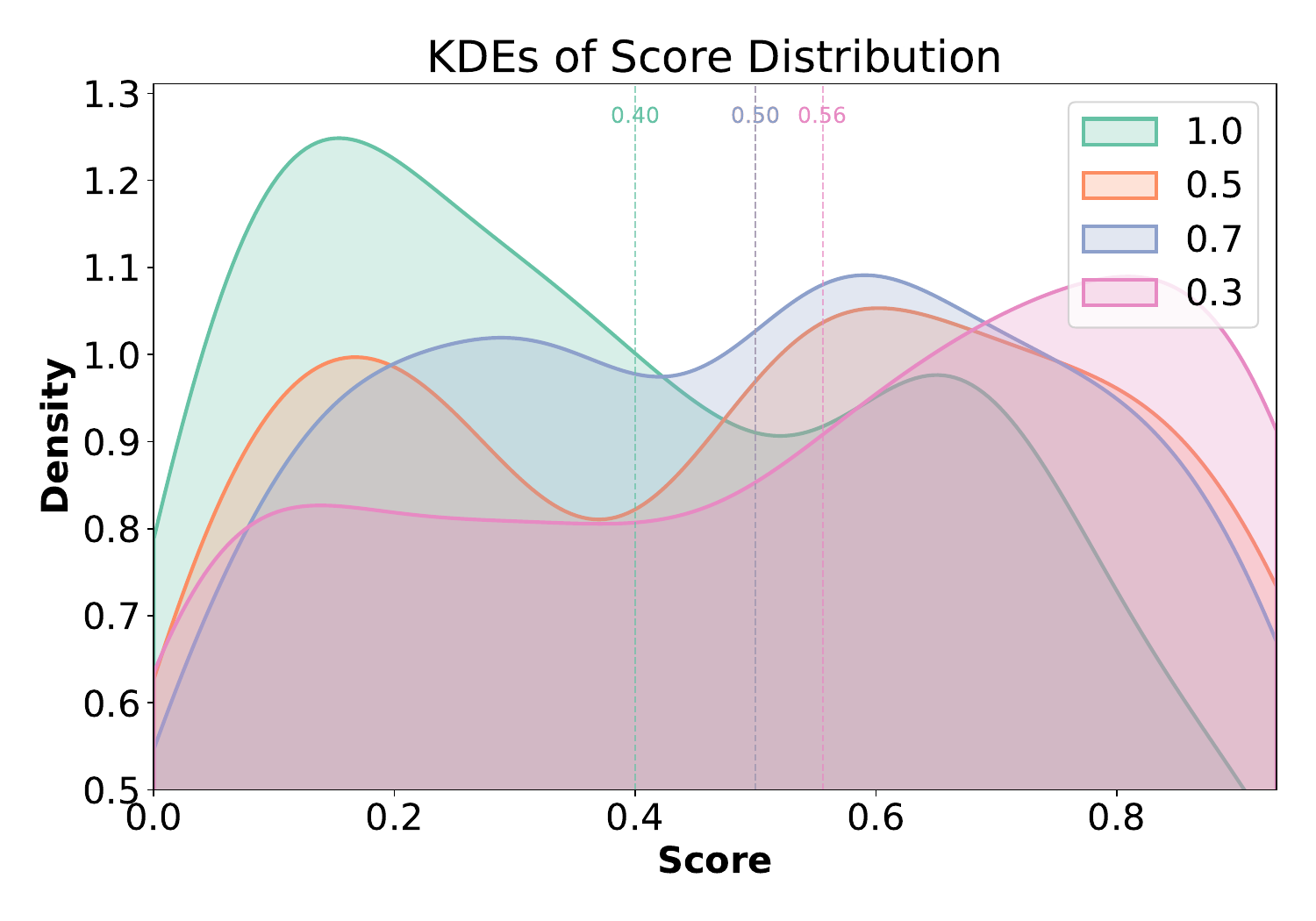}
        \caption{Gemma3-1B-PT}
       % \label{fig:kde1}
    \end{subfigure}
    \hspace{0.01\textwidth}
    \begin{subfigure}{0.3\textwidth}
        \centering
        \includegraphics[width=\linewidth]{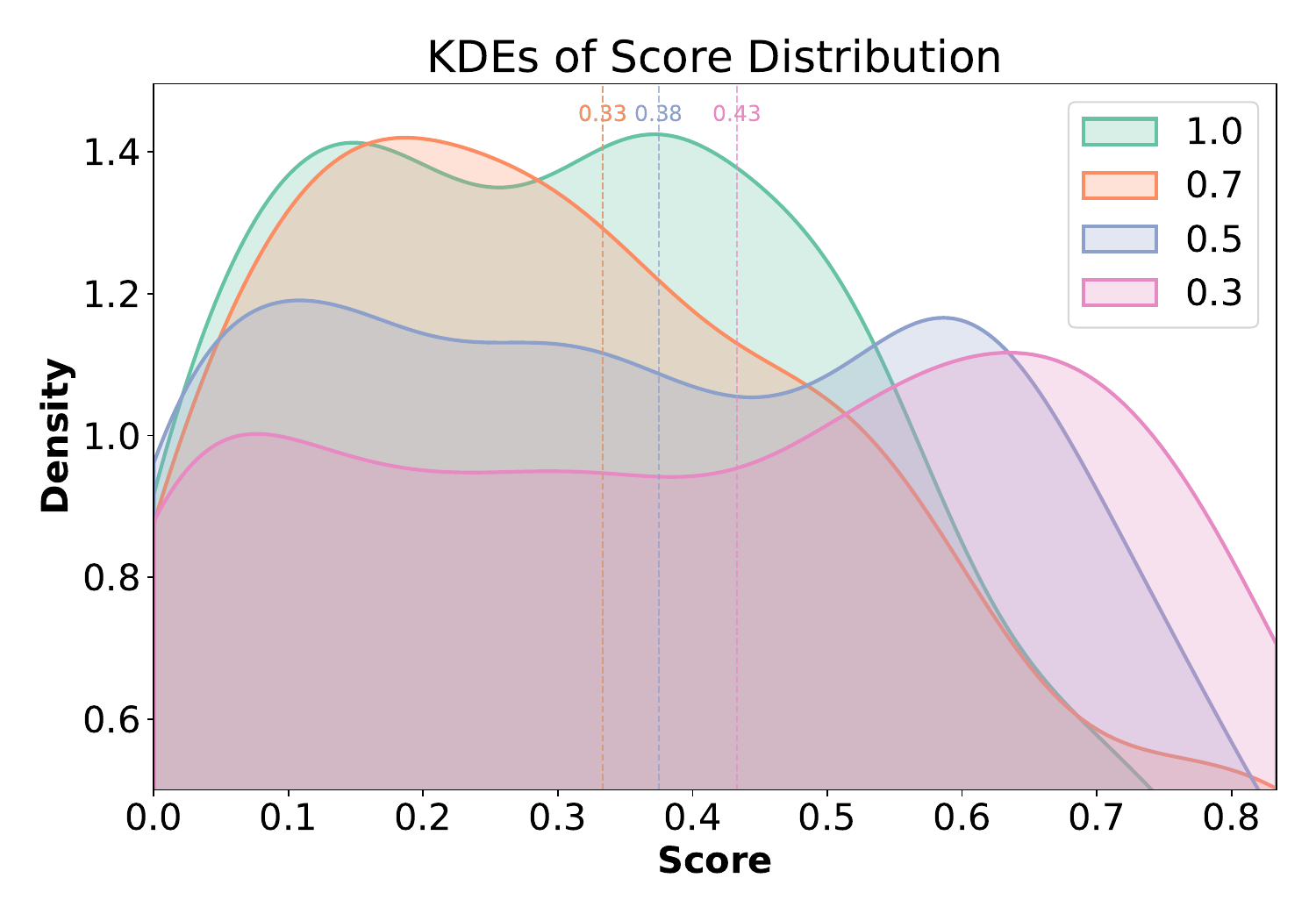}
        \caption{VaultGemma-1B}
       % \label{fig:kde2}
    \end{subfigure}
    \hspace{0.01\textwidth}
    \begin{subfigure}{0.3\textwidth}
        \centering
        \includegraphics[width=\linewidth]{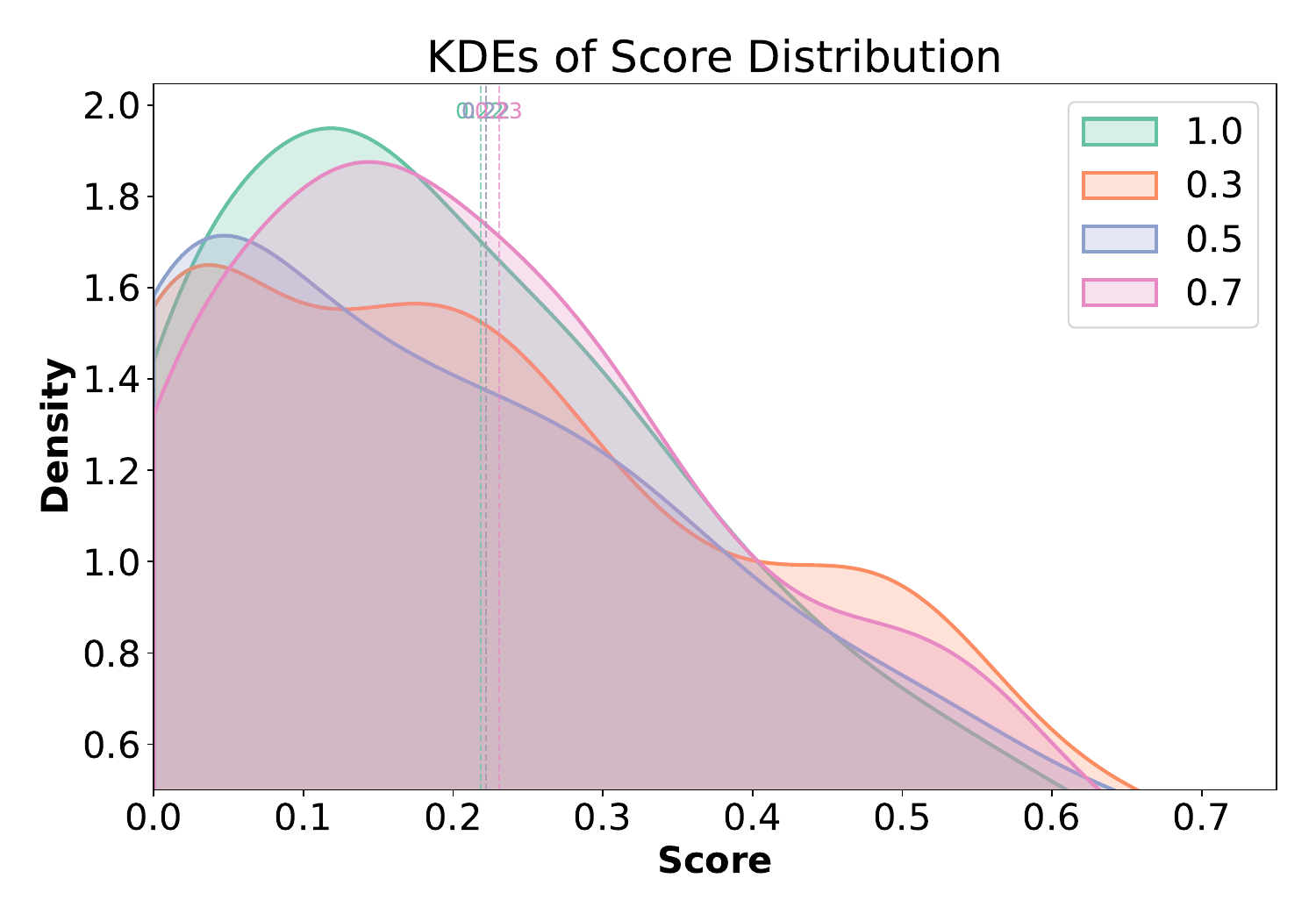}
        \caption{GPT2-XL}
       % \label{fig:kde3}
    \end{subfigure}
    \hspace{0.01\textwidth}
    \caption{
        KDE plots of FactScore distributions of texts generated from topics in the Wikipedia AI data for the pre-trained models.
    }
    \label{fig:kde_factscore_wiki_ai_pretrained}
\end{figure*}

\begin{figure*}[!htbp]
    \centering
    \begin{subfigure}{0.3\textwidth}
        \centering
        \includegraphics[width=\linewidth]{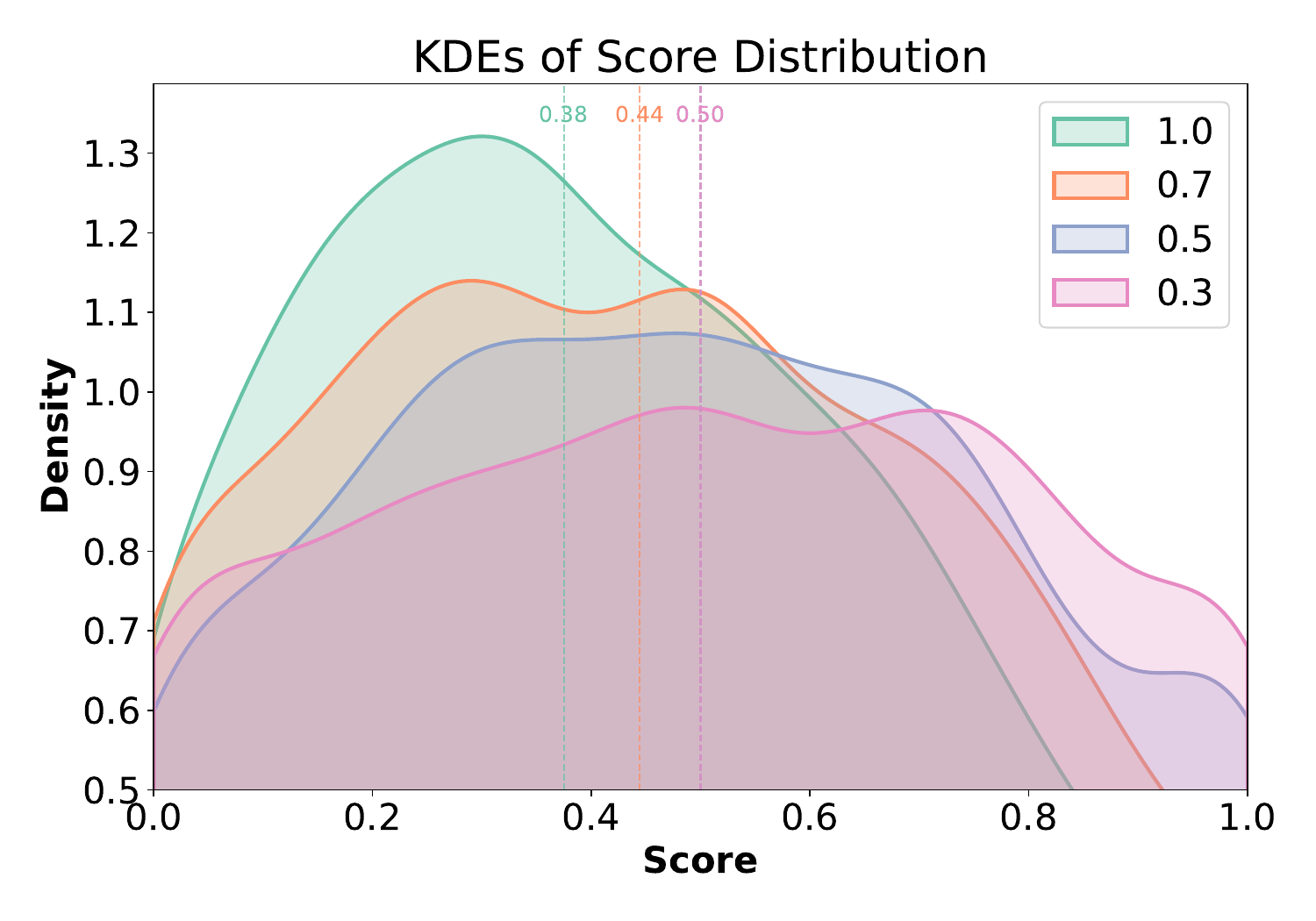}
        \caption{Gemma3-1B-PT}
        % \label{fig:kde1}
    \end{subfigure}
    \hspace{0.01\textwidth}
    \begin{subfigure}{0.3\textwidth}
        \centering
        \includegraphics[width=\linewidth]{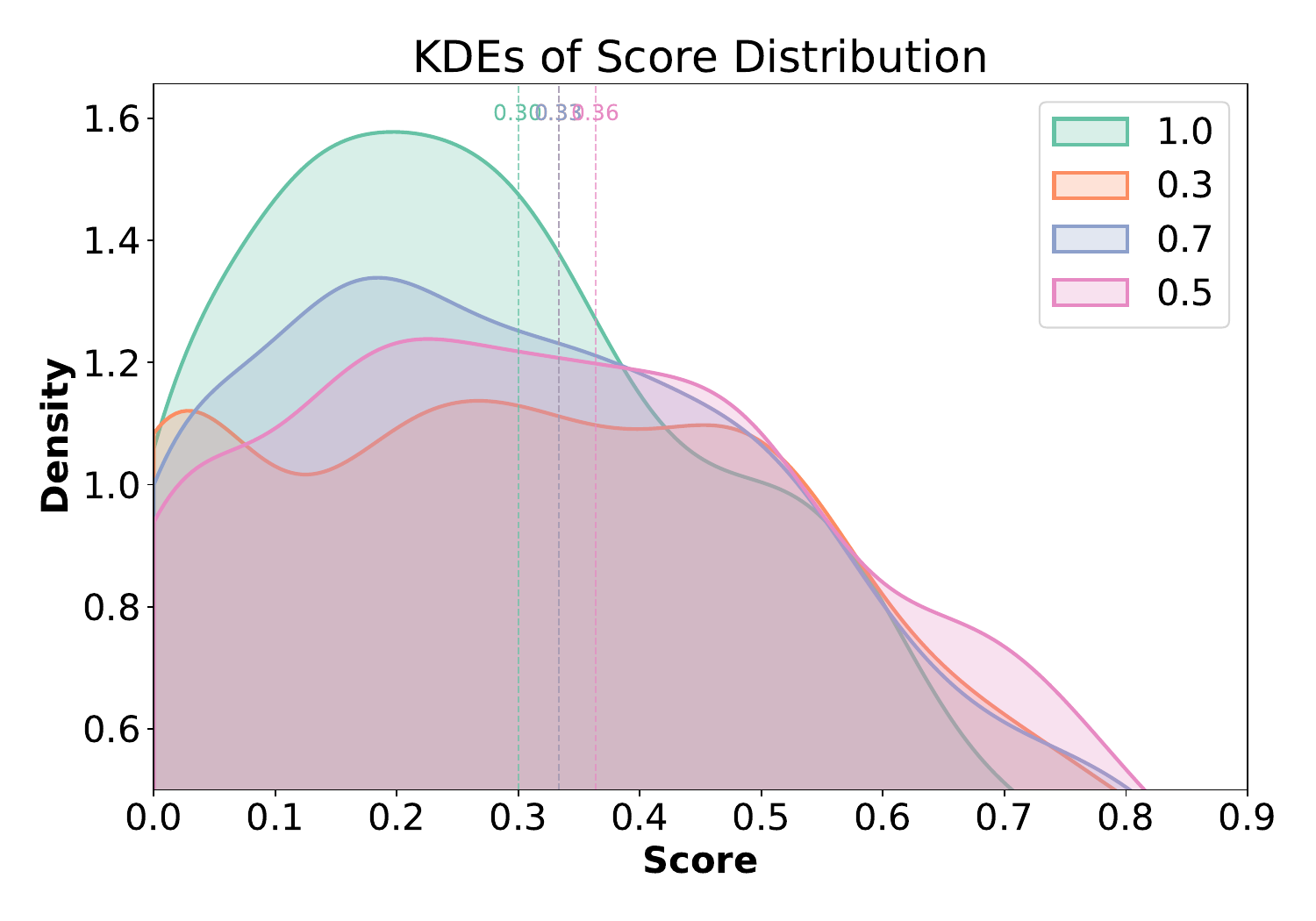}
        \caption{VaultGemma-1B}
       %  \label{fig:kde2}
    \end{subfigure}
    \hspace{0.01\textwidth}
    \begin{subfigure}{0.3\textwidth}
        \centering
        \includegraphics[width=\linewidth]{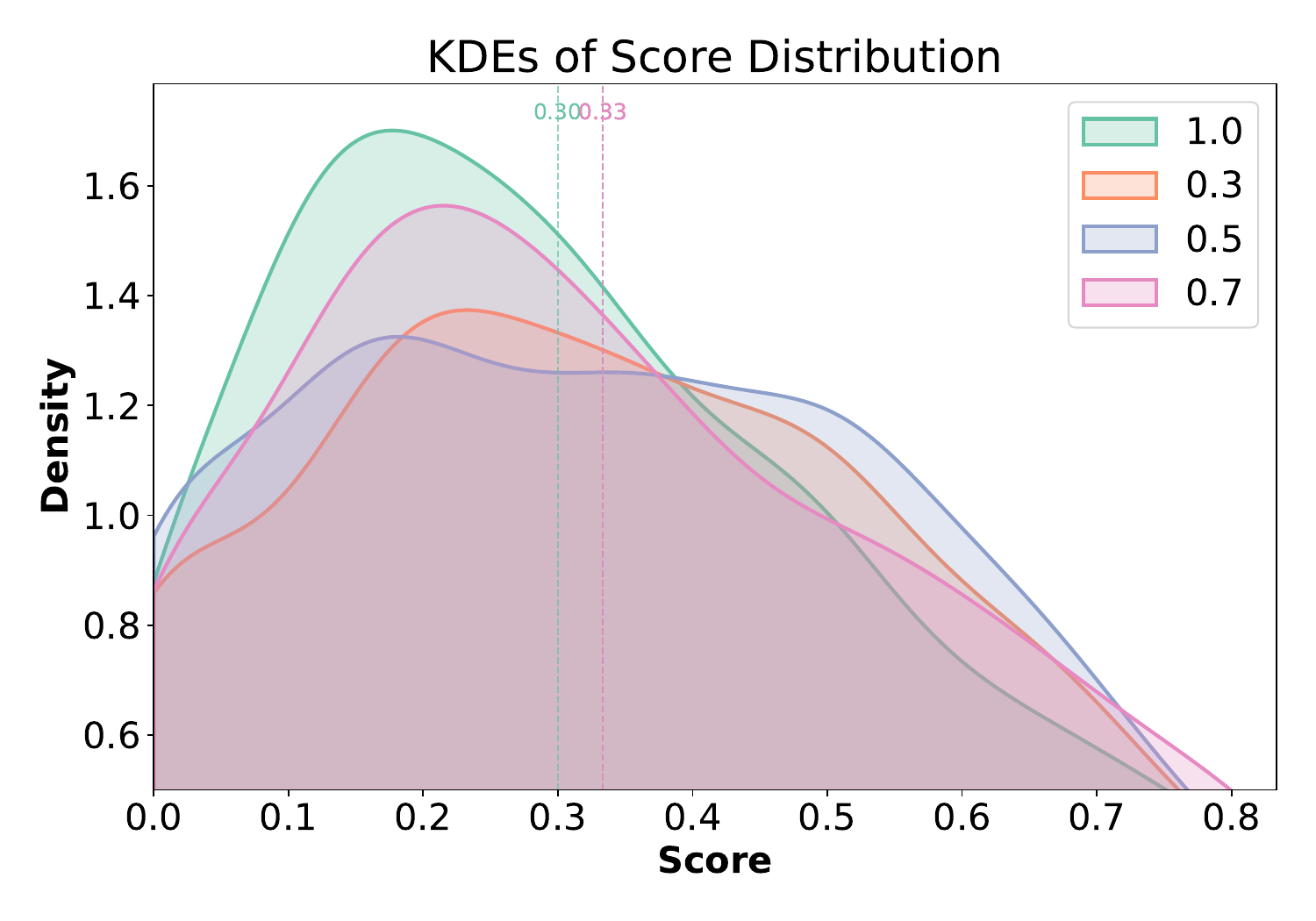}
        \caption{GPT2-XL}
       % \label{fig:kde3}
    \end{subfigure}
    \hspace{0.01\textwidth}
    \caption{
        KDE plots of FactScore distributions of texts generated from topics in the Wikipedia Science data for the pre-trained models.
    }
    \label{fig:kde_factscore_wiki_science_pretrained}
\end{figure*}

\begin{figure*}[!htbp]
    \centering
    \begin{subfigure}{0.3\textwidth}
        \centering
        \includegraphics[width=\linewidth]{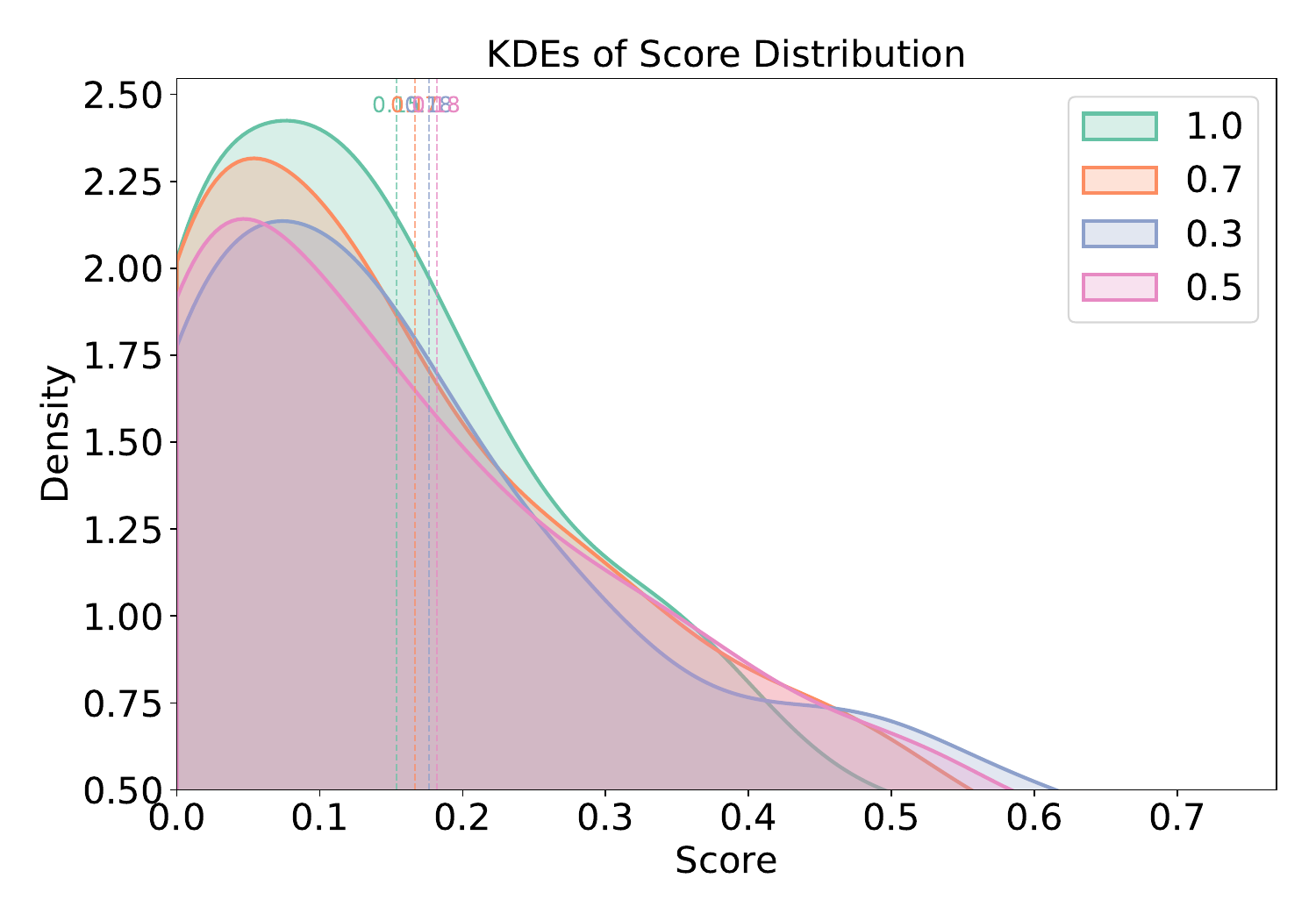}
        \caption{Gemma3-1B-PT}
       % \label{fig:kde1}
    \end{subfigure}
    \hspace{0.01\textwidth}
    \begin{subfigure}{0.3\textwidth}
        \centering
        \includegraphics[width=\linewidth]{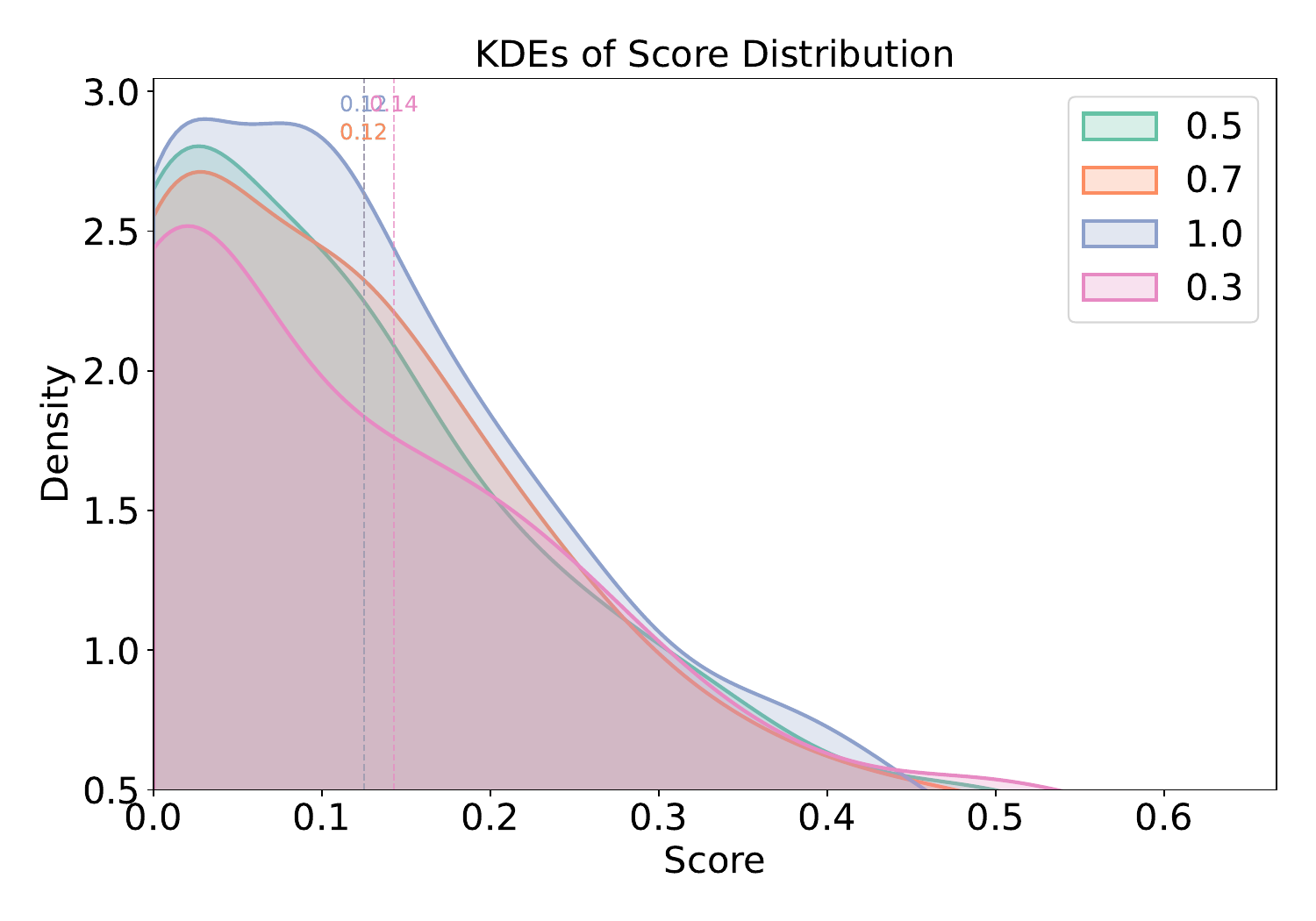}
        \caption{VaultGemma-1B}
       % \label{fig:kde2}
    \end{subfigure}
    \hspace{0.01\textwidth}
    \begin{subfigure}{0.3\textwidth}
        \centering
        \includegraphics[width=\linewidth]{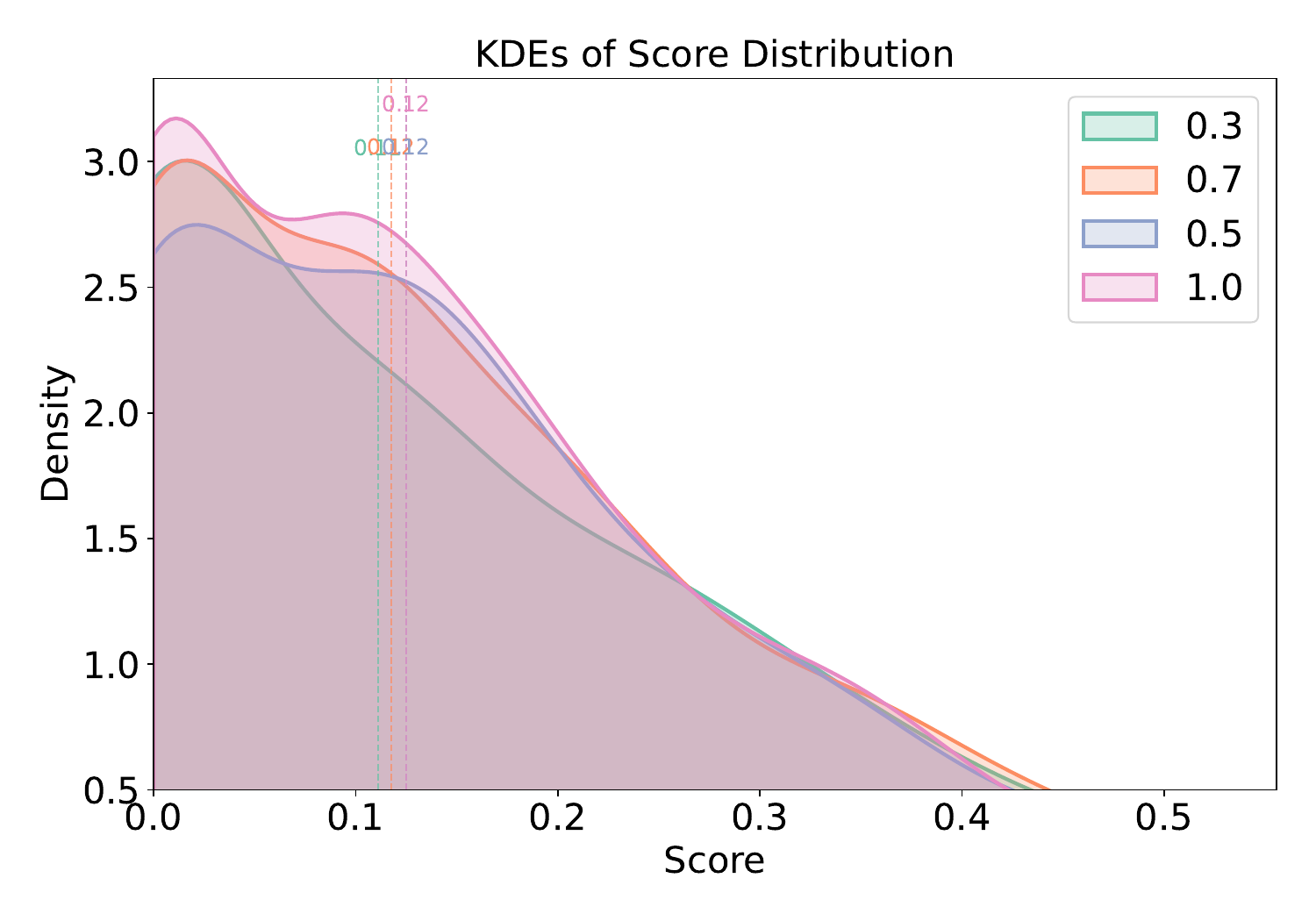}
        \caption{GPT2-XL}
       % \label{fig:kde3}
    \end{subfigure}
    \hspace{0.01\textwidth}
    \caption{
        KDE plots of FactScore distributions of texts generated from topics in the Wikipedia Pretraining data for the pre-trained models.
    }
    \label{fig:kde_factscore_wiki_pretrain_pretrained}
\end{figure*}

We report additional experiments in the following tables:

\begin{itemize}
\item FactScore evaluations with different claim decomposition and evaluation models in \Cref{tab:llama-3b-factscore-claim-eval} and \Cref{tab:deepseek-factscore-claim-eval}
\item FactScore distributions per temperature setting under different models in \Cref{fig:kde_factscore_wiki_ai}, \Cref{fig:kde_factscore_wiki_pretrain}, \Cref{fig:kde_factscore_wiki_science}, \Cref{fig:kde_factscore_wiki_science_pretrained}, \Cref{fig:kde_factscore_wiki_pretrain_pretrained}, \Cref{fig:kde_factscore_wiki_ai_pretrained}.
\item FactScores on unseen Wikipedia articles after fine-tuning on only the Wikipedia training data in \Cref{tab:factscores-wikipedia-unseen-only}.
\end{itemize}

%\section{FactScore evaluations with different claim decomposition and evaluation models}\label{sec:factscore_evals_across_claim}

\begin{table*}[h]
\renewcommand{\arraystretch}{1.2}
\small
\centering
\begin{tabular}{@{}cccccccccc@{}}
\toprule
\textbf{Temperature} & \textbf{\begin{tabular}[c]{@{}c@{}}Pretrained \\ Model\end{tabular}} & \textbf{\begin{tabular}[c]{@{}c@{}}Average \\ FS\end{tabular}} & \textbf{\begin{tabular}[c]{@{}c@{}}Median \\ FS\end{tabular}} & \textbf{\begin{tabular}[c]{@{}c@{}}Q1 \\ FS\end{tabular}} & \textbf{\begin{tabular}[c]{@{}c@{}}Q3 \\ FS\end{tabular}} & \textbf{\begin{tabular}[c]{@{}c@{}}Avg Max\\ FS / Topic\end{tabular}} & \textbf{\begin{tabular}[c]{@{}c@{}}Avg Min \\ FS / Topic\end{tabular}} & \textbf{\begin{tabular}[c]{@{}c@{}}\# of \\ FS \\ \textgreater{}0.5\end{tabular}} & \textbf{\begin{tabular}[c]{@{}c@{}}\# of \\ FS \\ \textless{}=0.5\end{tabular}} \\ \midrule
 & $\infty$ & 56.3 & 57.1 & 33.3 & 80 & 71.2 & 41.2 & 397 & 491 \\
0.3 & 16 & 54.3 & 57.1 & 28.6 & 80 & 69.7 & 38.9 & 314 & 564 \\
 & 8 & 53.2 & 54.5 & 28.6 & 76.9 & 68.3 & 37.9 & 294 & 588 \\ \midrule
 & $\infty$ & 55.2 & 57.7 & 33.3 & 77.8 & 68.6 & 41 & 418 & 470 \\
0.5 & 16 & 51.7 & 50 & 25 & 77.8 & 67.6 & 35.5 & 335 & 547 \\
 & 8 & 52.1 & 54.2 & 28.6 & 77.1 & 67.5 & 36.7 & 332 & 548 \\ \midrule
 & $\infty$ & 55 & 57.1 & 30 & 80 & 69.8 & 40.6 & 436 & 452 \\
0.7 & 16 & 52.6 & 54.2 & 28.6 & 77.8 & 67.4 & 37.9 & 380 & 502 \\
 & 8 & 51.5 & 50 & 28.6 & 75 & 67.2 & 35.8 & 366 & 522 \\ \midrule
 & $\infty$ & 52.6 & 53.8 & 30 & 77.8 & 67.5 & 37.8 & 442 & 446 \\
1.0 & 16 & 51.2 & 50 & 27.3 & 75 & 66.1 & 36.2 & 420 & 464 \\
 & 8 & 49.8 & 50 & 25 & 72.7 & 65.4 & 34.4 & 381 & 505 \\ \bottomrule
\end{tabular}
\caption{Factuality evaluation scores for different temperature settings for the Wikipedia Science articles, using Meta Llama 3.2-3b-Instruct for claim evaluation and decomposition.}
\label{tab:llama-3b-factscore-claim-eval}
\end{table*}

\begin{table*}[h]
\renewcommand{\arraystretch}{1.2}
\small
\centering
\begin{tabular}{@{}cccccccccc@{}}
\toprule
\textbf{Temperature} & \textbf{\begin{tabular}[c]{@{}c@{}}Pretrained \\ Model\end{tabular}} & \textbf{\begin{tabular}[c]{@{}c@{}}Average \\ FS\end{tabular}} & \textbf{\begin{tabular}[c]{@{}c@{}}Median \\ FS\end{tabular}} & \textbf{\begin{tabular}[c]{@{}c@{}}Q1 \\ FS\end{tabular}} & \textbf{\begin{tabular}[c]{@{}c@{}}Q3 \\ FS\end{tabular}} & \textbf{\begin{tabular}[c]{@{}c@{}}Avg Max\\ FS / Topic\end{tabular}} & \textbf{\begin{tabular}[c]{@{}c@{}}Avg Min \\ FS / Topic\end{tabular}} & \textbf{\begin{tabular}[c]{@{}c@{}}\# of \\ FS \\ \textgreater{}0.5\end{tabular}} & \textbf{\begin{tabular}[c]{@{}c@{}}\# of \\ FS \\ \textless{}0.5\end{tabular}} \\ \midrule
 & $\infty$ & 48.6 & 50.0 & 28.9 & 66.7 & 71.4 & 27.2 & 108 & 765 \\
0.3 & 16 & 47.5 & 45.5 & 25.0 & 66.7 & 70.9 & 24.2 & 79 & 788 \\
 & 8 & 46.9 & 44.4 & 25.0 & 66.7 & 69.0 & 26.0 & 61 & 809 \\ \midrule
 & $\infty$ & 48.9 & 50.0 & 30.0 & 66.7 & 68.7 & 29.1 & 112 & 759 \\
0.5 & 16 & 44.6 & 44.4 & 25.0 & 62.5 & 70.8 & 18.8 & 82 & 789 \\
 & 8 & 46.0 & 50.0 & 25.0 & 66.7 & 65.0 & 27.3 & 81 & 784 \\ \midrule
 & $\infty$ & 45.5 & 44.4 & 27.3 & 63.6 & 63.8 & 27.4 & 123 & 752 \\
0.7 & 16 & 45.5 & 44.4 & 27.3 & 62.5 & 68.0 & 23.1 & 99 & 766 \\
 & 8 & 45.8 & 44.4 & 25.0 & 66.7 & 66.0 & 25.9 & 85 & 785 \\ \midrule
 & $\infty$ & 44.4 & 42.9 & 25.0 & 62.5 & 69.2 & 22.6 & 128 & 738 \\
1.0 & 16 & 43.9 & 42.9 & 25.0 & 60.0 & 62.7 & 26.1 & 85 & 783 \\
 & 8 & 44.5 & 42.9 & 27.3 & 62.5 & 63.0 & 25.5 & 105 & 772 \\ \bottomrule
\end{tabular}
\caption{Factuality evaluation scores for different temperature settings for the Wikipedia Science articles, using DeepSeek-R1-Distill-Qwen-7B for claim evaluation and decomposition.}
\label{tab:deepseek-factscore-claim-eval}
\end{table*}

%\section{FactScore distributions per  temperature setting under different models}

\subsection{Variability in FactScore}
\label{sec:stddev-factscore}

\begin{figure*}[!htbp]
    \centering
    \begin{subfigure}[t]{0.48\linewidth}
        \centering
        \includegraphics[width=\linewidth]{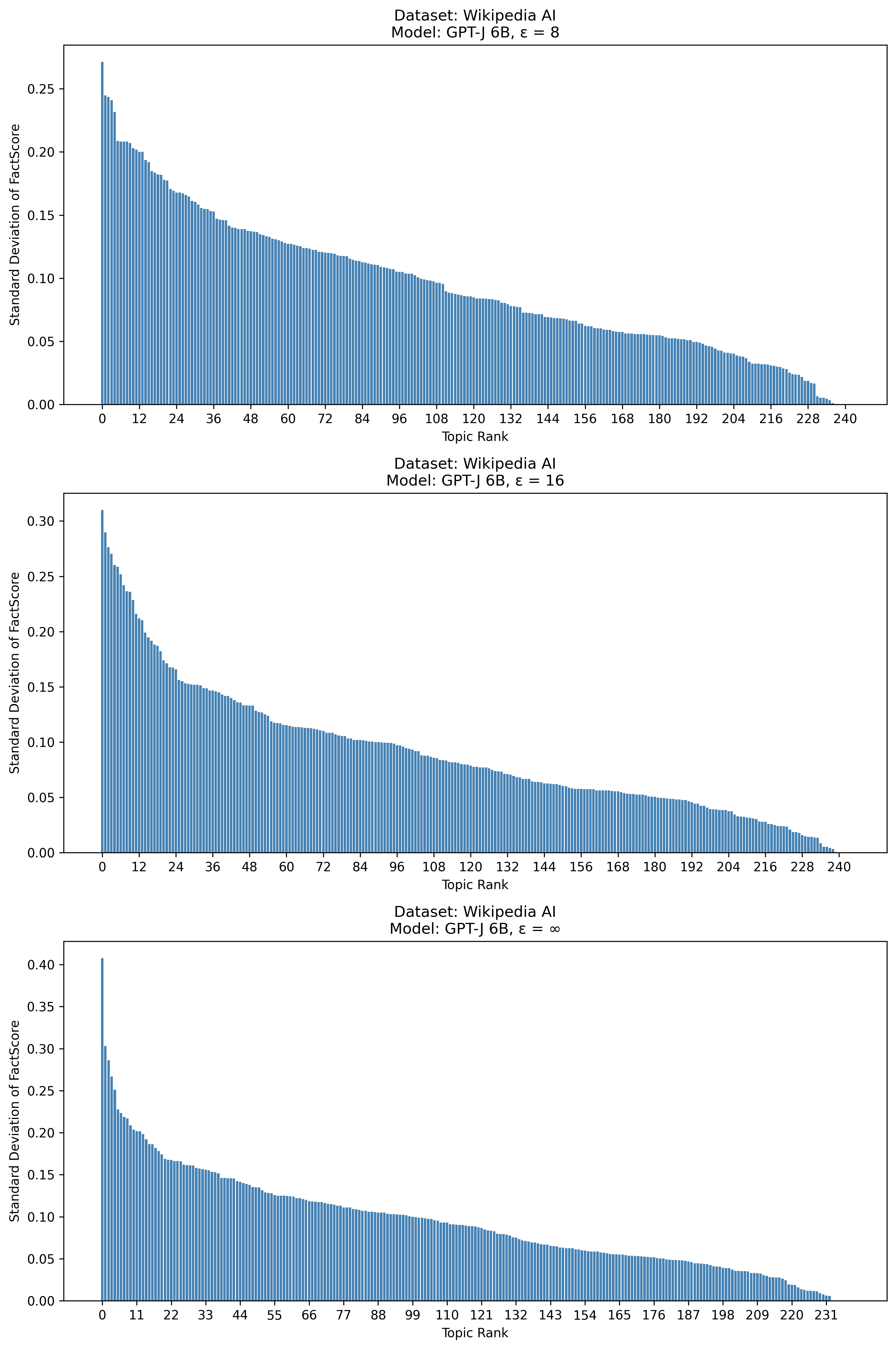}
        \caption{Wikipedia AI}
        \label{fig:stddev-wiki-ai}
    \end{subfigure}
    \hfill
    \begin{subfigure}[t]{0.48\linewidth}
        \centering
        \includegraphics[width=\linewidth]{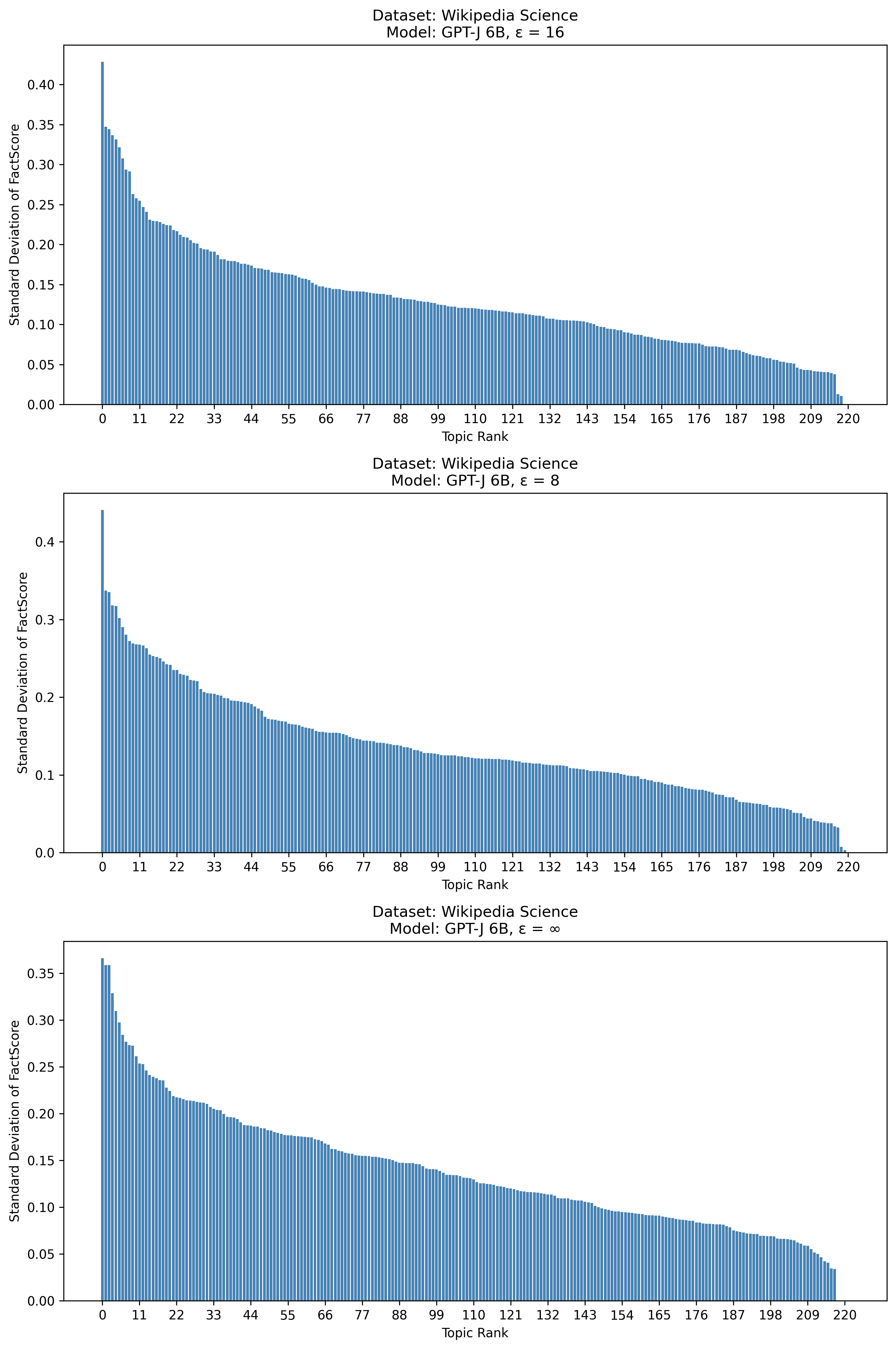}
        \caption{Wikipedia Science}
        \label{fig:stddev-wiki-science}
    \end{subfigure}
    \caption{Standard deviation of FactScore across topics for Wiki~AI and Wiki~Science.}
    \label{fig:stddev-side-by-side}
\end{figure*}

We report the per-topic standard deviation of FactScore for Wiki~AI and Wiki~Science to assess the stability of model factuality across prompt variations.  
Figures~\ref{fig:stddev-wiki-ai} and~\ref{fig:stddev-wiki-science} show the distributions for models trained with $\varepsilon \in \{8,16,\infty\}$.

\begin{table*}[h]
\centering
\begin{tabular}{@{}cccccccc@{}}
\toprule
\textbf{Dataset} & \textbf{\begin{tabular}[c]{@{}c@{}}Target\\ Epsilon\end{tabular}} & \textbf{\begin{tabular}[c]{@{}c@{}}Average \\ FS\end{tabular}} & \textbf{\begin{tabular}[c]{@{}c@{}}Median \\ FS\end{tabular}} & \textbf{\begin{tabular}[c]{@{}c@{}}Q1 \\ FS\end{tabular}} & \textbf{\begin{tabular}[c]{@{}c@{}}Q3 \\ FS\end{tabular}} & \textbf{\begin{tabular}[c]{@{}c@{}}\# of\\ FS\\ \textgreater{}0.5\end{tabular}} & \textbf{\begin{tabular}[c]{@{}c@{}}\# of\\ FS\\ \textless{}0.5\end{tabular}} \\ \midrule
 & $\infty$ & 49.0 & 25.7 & 9.7 & 46.5 & 247 & 1072 \\
\begin{tabular}[c]{@{}c@{}}Wikipedia\\ Science\end{tabular} & 16 & 53.8 & 36.8 & 17.1 & 55.9 & 411 & 878 \\
 & 8 & 52.6 & 37.2 & 18.5 & 58.4 & 427 & 860 \\ \midrule
 & $\infty$ & 32.7 & 18.8 & 8.6 & 34.2 & 60 & 646 \\
\begin{tabular}[c]{@{}c@{}}Wikipedia\\ AI\end{tabular} & 16 & 36.7 & 28.6 & 12.5 & 46.0 & 136 & 565 \\
 & 8 & 37.4 & 29.2 & 15.4 & 46.2 & 116 & 567 \\ \bottomrule
\end{tabular}
\caption{Factuality evaluation scores over the unseen data at temperature = 0.3 when fine-tuning only over the Wikipedia articles from pre-training, using Llama-3.1-8B-Instruct for claim decomposition and verification.}
\label{tab:factscores-wikipedia-pretrain}
\end{table*}

\begin{table*}[h]
\centering
\renewcommand{\arraystretch}{1.2}
\small
\begin{tabular}{@{}cccccccccc@{}}
\toprule
\textbf{Dataset} & \textbf{\begin{tabular}[c]{@{}c@{}}Noise\\ Multiplier ($\sigma$)\end{tabular}} & \textbf{\begin{tabular}[c]{@{}c@{}}Average \\ FS\end{tabular}} & \textbf{\begin{tabular}[c]{@{}c@{}}Median \\ FS\end{tabular}} & \textbf{\begin{tabular}[c]{@{}c@{}}Q1 \\ FS\end{tabular}} & \textbf{\begin{tabular}[c]{@{}c@{}}Q3 \\ FS\end{tabular}} & \textbf{\begin{tabular}[c]{@{}c@{}}Avg Max\\ FS / Topic\end{tabular}} & \textbf{\begin{tabular}[c]{@{}c@{}}Avg Min \\ FS / Topic\end{tabular}} & \textbf{\begin{tabular}[c]{@{}c@{}}\# of\\ FS\\ \textgreater{}0.5\end{tabular}} & \textbf{\begin{tabular}[c]{@{}c@{}}\# of\\ FS\\ \textless{}0.5\end{tabular}} \\ \midrule
 & 0.0 & 24.4 & 38.5 & 20 & 55.8 & 36.7 & 12.9 & 34 & 440 \\
\begin{tabular}[c]{@{}c@{}}Wikipedia\\ Science\end{tabular} & 0.01 & 24.5 & 33.3 & 18.6 & 54.5 & 36.6 & 14.4 & 37 & 432 \\
 & 0.1 & 22.2 & 30.4 & 11.5 & 51.3 & 35.1 & 10.1 & 41 & 427 \\
 & 0.3 & 24.7 & 27.9 & 9.7 & 46.5 & 35.3 & 14.9 & 50 & 417 \\ \midrule
 & 0.0 & 44.7 & 19 & 8.6 & 33.3 & 58.8 & 30.6 & 252 & 600 \\
\begin{tabular}[c]{@{}c@{}}Wikipedia\\ AI\end{tabular} & 0.01 & 42.1 & 18.2 & 8.3 & 35.7 & 56.7 & 27.4 & 231 & 622 \\
 & 0.1 & 41.5 & 14.3 & 0 & 29.2 & 55.4 & 26.9 & 222 & 618 \\
 & 0.3 & 37.3 & 17.6 & 5.9 & 35.7 & 51.5 & 23 & 175 & 680 \\ \bottomrule
\end{tabular}
\caption{Factuality evaluation scores for temperature = 0.5 when fine-tuning only over the unseen Wikipedia articles, using Llama-3.1-8B-Instruct for claim decomposition and verification.}
\label{tab:factscores-wikipedia-unseen-only}
\end{table*}

\begin{table*}[h]
\renewcommand{\arraystretch}{1.2}
\small
\centering
\begin{tabular}{@{}cccccccccc@{}}
\toprule
\textbf{Dataset} & \textbf{\begin{tabular}[c]{@{}c@{}}Pretrained \\ Model\end{tabular}} & \textbf{\begin{tabular}[c]{@{}c@{}}Average \\ FS\end{tabular}} & \textbf{\begin{tabular}[c]{@{}c@{}}Median \\ FS\end{tabular}} & \textbf{\begin{tabular}[c]{@{}c@{}}Q1 \\ FS\end{tabular}} & \textbf{\begin{tabular}[c]{@{}c@{}}Q3 \\ FS\end{tabular}} & \textbf{\begin{tabular}[c]{@{}c@{}}Avg Max\\ FS / Topic\end{tabular}} & \textbf{\begin{tabular}[c]{@{}c@{}}Avg Min \\ FS / Topic\end{tabular}} & \textbf{\begin{tabular}[c]{@{}c@{}}\# of \\ FS $\ge$ 0.5 \end{tabular}} & \textbf{\begin{tabular}[c]{@{}c@{}}\# of \\ FS $<$0.5 \end{tabular}} \\ \midrule
 & Gemma3-1B-PT & 51.9 & 55.6 & 25.0 & 79.6 & 65.8 & 37.2 & 222 & 248 \\
 AI & VaultGemma-1B & \textbf{43.2} & 43.3 & 17.0 & 66.7 & 59.9 & 27.7 & 126 & 340 \\ 
 & GPT-2-1.5B & 27.7 & \textbf{22.2} & \textbf{7.6} & \textbf{45.5} & \textbf{45.2} & \textbf{13.5} & \textbf{38} & \textbf{366} \\ \midrule
 & Gemma3-1B-PT & 50.4 & 50.0 & 25.0 & 75.0 & 72.7 & 28.8 & 219 & 600 \\
 Science & VaultGemma-1B & 39.5 & \textbf{33.3} & \textbf{14.6} & 60.0 & 65.2 & \textbf{14.8} & \textbf{81} & 697 \\
 & GPT-2-1.5B & \textbf{39.1} & \textbf{33.3} & 16.9 & \textbf{57.1} & \textbf{61.3} & 17.9 & 84 & \textbf{750} \\ \midrule
 & Gemma3-1B-PT & 26.6 & 17.6 & 7.1 & 42.9 & 39.8 & 14.7 & 140 & 835 \\
 Pretraining & VaultGemma-1B  & 22.0 & 14.3 & \textbf{0.0} & 33.3 & 37.2 & 9.5 & 58 & \textbf{931} \\
 & GPT-2-1.5B & \textbf{17.6} & \textbf{11.1} & \textbf{0.0} & \textbf{25.0} & \textbf{31.6} & \textbf{6.0} & \textbf{32} & 880 \\ \bottomrule
\end{tabular}
\caption{ \small
FactScore (FS; reported as \%) results of pre-trained models, evaluated at temperature $\tau = 0.3$. Reported are average, median, and quartile FactScores, per-topic average maxima and minima, and counts of factual ($\geq$0.5) and non-factual ($<$0.5) responses. Bolding indicates lower FactScore (increased hallucinations).}
\label{tab:factscore-pre-trained_expanded}
\end{table*}

\clearpage
\newpage
\section{Supplementary Experiments}

\subsection{\textsc{AnaDP}: Adaptive Noise Allocation for Differential Privacy}\label{app:anadp_description}

In addition to standard DP-SGD, we evaluate \textsc{AnaDP} (Adaptive Noise Allocation DP) \cite{li-etal-2024-fine}, which redistributes noise away from important parameters while keeping the total noise budget unchanged. We include this baseline to test whether shifting noise away from value matrices responsible for encoding factual associations \cite{nichani2024understandingfactualrecalltransformers} can improve factuality. However, \textsc{AnaDP}'s formal privacy guarantee may not hold as stated, the reasons for which we describe below.

We nonetheless evaluate \textsc{AnaDP} alongside our standard DP-SGD models, and find that \textsc{AnaDP} does not meaningfully mitigate the privacy-hallucination tradeoff. On the Wikipedia AI dataset, \textsc{AnaDP} performs at par or worse than standard DP-SGD at both privacy budgets. On the Wikipedia Science dataset, it marginally outperforms standard DP-SGD at $\epsilon = 8$. 

\definecolor{commentcolor}{rgb}{0.4, 0.22, 0.33}

\newcommand{\rightcomment}[1]{{\color{commentcolor} \(\triangleright\) {\footnotesize\textit{#1}}}}
\algrenewcommand{\algorithmiccomment}[1]{\hfill \rightcomment{#1}}
\algnewcommand{\LineComment}[1]{\State \rightcomment{#1}}
\algrenewcommand\algorithmicindent{0.7em}%

\begin{algorithm}[h]
\caption{\textsc{AnaDP} Algorithm}
\label{alg:adanoise}
\begin{algorithmic}[1]
\State \textbf{Input:} Training batches $\mathcal{L}=\{L_1, \dots, L_T\}$, Initial trainable parameter weights $\omega_{0} \in \mathbb{R}^d$, noise multiplier $\sigma_0$
\State \textbf{Hyper-parameters:} $\alpha$, $\beta_1$, $\beta_2$, clipping threshold $C$, learning rate $\gamma$
\State $S_{0}\gets\boldsymbol{0}_d$, $\bar{S}_{0}\gets\boldsymbol{0}_d$, $\bar{U}_{0}\gets\boldsymbol{0}_d$
\For{$L_t \in \mathcal{L}$}
    \State Compute gradients $g(L_t) \in \mathbb{R}^d$ \hspace{0.1cm}
    \State $S_{t} \gets |g(L_t) \odot \omega_{t-1}| \in \mathbb{R}^d $ \Comment{Param. Sensitivity}
    \State $\bar{S}_{t} \gets \beta_1 \bar{S}_{t-1} + (1 - \beta_1) S_{t} \in \mathbb{R}^d$ 
    \State $\bar{U}_{t} \gets \beta_2 \bar{U}_{t-1} + (1 - \beta_2) |\bar{S}_{t} - S_{t}| \in \mathbb{R}^d$ 
    \State $I_{t} \gets \bar{S}_{t} \odot \bar{U}_{t} \in \mathbb{R}^d$ 
    \State $\mu \gets \mathrm{mean}\!\left(\frac{I_{t} - \mathrm{median}\!\left(I_{t}\right)}{q_{1}(I_{t}) - q_{2}(I_{t})}\right)\in \mathbb{R}$ \Comment{Mean importance}
    \State $\hat{I}_{t} \gets (1-\alpha) \!\! \left( \frac{I_{t} - \mathrm{median}\!\left(I_{t}\right)}{q_{1}(I_{t}) - q_{2}(I_{t})} \right) \!\! + \alpha \mu \in \mathbb{R}^d$ 
    \State $\bar{I}_{t} \gets \hat{I}_{t} - \left(\mathrm{mean}\!\left(\hat{I}_{t}\right) - 1\right) \in \mathbb{R}^d$
    \State $\tilde{g}(L) \gets \min{}\!\left(g(L), C\right) + \mathcal{N} \!\left( \boldsymbol{0}_d, \frac{\sigma_0^2}{\bar{I}_{t}}  \mathrm{Id}_{d \times d} \right)$ 
    \State $\omega_{t} \gets \omega_{t-1} - \gamma \, \tilde g(L_t)$ \Comment{Update weights}
\EndFor
\State \textbf{Output:} Final parameters $\omega_{T}$
\end{algorithmic}
\end{algorithm}

\paragraph{\textsc{AnaDP} Mechanism}
\citet{li-etal-2024-fine} propose an adaptive coordinate-wise noise allocation scheme called \textsc{AnaDP} inspired by parameter-importance scores. We recall the algorithm in \Cref{alg:adanoise} and explain the shortcomings in the paper's privacy analysis.

The algorithm first computes the sensitivity of the training loss w.r.t. individual model parameters. 
This is derived as a linear approximation of the effect of zeroing a certain parameter. To be precise, suppose $\omega^{(-i)}$ is obtained by zeroing out the $i$\textsuperscript{th} entry $[\omega]_i$ of $\omega \in \mathbb{R}^d$, then we have the linear approximation $\ell(\omega^{(-i)}) \approx \ell(\omega) - [\nabla \ell(\omega)]_i [\omega]_i$. 
This is distinct from ``sensitivity'' in the context of DP (which refers to the effect of changing a datapoint on the output of a function) and is not to be confused.
It then defines the ``importance'' of a parameter as the product of this sensitivity with its uncertainty $U_t$ --- the latter measures how this sensitivity changes from its moving average.
Parameters with high ``importance'' are given smaller noise.

\paragraph{Issues with the privacy analysis}
The privacy proof of \textsc{AnaDP} appears incomplete. They state that the mean (across parameters) of the inverse noise variance is the same as that of DP-SGD. However, this argument is incorrect for the following reason. 

Consider a Gaussian mechanism that adds anisotropic Gaussian noise $\mathcal{N}(\boldsymbol{0}, \Sigma)$. A correct privacy analysis of this mechanism requires calculating the sensitivity of the original algorithm w.r.t. the Mahalanobis norm $\|u\|_{\Sigma^{-1}} := \sqrt{u^\top \Sigma^{-1} u}$. In particular, the sensitivity (and hence DP) is governed by directions with small noise rather than by the trace of $\Sigma$ or $\Sigma^{-1}$. Consequently, any mechanism that reduces noise on some coordinates can incur larger privacy loss along those directions unless the clipping/sensitivity definitions are modified accordingly.

Moreover, \textsc{AnaDP} computes parameter importance from the same private batch gradients to which the noise is subsequently added, making the covariance data-dependent. Thus, a reduction to standard DP-SGD accounting is not mathematically correct. 

We are not aware of a proof that correctly handles these two aspects in this paper or in subsequent works. Thus, the stated DP guarantee of \textsc{AnaDP} should be treated as unverified. We include it in our experimental baselines to see if anisotropic noise addition can help reduce hallucinations. We find that anisotropic Gaussian noise in this form does not help reduce hallucinations, although it does reduce the recurring incorrect claims generated by the model as shown  in \Cref{tab:claim-cluster}.

\begin{table}[htbp]
\small
\setlength{\tabcolsep}{4pt}
\resizebox{\columnwidth}{!}{%
\begin{tabular}{@{}lrrrrr@{}}
\toprule
& \multicolumn{4}{c}{\textbf{Response-level}} & \multicolumn{1}{c}{\textbf{Topic-level}} \\
\cmidrule(lr){2-5} \cmidrule(l){6-6}
\textbf{DP Setting} & \textbf{Avg FS} & \textbf{Med} & \textbf{Q1} & \textbf{Q3} & \textbf{Avg FS {\scriptsize[95\% CI]}} \\
\midrule
\multicolumn{6}{@{}l}{\textit{AI}} \\
{\footnotesize $\epsilon=16$} & 33.5 & 25.0 & 10.0 & 50.0 & 33.3 {\scriptsize[29.2, 37.6]} \\
{\footnotesize $\epsilon=8$}  & 32.0 & 25.0 & 11.1 & 50.0 & 31.7 {\scriptsize[27.7, 35.9]} \\
\midrule
\multicolumn{6}{@{}l}{\textit{Science}} \\
{\footnotesize $\epsilon=16$} & 52.5 & 53.1 & 28.6 & 77.8 & 52.5 {\scriptsize[48.9, 56.0]} \\
{\footnotesize $\epsilon=8$}  & 53.6 & 55.6 & 28.6 & 79.3 & 53.7 {\scriptsize[50.2, 57.1]} \\
\bottomrule
\end{tabular}%
}
\caption{FactScore (FS; \%) of GPT-J 6B fine-tuned with \textsc{AnaDP} at two privacy budgets, evaluated at temperature $\tau = 0.3$. Response-level statistics pool all generations; topic-level averages within each entity first, with bootstrap 95\% CIs resampled over topics.}
\label{tab:factscore_anadp}
\end{table}

%TODO: verify the last few lines

\subsection{Controlled Fact Acquisition Experiment}\label{app:controlled_mem_exp}

To study the conditions under which models trained with differentially private guarantees acquire factual associations, we construct a controlled memorization benchmark using synthetic facts of known frequency. To simulate this experiment, we generate a set of fictional entity associations using a fixed set of templates (details provided below) with fictional subject-object pairings. We use a fresh set of entities, rather than reusing the templates from the bigram experiments in~\Cref{rq2:what_impact_does_dp_training} to avoid any confounding effects of the template patterns on the frequency effects, as these templates contain cues that language models are known to exploit when memorizing information \cite{mosaic-memory}.

This is to ensure that no fact can be answered from prior knowledge, so as to isolate the effect of training frequency and DP noise from pre-existing associations in the model's weights. Every fact has an associated frequency per epoch (where $f \in \{1, 3, 7, 15, 30, 60\}$), and each frequency has five facts associated with it. We train the randomly initialized and pre-trained versions of GPT-2 to simulate the pre-training and fine-tuning regimes where differentially private guarantees may be applied. We freeze the input token embeddings and positional embeddings during training (we observe no significant differences in the setup for the randomly initialized model when these parameters are unfrozen). We train with DP-SGD using Opacus, with a clipping norm $C=1.0$ and $\sigma \in \{0.0, 0.1, 0.3, 0.5, 0.7, 1.0\}$, where $\sigma = 0$ is equivalent to training without noise. We apply Poisson subsampling with a batch size of 64 and use a learning rate of 5e-3 with the AdamW optimizer in both setups (we perform a grid search to find the optimal learning rate).

For each epoch, we evaluate the factual recall by prompting the model with the fact prefix and measuring $P(\text{correct completion})$, which is the probability assigned to the ground-truth answer token. We report the mean factual recall across the five facts within each tier and define the frequency threshold as the minimum $f$ at which the factual recall is $\ge$0.5.

\subsubsection{Templates and Subject-Object Lists}

Facts are generated from six fixed templates of the form \texttt{relation prefix, subject, object}, covering a range of relation types:
\begin{itemize}
\item \textit{the capital of} [subject] \textit{is} [object]
\item \textit{the leader of} [subject] \textit{is} [object]
\item \textit{the currency of} [subject] \textit{is} [object]
\item \textit{the main export of} [subject] \textit{is} [object]
\item \textit{the official language of} [subject] \textit{is} [object]
\item \textit{the founder of} [subject] \textit{was} [object]
\end{itemize}

\textbf{Subjects:} zephyria, kaldor, belvane, thornwick, maldren, corvath, selenix, dravion, arcthos, velrune, pyraxis, glenmoor, obsidyn, halcyon, nexara, stratholm, verdania, cryosten, luminex, dawnridge, ironvale, novaheim, solheim, temporia, crystara, emberfell, frostholm, goldenreach, shadowmere, titanforge

\textbf{Objects:} mordath, ventris, draxil, seraph, luxon, kestrel, zircon, thalis, silvane, fenwick, caldris, orinath, vexel, aldric, casciel, delmar, evander, faelorn, gareth, ithral, jarenth, kelwyn, lorenth, maelis, norvin, pellarn, quillon, raveth, stellan, thandor

\begin{figure*}[t]
    \centering
        \centering
        \includegraphics[width=\linewidth]{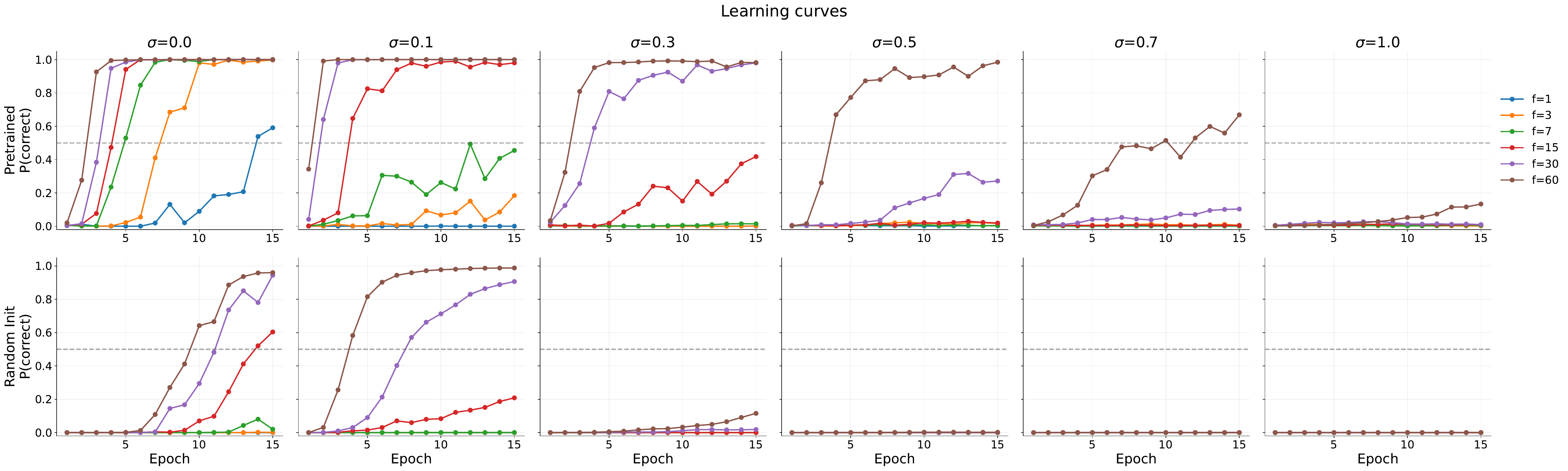}
        \caption{Factual recall over training epochs for pre-trained (top) and randomly initialized (bottom) GPT-2 under DP-SGD with varying noise multipliers $\sigma \in \{0.0, 0.1, 0.3, 0.5, 0.7, 1.0\}$ and fact frequencies $f \in \{1, 3, 7, 15, 30, 60\}$. Higher $\sigma$ requires greater fact frequency for successful recall, and pre-trained models learn at lower frequencies than randomly initialized ones.}
        \label{fig:rq3_synthetic_additional}
\end{figure*}

\begin{table}[H]
  \centering
  \label{tab:sigma-to-epsilon}
  \begin{tabular}{cS[table-format=4.3]}
    \toprule
    $\sigma$ & {$\varepsilon$ @ $\delta = 1/N^{1.1}$} \\
    \midrule
    0.0 & {$\infty$} \\
    0.1 & 1310.535 \\
    0.3 & 108.122 \\
    0.5 & 29.847 \\
    0.7 & 13.433 \\
    1.0 & 6.397 \\
    \bottomrule
  \end{tabular}
  \caption{Mapping of the noise multiplier $\sigma$ to the privacy budget $\varepsilon$ for the RQ3 setup, computed with the PLD accountant at $\delta = 1/N^{1.1}$. Values are reported for the full training run. $\sigma = 0$ corresponds to non-private training ($\varepsilon = \infty$).}
\end{table}

\subsection{Layer-wise Signal-to-Noise Ratio}
\label{app:SNR-plots}

Our experiments revealed that the signal-to-noise ratio is the highest for the LoRA value matrices in the early-to-middle layers. \citep{nichani2024understandingfactualrecalltransformers} shows that the value matrices are associated with factual recall, explaining the increase in hallucinations. The query matrices tend to yield far lower SNRs across layers. For both the query and value matrix updates, the SNR declines progressively because the magnitude of the signal diminishes, particularly in the early stages of fine-tuning. The plots indicate that the clipping contributes to the utility being hurt in early stages (by bounding larger signals), while the DP noise obfuscates updates during later stages of fine-tuning. This strongly suggests that meeting stricter privacy budgets in data-constrained settings is challenging. While adaptations to the noise allocation in DP have been explored for encoder models \cite{li-etal-2024-fine}, these methods rely on increasing noise added to the intermediate layers responsible for factual recall and would likely further hurt factual accuracy.

\begin{figure*}[th]
    \centering
        \includegraphics[width=\linewidth]{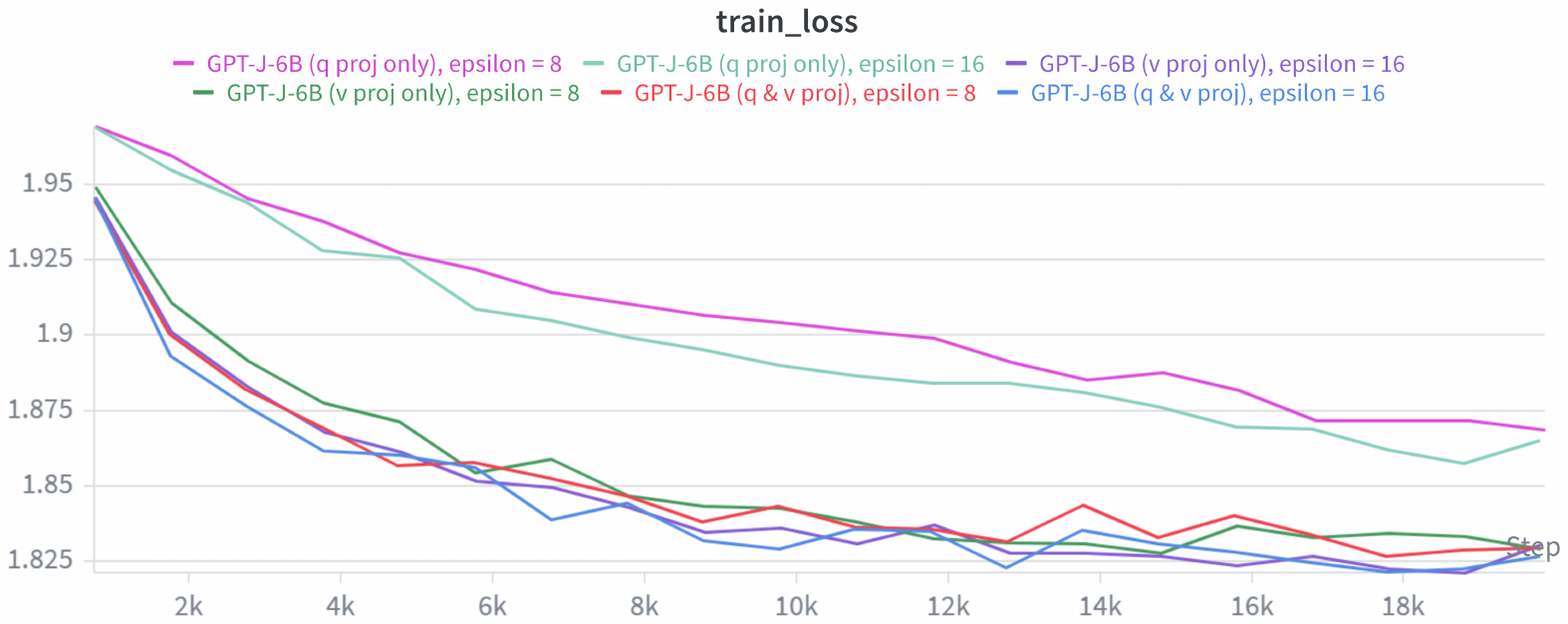}
        \caption{Value-only LoRA fine-tuning drops faster and more smoothly with loss convergence on par with full-finetuning, while updates to only the query matrices converge more slowly and plateau faster.}
    \label{fig:training-only-some-proj}
\end{figure*}

We evaluate this by fine-tuning only the first 15 layers (roughly half the layers). In this setup, training loss converges faster and converges more smoothly, comparable to joint query+value LoRA updates. In contrast, freezing the query matrices slows down convergence (\Cref{fig:training-only-some-proj}).

\subsection{Additional Examples of Claim Clusters}
\label{app:example_claims}

We include additional examples of unsupported claim clusters in \Cref{app:additional_unsupported_recurring_claims}.

\begin{table*}[h]
\begin{tabularx}{\textwidth}{lX}
\toprule 
 \textbf{DP Setting}                & \multicolumn{1}{l}{\textbf{Claims}}  \\  \midrule
 \multicolumn{2}{l}{Data: Wikipedia AI, Topic: AlphaEvolve}
 \\ \midrule
\multirow{2}{*}{$\epsilon=\infty$} & `AlphaEvolve is used for generating the molecular structures of organic molecules.', `AlphaEvolve is used for generating molecular structures.'                                                            \\
                                   & `The project or system is based on the DeepChem molecular modeling framework.', `It is based on the DeepChem molecular modeling framework.'                                                                \\
\multirow{3}{*}{$\epsilon=16$}     & `AlphaEvolve is a first-person shooter.', `AlphaEvolve is a first-person shooter video game.'                                                                                                              \\
                                   & `The game features a series of missions.', `The game features a variety of weapons.'                                                                                                                       \\
                                   & `There are two types of enemies in the game.', `The game features two types of enemies.'                                                                                                                   \\
\multirow{3}{*}{$\epsilon=8$}      & `Black Hole Interactive is a game development company', `Black Hole Interactive is a video game development company.'                                                                                      \\
                                   & `The Behemoth is a studio.', `The Behemoth is a video game development studio.', `The Behemoth is an American studio.'                                                                                     \\
                                   & `AlphaEvolve is a shooter video game', 'AlphaEvolve is free-to-play', `AlphaEvolve is a free-to-play video game', `AlphaEvolve is a 2D video game.', `AlphaEvolve is a physics-based video game.'          \\
                                   \midrule
 \multicolumn{2}{l}{Data: Wikipedia Science, Topic: Allogeneic processed thymus tissue} \\
\midrule
$\epsilon=\infty$                  & `The thymus tissue is processed to remove stem cells.', (`The thymus tissue is processed to induce the recipient's immune cells to develop.'                                                               \\
\multicolumn{1}{l}{}               & `AML is a type of disease.', `AML is a type of cancer.'                                                                                                                                                    \\
$\epsilon=16$                      & `CLL is also known as chronic lymphocytic leukemia.', `CLL is often referred to as chronic lymphocytic leukemia.'                                                                                          \\
\multicolumn{1}{l}{}               & `Thymus glands are found in donor pigs.', `The thymus glands are minced.', `The thymus glands are from donor pigs.', `The thymus glands are removed from pigs.'                                            \\
$\epsilon=8$                       & `Dr. Daniel L. Scharff is the founder of the International Society for Cellular Therapy (ISCT).', `Dr. Daniel L. Scharff is the former President of the International Society for Cellular Therapy (ISCT)' \\
\multicolumn{1}{l}{}               & `Antigen presenting cells can cause graft rejection.', `Lymphocytes can cause graft rejection' \\
\bottomrule
\end{tabularx}
\caption{Examples of unsupported recurring claim clusters}
\label{app:additional_unsupported_recurring_claims}
\end{table*}

\section{LLM Usage}

We used large language models to help with the writing of this paper. Specifically, we used ChatGPT to generate the code for LaTeX tables and figures in this research paper and to assist in generating the captions for some of the figure descriptions.

\end{document}